\documentclass[twoside]{article}

\usepackage{preprint}

\usepackage[american]{babel}

\usepackage[round]{natbib}
\usepackage{booktabs} %
\usepackage{tikz} %

\usepackage[T1]{fontenc} %
\usepackage{graphicx} %
\usepackage{microtype}
\usepackage[breaklinks]{hyperref} %

\input{preamble}
\newcommand{\circright}[1][]{%
    \ifx\relax#1\relax
        \mathrel{\circ\mkern-7.4mu\rightarrow}%
    \else 
        \mathrel{\stackrel{#1}{\circ\mkern-7.4mu\rightarrow}}%
    \fi 
}

\newcommand{\circleft}[1][]{%
  \ifx\relax#1\relax
    \mathrel{\leftarrow\mkern-7.4mu\circ}%
  \else
    \mathrel{\stackrel{#1}{\leftarrow\mkern-7.4mu\circ}}%
  \fi
}

\newcommand{\circboth}[1][]{%
  \ifx\relax#1\relax
    \mathrel{\circ\mkern-7mu-\mkern-7.4mu\circ}%
  \else
    \mathrel{\stackrel{#1}{\circ\mkern-7mu-\mkern-7.4mu\circ}}%
  \fi
}

\newcommand{\bidiedge}[1][]{%
  \ifx\relax#1\relax
    \mathrel{\leftarrow\mkern-10mu\rightarrow}%
  \else
    \mathrel{\stackrel{#1}{\leftarrow\mkern-10mu\rightarrow}}%
  \fi
}

\newcommand{\myleftarrow}[1][]{%
  \ifx\relax#1\relax \leftarrow \else \xleftarrow{#1} \fi
}

\newcommand{\myrightarrow}[1][]{%
  \ifx\relax#1\relax \rightarrow \else \xrightarrow{#1} \fi
}

\newcommand{\mysim}[1][]{%
  \ifx\relax#1\relax \sim \else \stackrel{#1}{\sim} \fi
}

\newcommand{\rcirclearrow}{\hspace{-0.12em} \leftarrow \hspace{-0.15cm}\circ\hspace{0.06em}} %
\newcommand{\lcirclearrow}{\hspace{+0.06em}\circ\hspace{-0.16cm}\rightarrow \hspace{-0.12em}} %

\newcommand{\cE}{\ensuremath \mathcal{E}}
\newcommand{\cK}{\ensuremath \mathcal{K}}
\newcommand{\cU}{\ensuremath \mathcal{U}}
\newcommand{\cI}{\ensuremath \mathcal{I}}
\newcommand{\cP}{\ensuremath \mathcal{P}}
\newcommand{\cG}{\ensuremath \mathcal{G}}

\newcommand{\scc}{\ensuremath \operatorname{scc}}
\newcommand{\anc}{\ensuremath \operatorname{anc}}
\newcommand{\des}{\ensuremath \operatorname{des}}
\newcommand{\sib}{\ensuremath \operatorname{sib}}
\newcommand{\pa}{\ensuremath \operatorname{pa}}
\newcommand{\ch}{\ensuremath \operatorname{ch}}
\newcommand{\coll}{\ensuremath \operatorname{coll}}
\newcommand{\ncoll}{\ensuremath \operatorname{ncoll}}
\newcommand{\edges}{\ensuremath \operatorname{edges}}
\newcommand{\nodes}{\ensuremath \operatorname{nodes}}
\newcommand{\en}{\ensuremath \operatorname{endnodes}}

\DeclareMathOperator{\indep}{\perp\!\!\!\perp} %
\DeclareMathOperator{\notindep}{\not\! \perp\!\!\!\perp} %

\newcommand\restr[2]{{%
  \left.\kern-\nulldelimiterspace %
  #1 %
  \vphantom{\big|} %
  \right|_{#2} %
  }}
\newcommand{\cond}{\: : \:}
\newcommand{\equivcls}[1]{%
  #1/\!{\sim}%
}

\newcommand{\defined}[1]{{\color{NavyBlue}\emph{#1}}}
\newcommand{\symbdef}[1]{{\color{ForestGreen}#1}}

\usepackage[dvipsnames]{xcolor}
\definecolor{DarkRed}{rgb}{0.7, 0.1, 0.1}

\newcommand{\xhdr}[1]{\noindent \textbf{#1}\:}

\begin{document}

\runningauthor{Simon Kempf, Niki Kilbertus, Georg Manten}

\setlength\footskip{25pt}\fancyfoot[C]{\thepage}

\renewcommand{\thefootnote}{\fnsymbol{footnote}}
\twocolumn[

\papertitle{Markov-Equivalent Conditional Independence Graphs on Path Spaces}

\paperauthor{ Simon Kempf\footnotemark[1] \And Niki Kilbertus \And Georg Manten }

\paperaddress{ LMU Munich \\ MCML \And Technical University of Munich \\ Helmholtz Munich \\ MCML \And Technical University of Munich \\ Helmholtz Munich \\ MCML } ]
\footnotetext[1]{Work done while at Technical University of Munich and Helmholtz Munich.}
\renewcommand{\thefootnote}{\arabic{footnote}}

\begin{abstract}
Causal discovery for continuous-time stochastic processes with latent confounding typically faces a trade-off between (i) approaches that are highly informative about the causal structure but are difficult to test statistically, and (ii) approaches that admit practical tests from data but yield more coarse-grained conclusions, potentially not leveraging the direction of time. We introduce W-separation, an asymmetric graphical criterion for directed mixed graphs defined as a quaternary relation that allows distinct conditioning on past and future components.
This added flexibility yields a highly informative graphical independence model while preserving testability. W-separation satisfies a global Markov property for a broad class of stochastic differential equations with latent confounding, linking it to conditional independence relations among coordinate processes that are testable from data.
A central result of this work is the full characterization of W-Markov equivalence classes: we show that every equivalence class admits a unique greatest element that serves as a canonical representative and a principled target for learning algorithms. In addition, we provide necessary and sufficient conditions under which directed or bidirected edges can be added without altering the induced independence model.
\end{abstract}
\section{Introduction \& Related Work}\label{sec:intro}

Causal relationships underpin scientific reasoning across disciplines, and inferring these relations from observational data remains a primary objective in many domains \citep{glymour2019review,brouillard2025}. Such relationships are predominantly represented through graphical models, by leveraging a correspondence between graph structures and independence relations in the observed data distribution \citep{pearl1995causal,lauritzen2001causal}. While the majority of existing literature focuses on static models with symmetric independence relations, applying such a framework directly to dynamical systems fails to leverage that causal influence strictly flows from past to future.
Capturing this fundamental constraint requires an asymmetric independence model, which can provide more informative causal inferences than symmetric alternatives.

Various frameworks have been proposed to address causal discovery in dynamical systems, yet each faces distinct limitations. In the following, we review several prominent approaches to highlight the current gaps in this field and motivate the need for a more expressive, time-asymmetric separation criterion. One prominent framework is local independence, which models the dependence of the present on the past history of coordinate processes, capturing arbitrarily fast interactions \citep{schweder1970composable,mogensen2018causal,didelez2008}. While theoretically elegant and capable of handling latent processes \citep{mogensen2020markov}, local independence remains difficult to test statistically \citep{christgau2023,liu2025conditionallocalindependencetesting}; currently, no general tests exist for broad classes of processes such as diffusions. Alternatively, many methods rely on discretized time grids, ``unrolling'' the causal graph into a discrete autoregressive model \citep{runge2018causal,runge2020discovering,runge2019detecting,runge2023causal,quinn2015directedinformationgraphs}. However, these approaches are fundamentally limited by their reliance on fixed, regular sampling and are highly sensitive to the chosen time lag and sampling frequency.

Recent literature has shifted toward path-segment approaches to handle continuous-time data.  \citet{laumann2023kernel} developed a conditional independence test for such data; however, their framework relies on established results from static, acyclic models and thus ignores the inherent temporal direction. \citet{manten2024sigker} established both a conditional independence test on path-space and a time-aware Markov property for SDEs by splitting the time into a past and a future segment, yielding more informative discovery, but remaining essentially acyclic (allowing only self-loops). \citet{boekenmooij2024DSCMs} uses symmetric $\sigma$-separation in simple SCMs \citep{Forre_Constraint} for path-valued variables to accommodate cycles, but without leveraging the time direction. While this allows for cyclic dependencies, %
the reliance on a symmetric independence notion fails to exploit the ``arrow of time'', leading to less informative causal discovery.

To address these shortcomings, \citet{manten2025} introduced a novel asymmetric separation criterion, named E-separation. This framework is practically testable via the conditional independence tests of \citet{manten2024sigker} and simultaneously accounts for cycles, latent confounding, and the direction of time.
However, E-separation is limited by its characterization properties; importantly, \textbf{(i)} we show that it fails to establish a fundamental equivalence between graphical inseparability and the existence of inducing walks in DMGs, and \textbf{(ii)} the Markov equivalence class of directed mixed graphs (DMGs) has not been identified, and thus for systems with latent confounding the learned graphs would lack a clear causal interpretation.

Our work addresses this gap by introducing W-separation, a more expressive graphical criterion that overcomes the fundamental shortcomings of E-separation. We characterize separability under W-separation and fully characterize the corresponding Markov equivalence classes of DMGs, providing graphical foundations for causal discovery in cyclic dynamical systems with latent confounding that leverages the direction of time.
We further establish a global Markov property for a broad class of path-dependent SDE models, which ties W-separation to conditional independences between the coordinate processes that are testable from data, e.g., via a signature-kernel-based conditional independence test on path space \citep{manten2024sigker}.

\section{Dynamical Systems, Graphs and Separation}
To motivate our graphical formalism with a concrete data-generating mechanism, we consider a multivariate stochastic process $X = (X^1, \ldots, X^d)$ with $X^j = (X_t^j)_{t \in [0,1]}$ for $j \in [d]$ evolving according to a stationary, path-dependent SDE-model of the form 
    \begin{align}\label{eq:SDE_model}
        \begin{cases}
        dX_t^{k} = \mu^{k} (X_{[0,t]}) dt + \sigma^{k} (X_{[0,t]}) dW^{k}_t,  \\
        X_0^{k} = x_0^{k},
        \end{cases}
    \end{align}
for $t \in [0,1]$, $k \in [d]$, with $X_t^k \in \mathbb{R}^{n_k}$, $n_k \in \mathbb{N}_{>0}$, Brownian motions $W_t^k \in \mathbb{R}^{m_k}$, $m_k \in \mathbb{N}_{>0}$ and drift- and diffusion functionals $\mu^k : C^{0}([0,1],\mathbb{R}^{n}) \rightarrow \mathbb{R}^{n_k}$, $\sigma^k : C^{0}([0,1],\mathbb{R}^{n}) \rightarrow \mathbb{R}^{n_k \times m_k}$ that may depend on the full solution history up to time $t$ and that satisfy the standard Lipschitz and growth conditions, ensuring existence and pathwise uniqueness; see \Cref{app:sde_assumptions}. Furthermore, we assume the initial conditions $x^k_0$ and Brownian motions are jointly independent. The coordinate process dependence in \Cref{eq:SDE_model} defines a directed ``dependency graph'' $\cG = (V, \cE_{d})$ with vertices $V = [d] := \{1, \ldots, d\}$ representing the individual coordinate processes and where, for any $i,j\in[d]$, we include the directed edge $i\to j$ whenever $\mu^j$ or $\sigma^j$ is non-constant in its $i$-th argument.

In practice, typically only a subset $V_{obs} \subseteq V$ of the coordinate processes is observed, while latent processes $V_{lat} = V \setminus V_{obs}$ may drive several observed coordinates simultaneously. Such latent confounding cannot be expressed by directed edges among observed nodes alone. Marginalizing the latent processes from the dependency graph yields its latent projection onto $V_{obs}$ (\cref{def:latent_projection}), a directed mixed graph (DMG) whose directed edges encode (possibly latent-mediated) dependences among observed processes and whose bidirected edges encode latent confounding. The global Markov property carries over to this setting (\cref{prop:global_MP_latent_models}); hence we develop the theory for general DMGs.

While \cref{eq:SDE_model} serves as a concrete data-generating mechanism, the graph-theoretic results below are not tied to SDEs: they concern DMGs and their W-separation independence models, and SDEs only enter through the global Markov property (\cref{prop:W-sep_global_MP}), which links the graph to conditional independences on path space that are testable from data. The SDE model \eqref{eq:SDE_model} includes deterministic ODEs ($\sigma^k \equiv 0$) and requires no equilibrium assumption, and the graphical theory applies equally to, e.g., discrete-time vector autoregressive (VAR) processes with lagged dependences and jointly independent noise.

\subsection{Basic Graph Terminology}
Our notation and terminology follows \citet{mogensen2020markov,forre2023causality,peters2017elements, manten2025}; we recall the necessary concepts below.

\begin{definition}[Directed (Mixed) Graph, D(M)G]
A \defined{directed mixed graph (DMG)} is a triple $\cG = (V, \cE_{d}, \cE_{bi})$ consisting of a \textit{node set} $V\cong [d]$, $d \in \mathbb{N}$, together with two edge sets: the \defined{directed edges} $\cE_{d} \subset V \times V =: \cE_{d}(V)$ and the \defined{bidirected edges} $\cE_{bi} \subset \equivcls{\left((V \times V)\right)}=: \cE_{bi}(V)$, where $\Delta:=\lbrace (v,v) \mid v \in V \rbrace$ and the equivalence relation is given by $(v_1, v_2) \sim (v_3, v_4) : \Leftrightarrow \left( (v_1 = v_3) \land (v_2 = v_4) \right) \lor \left( (v_1 = v_4) \land (v_2 = v_3) \right)$, with equivalence classes written \symbdef{$[(j,k)]$}. If $\cE_{bi} = \emptyset$, then $\cG$ is called a \defined{directed graph (DG)}.
\end{definition}

\begin{remark}
Unlike ADMGs in the static setting, a DMG may be cyclic, may contain self-loops $v \to v$ (as $\Delta \subseteq \cE_{d}(V)$), and may contain both a directed and a bidirected edge between the same pair of nodes, e.g., $v \to w$ and $v \leftrightarrow w$. Both arise naturally in dynamical systems \citep{mogensen2020markov}, where a process may depend on its own past and $v$ may affect $w$ directly while sharing a latent cause with it.
\end{remark}

Throughout, $\cG$ denotes a DG or DMG, $V \cong [d]$ its node set (associated with coordinate processes), and $\cE_d$, $\cE_{bi}$ its respective edge sets.
For brevity, we write \symbdef{$v \in \cG$} for $v \in V$, \symbdef{$v_1 \to v_2 \in \cG$} for $(v_1, v_2) \in \cE_{d}$ (the edge is said to have a \defined{tail} at $v_1$ and a \defined{head} at $v_2$, collectively referred to as \defined{edge marks}), and \symbdef{$v_1 \leftrightarrow v_2 \in \cG$} for $[(v_1,v_2)]_{bi} \in \cE_{bi}$.
We further write \symbdef{$v_1 \circright v_2 \in \cG$} to mean $v_1 \to v_2 \in \cG$ or $v_1 \leftrightarrow v_2 \in \cG$, and \symbdef{$v_1 \sim v_2 \in \cG$} or \symbdef{$v_1 \circboth v_2 \in \cG$} to mean that $v_1 \to v_2 \in \cG$ or $v_1 \leftarrow v_2 \in \cG$ or $v_1 \leftrightarrow v_2 \in \cG$, i.e., that $v_1$ and $v_2$ are \defined{adjacent} in $\cG$.
The symbol $\circ$ acts as a placeholder for either an `arrowhead' or a `tail'.

For $v,w \in \cG$, a \defined{walk from $v$ to $w$ in $\cG$} is a finite sequence $\{ (v_k, e_k) \}_{k \in [n]}$ with $v_0 := v$, $v_{n+1}:= w$, and $e_k \in \{ (v_k, v_{k+1}), (v_{k+1}, v_{k}),[v_k, v_{k+1}]_{bi} \}$ for all $k \in [n]$, often written as $v = v_0 \overset{e_0}{\sim} v_1 \overset{e_1}{\sim} \cdots\overset{e_n}{\sim} v_{n+1} = w$.
The \defined{inverse walk} is denoted \symbdef{$\pi^{-1}$}, i.e., $w = v_{n+1} \overset{e_n}{\sim} v_n \overset{e_{n-1}}{\sim} \cdots \overset{e_0}{\sim} v_{0} = v$.
The \defined{trivial walk} from $v$ to $v$, written \symbdef{$\emptyset$}, consists of a single node $v \in \cG$ and no edges.
A walk $\pi: v \sim \cdots \sim w$ is called \defined{bidirected} (\defined{directed}) if $e_k = [ v_k, v_{k+1} ]_{bi}\in \cE_{bi}$ (resp.\ $e_k = (v_k, v_{k+1}) \in \cE_d$, i.e., $v \to \cdots \to w$) for every $k \in [n]$.
A walk $v=v_{0} \sim \cdots \sim v_{n+1}=w$ is a \defined{path from $v$ to $w$ in $\cG$} if $\lvert \{ v_0, \cdots , v_n \} \rvert = n+1$, i.e., no node other than the endpoint appears more than once. The notions of (bi)directedness transfer directly from walks to paths.

Two walks $\pi_1 = v \overset{e_0^1}{\sim} v^1_1 \overset{e_1^1}{\sim} \cdots \overset{e_n^1}{\sim} w$ and $\pi_2 = v \overset{e_0^2}{\sim} v^2_1 \overset{e_1^2}{\sim} \cdots \overset{e_m^2}{\sim} w$ are called \defined{endpoint-identical} if the marks of $e_0^1$ and $e_0^2$ at $v$, and of $e_n^1$ and $e_m^2$ at $w$, coincide.
A DMG $\cG = (V, \cE_{d}, \cE_{bi})$ is \defined{acyclic} if no non-trivial directed walk $v \rightarrow \cdots \rightarrow v$ exists for any $v \in V$; such a graph is an \defined{acyclic directed mixed graph (ADMG)}, and a \defined{directed acyclic graph (DAG)} if additionally $\cE_{bi} = \emptyset$.

Given DMGs $\cG_1 = (V, \cE^1_{d}, \cE^1_{bi})$ and $\cG_2 = (V, \cE^2_{d}, \cE^2_{bi})$, we call $\cG_2$ a \defined{supergraph} of $\cG_1$ (and $\cG_1$ a \defined{subgraph} of $\cG_2$), written \symbdef{$\cG_1 \subseteq \cG_2$}, if $\cE^1_d \subseteq \cE^2_d$ and $\cE^1_{bi} \subseteq \cE^2_{bi}$. Note that $\subseteq$ defines a partial order on the set of DMGs over $V$. The \defined{complete DG (DMG)} is defined as the graph containing all possible edges, $(V, \cE_d = V \times V)$ (resp.\ $(V, \cE_d = V \times V, \cE_{bi}=\equivcls{(V \times V)})$).

For a node $v$ in a DMG $\cG$, we define its
\defined{parents} \symbdef{$\pa^\cG(v)$} $:= \{w \in \cG \mid w \to v \in \cG\}$,
\defined{children} \symbdef{$\ch^\cG(v)$} $:= \{w \in \cG \mid v \to w \in \cG\}$,
\defined{siblings} \symbdef{$\sib^\cG(v)$} $:= \{w \in \cG \mid w \leftrightarrow v \in \cG\}$,
\defined{ancestors} \symbdef{$\anc^\cG(v)$} $:= \{w \in \cG \mid \exists \: \text{directed path} \: w \to \cdots \to v \in \cG \}$, and
\defined{descendants} \symbdef{$\des^\cG(v)$} $:= \{w \in \cG \mid \exists \: \text{directed path} \: v \to \cdots \to w \in \cG\}$.
All of these extend to node sets via unions, e.g., $\pa^\cG(A) := \bigcup_{v\in A} \pa^\cG(v)$.
Since trivial paths are permitted, we have $v \in \anc^\cG(v)$, $A \subseteq \anc^\cG(A)$, $v \in \des^\cG(v)$, and $A \subseteq \des^\cG(A)$.

In the presence of cycles, we additionally require the \defined{strongly connected component of $v$ in $\cG$}, defined as \symbdef{$\scc^\cG(v)$} $:= \anc^\cG(v) \cap \des^\cG(v)$, which satisfies $v \in \scc^\cG(v)$.
The \defined{set of strongly connected components} of $\cG$ is \symbdef{$\mathbf{S}(\cG)$} $:= \{ A \subseteq V \mid \exists\, j \in V : A = \scc^\cG(j) \}$; the relation $v \sim_{1} w :\Leftrightarrow w \in \scc^\cG(v)$ defines an equivalence relation on $V$ whose equivalence classes partition $V$.

\xhdr{$\sigma$-Separation.}
Let $\cG = (V, \cE_{d}, \cE_{bi})$ be a DMG and $\pi = v_0 \overset{e_0}{\sim} \cdots \overset{e_n}{\sim} v_{n+1}$ a walk. A node $v_k$ (equivalently, a position $k \in \{0,\ldots,n+1\}$) on $\pi$ is called a \defined{non-collider on $\pi$} if it is an endpoint ($k \in \{0,n+1\}$), lies in a left-chain ($v_{k-1} \leftarrow v_{k} \circleft v_{k+1}$), lies in a right-chain ($v_{k-1} \circright v_{k} \rightarrow v_{k+1}$), or lies in a fork ($v_{k-1} \leftarrow v_k \rightarrow v_{k+1}$).
It is a \defined{collider on $\pi$} if $v_{k-1} \circright v_k \circleft v_{k+1}$.
We write $\symbdef{\coll_{\pi}}$ and $\symbdef{\ncoll_{\pi}}$ for the sets of colliders and non-colliders of $\pi$, respectively.

A non-collider $v_k$ on $\pi$ (with $k \notin \{0,n+1\}$) is called an \defined{unblockable non-collider on $\pi$} if it satisfies one of the following:
it lies in a left-chain ($v_{k-1} \leftarrow v_{k} \circleft v_{k+1}$) with $v_{k-1} \in \scc^\cG(v_k)$;
it lies in a right-chain ($v_{k-1} \circright v_{k} \rightarrow v_{k+1}$) with $v_{k+1} \in \scc^\cG(v_k)$;
or it lies in a fork ($v_{k-1} \leftarrow v_k \rightarrow v_{k+1}$) with $v_{k-1} \in \scc^\cG(v_k)$ and $v_{k+1} \in \scc^\cG(v_k)$.
Otherwise, $v_k$ is a \defined{blockable non-collider on $\pi$}. We write $\symbdef{\ncoll_{\pi,\mathrm{bl}}}$ and $\symbdef{\ncoll_{\pi,\mathrm{ubl}}}$ for these two sets, respectively.

We use \textit{$\sigma$-separation} \citep{forre2017markovpropertiesgraphicalmodels}, the extension of $d$-separation \citep{pearl_d_sep} to cyclic graphs.

\emph{$\sigma$-separation:} A walk $\pi =  v_0 \overset{e_0}{\sim} \cdots \overset{e_n}{\sim} v_{n+1}$ is \defined{$\sigma$-open given $C$} (or \defined{$\sigma$-$C$-open}) for $C \subseteq V$ if $\coll{\pi} \subseteq \anc^\cG(C)$ and $\ncoll_{\pi,\mathrm{bl}} \cap C = \emptyset$; otherwise it is \defined{$\sigma$-blocked given $C$} (or \defined{$\sigma$-$C$-blocked}), meaning $\coll{\pi} \not\subseteq \anc^\cG(C)$ or $\ncoll_{\pi,\mathrm{bl}} \cap C \neq \emptyset$.
Extending this to sets, we say $A$ is \defined{$\sigma$-separated from $B$ given $C$} and write \symbdef{$A \indep_{\sigma}^\cG B \mid C$}; otherwise we write $\symbdef{A \notindep_{\sigma}^\cG B \mid C}$.

\section{DMGs on Path Spaces}
To understand the motivation for our work, we first recall the graphical framework introduced by \citet{manten2025} and point out its limitation. This framework centers on the \emph{lifted dependency graph} (see \cref{fig:counter_example} for an illustration), which explicitly represents the temporal dependence structure between coordinate processes.

\xhdr{Lifted dependency graph.}
For a DMG $\cG = (V, \cE_d, \cE_{bi})$, the \defined{lifted dependency graph} \symbdef{$\tilde{\cG}$} is the DMG $\tilde{\cG} = (\tilde{V}, \tilde{\cE}_d, \tilde{\cE}_{bi})$, where $\tilde{V} := V_0 \sqcup V_1$ is a disjoint union of two copies of $V$, with nodes subscripted by $0/1$ to indicate set membership and edges 
$\tilde{\cE}_{d}:= \{(u_0,v_0), (u_0,v_1), (u_1,v_1) \text{ if } u\neq v \text{ and } (u_0,v_1) \text{ if } u=v \cond (u,v) \in \cE_{d} \}$ 
and $\tilde{\cE}_{bi}:= \{ [(u_0,v_0)], [(u_0,v_1)], [(u_1, v_0)], [(u_1,v_1)] \cond [(u,v)] \in \cE_{bi} \}$. 

We may refer to $V_0$ and $V_1$ as the set of past and future nodes, respectively.
Intuitively, $v_0$ and $v_1$ represent the path segments $X^v_{[0,s]}$ and $X^v_{[s,s+h]}$, making the direction of time explicit. Unlike unrolling a time-series graph, two copies suffice without fixing a maximal lag, as all lagged influences on the future segment are absorbed into its dependence on the past segment.
Based on this structure, \emph{E-separation} was proposed as a ternary separation criterion designed to capture the fundamental asymmetry of time. 

\xhdr{E-Separation.}
    For a DMG $\cG = (V , \cE_{d},\cE_{bi})$, $B$ is \defined{E-separated} from $A$ given $C$ in $\cG$ if
    \begin{align*}
        A_0 \indep_{\sigma}^{\tilde{\cG}} B_1 \mid C_0, C_1\setminus B_1,
    \end{align*}
    where $A,B,C \subset V$, \symbdef{$A_0$} $= \{ a_0 : a \in A \} \subseteq \tilde{V}$ and \symbdef{$B_1$} $= \{ b_1 : b \in B \} \subseteq \tilde{V}$, $C_0:= \{ c_0 : c \in C \} \subseteq \tilde{V}$, $C_1:= \{ c_1 : c \in C \} \subseteq \tilde{V}$ in the lifted dependency graph $\tilde{\cG}$ of $\cG$.
    We denote this E-separation by $\symbdef{A \indep_{E}^{\cG} B \mid C}$.

\emph{Inducing walks} provide a key structural concept which, in traditional causal discovery, are expected to fully characterize inseparability. %
The concept has been used in various graph classes and their graphical separation criteria (e.g. \citet{verma_pearl1990} for ADMGs, \citet{jiri_zhang_fci} for ancestral graphs, \citet[Chapter 11]{forre2023causality} for CDMGs 
and $\sigma$-separation, or \citet{mogensen2020markov} for DMGs and $\mu$-separation). 
\citet{manten2025} define an asymmetric version of inducing walks that adapts to the asymmetry in their E-separation. We recall their definition of \emph{asymmetric inducing walks}.

\xhdr{Asymmetric inducing walks.}
    Let $\cG = (V, \cE_d, \cE_{bi})$ be a DMG and $v,w \in V$. An \defined{(asymmetric) inducing walk from $v$ to $w$} in $\cG$ is a non-trivial walk $\pi: v=v_0 \overset{e_0}{\sim} v_1  \overset{e_1}{\sim} \cdots \overset{e_{n-1}}{\sim} v_n \overset{e_n}{\sim} v_{n+1} = w$ such that $\coll_\pi \subseteq \anc({\lbrace v,w \rbrace })$, each $v_i \in \ncoll_\pi \setminus \lbrace v_0 \rbrace$ is unblockable, and the last edge before entering $\scc(w)$ has to be into the $\scc(w)$. 

However, we show that E-separation is fundamentally limited in its characterization properties in the presence of latent confounding. Specifically, we prove that for general Directed Mixed Graphs (DMGs), the existence of an a-inducing walk is not equivalent to graphical inseparability. In the counterexample illustrated in \cref{fig:counter_example}, two nodes remain inseparable under E-separation even though no a-inducing walk connects them. This failure has direct implications for causal discovery: in classical DAG, MAG, and ADMG theory, inducing structures explain inseparability and underpin the interpretation of adjacencies in maximal graphical representations. Under E-separation, skeleton discovery can instead retain additional adjacencies that lack an a-inducing-walk interpretation, yielding a less informative supergraph and leaving no clear structural basis for proceeding beyond skeleton discovery.

We introduce \textit{W-separation}, a more expressive separation criterion designed to overcome these structural shortcomings. By extending E-separation into a quaternary relation, W-separation allows for independent conditioning on past and future nodes. This flexibility restores the equivalence between graphical inseparability and the existence of a-inducing walks, allowing fewer adjacencies to be retained while giving them a structural interpretation. Beyond this equivalence, we fully characterize the W-Markov equivalence classes of DMGs. Notably, we prove that each class admits a unique greatest element, which serves as its canonical representative and a principled target for structure learning.

\subsection{Separability in W-Separation DMGs}
We now define \emph{W-separation} via $\sigma$-separation on the lifted dependency graph and show that \emph{inseparability} is equivalent to the existence of a-inducing walks.
\begin{definition}[W-Separation] \label{def:s_sep}  
For a DMG $\cG = (V , \cE_{d},\cE_{bi})$, $B$ is \defined{W-separated} from $A$ given $C$ and $D$ in $\cG$ if
$$    
A_0 \indep_{\sigma}^{\tilde{\cG}} B_1 \mid C_0 \cup D_1
$$
where $A,B,C,D \subset V$, $A_0 = \{ a_0 : a \in A \} \subseteq \tilde{V}$ and $B_1 = \{ b_1 : b \in B \} \subseteq \tilde{V}$, $C_0:= \{ c_0 : c \in C \} \subseteq \tilde{V}$, $D_1:= \{ d_1 : d \in D \} \subseteq \tilde{V}$ in the lifted dependency graph $\tilde{\cG}$ of $\cG$.
We denote this W-separation by $\symbdef{A \indep_{W}^{\cG} B \mid C,D}$.
\end{definition}

\begin{definition}[W-independence model]\label{def:s_sep_quaternary}
    Let $\cG = (V , \cE_{d},\cE_{bi})$ be a DMG. Then $\indep_{W}^{\cG}$ defines the following quaternary relation on the node set $V$, which we call the \defined{W-independence model} of $\cG$:
    $$
    \symbdef{\cI_{W} := \cI_{W}^{\cG}} := \{  (A,B,C,D) \in \cP (V)^4 : A \indep_{W}^{\cG} B \mid C,D\},
    $$
    where \symbdef{$\cP(V)$} denotes the power set of $V$.
\end{definition}

\begin{remark}
    In the sequel, it suffices to focus on singleton sets in the first two components as W-separation decomposes there, i.e. $(A,B,C,D) \in \cI $ if and only if $ (\{a\},B,C,D) \in \cI \,\, \forall a \in A$ and $ (A,\{b\},C,D) \in \cI \,\, \forall b \in B$, following directly from its definition through $\sigma$-separation.
\end{remark}
We define \emph{separability} for abstract independence models.
\begin{definition}[Separability]
Let $\cI \subseteq \cP(V)^4$ be an independence model over $V$, $a,b \in V$. We call $b$ \defined{separable} from $a$ if there exists a $C \subseteq V \setminus \{ a \}, D \subseteq V \setminus \{ b \}$ such that $(\{ a \},\{ b \},C,D) \in \cI$ and otherwise
\defined{inseparable}.
We denote the set of nodes inseparable from a node $b$ by 
\begin{multline*}
    \symbdef{u (b, \cI)} := \left\{ a \in V \cond (\lbrace a \rbrace,\lbrace b \rbrace,C,D) \notin \cI, \right.\\
    \left. \forall C \subseteq V \setminus \lbrace a \rbrace,D \subseteq V \setminus \{ b \} \right\} .
\end{multline*}
\end{definition}

\begin{figure*}[t]  
    \centering
    \includegraphics[width=0.93\linewidth]
    {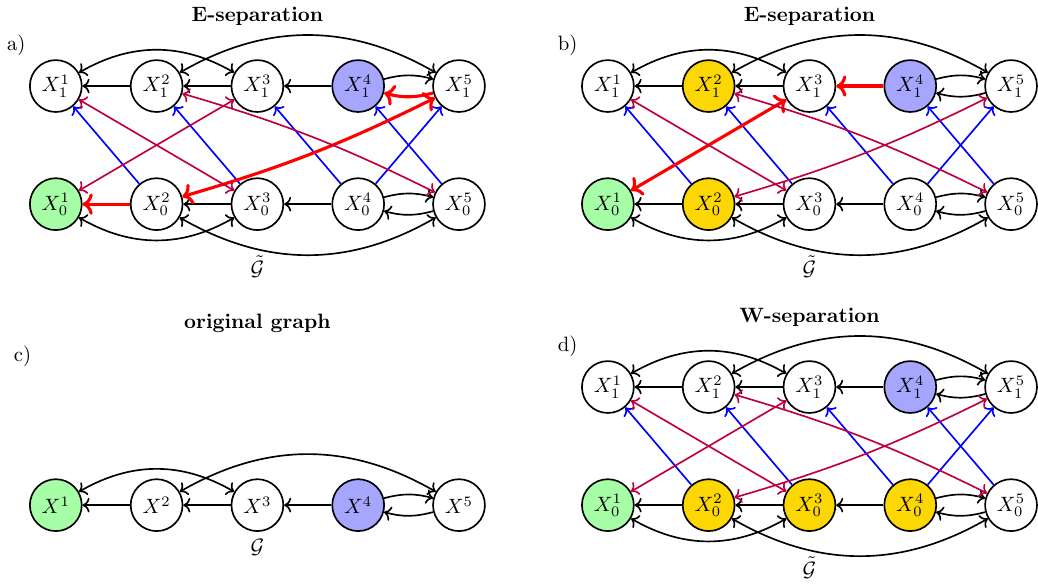}
    \caption{The DMG $\cG$ in c) shows that E-separation fails to characterize inseparability through a-inducing walks; a), b), d) show its lifted dependency graph $\tilde{\cG}$.
    There is no a-inducing walk from $X^1$ (green) to $X^4$ (blue) in $\cG$, yet $X^4$ cannot be E-separated from $X^1$ by any $C\subseteq\{X^2,X^3,X^4,X^5\}$:
    if $X^2 \notin C$, as in a), the walk $X^1_0 \myleftarrow X^2_0 \bidiedge X^5_1 \myrightarrow X^4_1$ (red) is $\sigma$-$(C_0 \cup C_1 \setminus \{X_1^4\})$-open; if $X^2 \in C$ (yellow, b)), the walk $X^1_0 \bidiedge X^3_1 \myleftarrow X^4_1$ opens, since $X^3_1\in \anc(X^2_1)$.
    In contrast, d) shows that $X^4$ is W-separated from $X^1$ by $C=\{X^2,X^3,X^4\}, D=\emptyset$ (see \cref{fig:future_conditioning} for an example where separation requires $D \neq \emptyset$): conditioning independently on past and future nodes blocks the walk in a) without opening the one in b).
    }
    \label{fig:counter_example}
\end{figure*}

\cref{prop:a_ind_eqv_s_sep} shows that two nodes are inseparable under W-separation if and only if they are connected by an a-inducing walk. This property is central in the characterization of the independence model and marks a significant advantage over E-separation for which this equivalence fails (see \cref{fig:counter_example}).

\begin{proposition} \label{prop:a_ind_eqv_s_sep}
Let $\cG = (V, \cE_d, \cE_{bi})$ a DMG, $v,w \in V$. Let $\tilde{\cG} = (\tilde{V}, \tilde{\cE}_d, \tilde{\cE}_{bi})$ be the lifted dependency graph of $\cG$. Then the following are equivalent:

\begin{enumerate}[label=\roman*)]
    \item There exists an a-inducing walk %
    from $v$ to $w$ in $\cG$.
    
    \item There exists an a-inducing path %
    from $v$ to $w$ in $\cG$.
    
    \item $(\{v\}, \{w\}, C, D) \notin \cI_W^\cG $ for every $ C,D \subseteq V $, i.e. $v\in u (w, \cI)$.
\end{enumerate}
\end{proposition}
The proof is provided in \cref{app:proof_aind_ssep} and leverages the flexibility of constructing independent conditioning sets for past and future nodes.

\subsection{Markov Property for the SDE-model}
Markov properties relate aspects of a graph to independence relations in the underlying causal process and can be viewed as rules on how independence relations can be read off the corresponding graph. We here introduce a time-aware asymmetric conditional independence notion for our SDE-model \cref{eq:SDE_model} and show that it obeys the global Markov property for directed graphs w.r.t W-separation.
\begin{proposition}[\defined{Global Markov property for $W$-separation}]\label{prop:W-sep_global_MP}
Let $X$ solve \eqref{eq:SDE_model} under assumptions (A1)--(A4) in \cref{app:sde_assumptions}, and let $\cG=(V,\cE_d)$ be its dependency graph.
For $s,h>0$ with $s+h\le1$, define $\hat{X}^k:=(X_t^k-X_s^k)_{t\in[s,s+h]}$.
Then, for all $A,B,C,D\subseteq V$,
\begin{align}
A \indep^{\cG}_{W} B \mid C,D
\Rightarrow X^A_{[0,s]} \indep \hat{X}^B
\mid X^C_{[0,s]},\hat{X}^D.
\end{align}
\end{proposition}
Using future increments removes the additive carry-over $X_s^k$, so the lifted edge $k_0\to k_1$ is present only when the drift or diffusion depends on $X^k$, as encoded by a self-loop $k\to k$ in $\cG$.
The proof in \cref{app:proof_markov_property} applies the $\sigma$-separation Markov property of \citet[Theorem~B.2]{Forre_Constraint} to an SCM-like functional representation of \eqref{eq:SDE_model}, following \citet{manten2024sigker,manten2025,boekenmooij2024DSCMs}.
The required measurable subsystem solution maps are constructed using \citet[Theorem~3.1]{Przybylowicz2024}, following its use in this context by \citet{boekenmooij2024DSCMs,boekenmooij2026SDEs}.\\
Latent confounding, i.e., dependences among observed coordinate processes induced by unobserved components of the underlying dynamical system can typically not be represented by directed edges but require bidirected edges. The resulting mixed graph of the fully observed dynamical systems graph is constructed via the latent projection: 
\begin{definition}[The latent projection]\label{def:latent_projection}
 Let $\cG = (V, \cE_d, \cE_{bi})$ be a DMG, $V = V_{obs} \sqcup V_{lat}$. The \defined{latent projection of $\cG$} onto the observed node set $V_{obs}$ is the DMG $\cG'=(V_{obs},\cE_d',\cE_{bi}')$ with edge-sets defined as follows:
 \begin{enumerate}
 \item[(i)]  $(v_1,v_2)\in \cE_d'$  iff $v_1,v_2\in V_{obs}$ and there exists a non-trivial directed walk $v_1\to \cdots \to v_2$ whose intermediate nodes lie in latent node set $V_{lat}$
 \item[(ii)] $[(v_1,v_2)]\in \cE_{bi}'$  iff $v_1,v_2\in V_{obs}$ and there exists a non-trivial walk $\pi = v_1 \rcirclearrow \cdots \lcirclearrow v_2$ whose intermediate nodes lie in $V_{lat}$ and that has no colliders, $\mathrm{coll}_{\pi}=\emptyset$.
\end{enumerate}
\end{definition}
Analogous to \citet[Prop. 3.16, Prop. 3.17]{manten2025}, marginalizing a directed graph to a DMG preserves the W-separation independence relations among observed nodes, from which we can extend the global Markov property to the setting with latent confounding:
\begin{proposition}[Global Markov property for latent models]\label{prop:global_MP_latent_models}
Let $X$ solve \eqref{eq:SDE_model} under assumptions (A1)--(A4) in \cref{app:sde_assumptions}, and let $\cG=(V,\cE_d)$ be its dependency graph.
Partition $V=V_{obs}\sqcup V_{lat}$ into observed and latent nodes, and let $\cG'$ be the latent projection of $\cG$ onto $V_{obs}$.
For $s,h>0$ with $s+h\le1$, define $\hat{X}^k:=(X_t^k-X_s^k)_{t\in[s,s+h]}$ for $k\in V_{obs}$.
Then, for all $A,B,C,D\subseteq V_{obs}$,
\begin{align}\label{eq:sigma_global_MP_hidden}
A \indep^{\cG'}_{W} B \mid C,D
\Rightarrow X^A_{[0,s]} \indep \hat{X}^B
\mid X^C_{[0,s]},\hat{X}^D.
\end{align}
\end{proposition}
The statement follows from \cref{prop:W-sep_global_MP} and \citet[Lemma~3.3.8]{forre2023causality}, since $\sigma$-separation is preserved under latent marginalization and our lifting commutes with latent projection.

\subsection{Markov Equivalence of DMGs}

The central result of this section is that every \emph{Markov equivalence class} of DMGs possesses a \emph{greatest element}, namely a DMG that is a supergraph of all other elements, serving as a canonical representative of the class. 
We establish this result by adapting an approach of \citet{mogensen2020markov} to our setting; for a DMG $\cG = (V,\cE_d,\cE_{bi})$  we define a set of \emph{potential siblings} and a set of \emph{potential parents} for each node $a\in V$. We use these sets to construct a Markov equivalent supergraph of $\cG$ that is maximal. By proving that all Markov equivalent graphs possess the same sets of potential siblings and parents, we show that this supergraph is identical for all elements of the equivalence class $[G]_W$ and is thus a greatest element. 

\begin{definition}[Markov equivalence]\label{def:markov_eqv}
Let $\cG^1 = (V, \cE_{d}^1, \cE_{bi}^1)$, $\cG^{2} = (V, \cE_{d}^{2}, \cE_{bi}^2)$ be two DMGs over a common node set $V$. We say that $\cG^1$ and  $\cG^2$ are \defined{(W-separation) Markov equivalent} if $\cI_{W}^{\cG^1} = \cI_{W}^{\cG^2}$. W-separation Markov equivalence induces an equivalence relation on the set of DMGs over $V$, and we denote the \defined{(Markov) equivalence class} of $\cG^{1}$  by \symbdef{$[\cG^1]_{W}$}.
\end{definition}

\begin{lemma}\label{lem:equivalent_DMGs_inducing_walks}
    Let $\cG^1 = (V, \cE_{d}^1, \cE_{bi}^1)$, $\cG^{2} = (V, \cE_{d}^{2}, \cE_{bi}^2)$ be two Markov equivalent DMGs and $v,w\in V$. If there exist  an a-inducing walks from $v$ to $w$ in $\cG^1$, then there exist an a-inducing walks from $v$ to $w$ in $\cG^2$.
\end{lemma}

In words: two Markov equivalent graphs share the same set of nodes connected by a-inducing walks.

\begin{definition}[Maximal DMGs]
We call a DMG $\cG = (V, \cE_d, \cE_{bi})$  \defined{maximal}, if $\cG$ is complete or if adding any edge changes the induced independence model $\cI_W^\cG$. 
\end{definition}

\begin{definition}[Greatest element]
    We say that $\cG_0$ is the \defined{greatest element} of an equivalence class $[\cG_1]_W$ if $\cG_0 \in [\cG_1]_W$ and $\cG_2 \subseteq \cG_0$ for all $\cG_2 \in [\cG_1]_W$.
\end{definition}

In words: the greatest element is a supergraph of all elements in its Markov equivalence class, and hence maximal. \Cref{thr:greatest_dmg} shows that it always exists and is the only maximal DMG in its class.

\paragraph{Potential Siblings.}
We define \emph{potential siblings} directly through the independence model. Consequently, all graphs within a single Markov equivalence class share the same set of potential siblings, making them identifiable from the independence constraints alone.

\begin{definition}[Potential Siblings]\label{def:potential_siblings}
    Let $\cI \subseteq \cP(V)^4$ be an independence model over $V$ and let $v,w \in V$. We call $v$ and $w$ \defined{potential siblings} in $\cI$ if %
    $v\in u (w, \cI)$ and $w\in u (v, \cI)$.
\end{definition}

\begin{proposition}\label{prop:pot_sib_in_G}
    Let $\cG = (V, \cE_d, \cE_{bi})$ be a DMG, $v,w \in V$. Then $v$ and $w$ are potential siblings in $\cI_{W}^{\cG}$ if and only if there exist both an a-inducing path from $v$ to $w$ in $\cG$ and an a-inducing path from $w$ to $v$ in $\cG$.
\end{proposition}

\begin{lemma}\label{lem:sib_are_pot_sib}
    Let $\cG = (V, \cE_d, \cE_{bi})$ be a DMG, $v,w \in V$ and $[(v,w)] \in \cE_{bi}$. Then $v$ and $w$ are potential siblings in $\cI_{W}^{\cG}$.
\end{lemma}

\begin{proposition}%
\label{prop:adding_bidi_edges}
Let $\cG=\left(V, \cE_{d}, \cE_{bi}\right)$ be a DMG and $v, w \in V$.
Let $[(v, w)] \notin \cE_{bi}$ and denote $\cG'=(V, \cE_{d},\cE_{bi}' )$, where $\cE_{bi}' := \cE_{bi} \cup\{[(v, w)]\}$. 
Then $\cI_{W}^{\cG}=\cI_{W}^{\cG'}$ if and only if $v$ and $w$ are potential siblings in $\cI_{W}^{\cG}$.
\end{proposition}

In words: If $v$ and $w$ are potential siblings, then adding $v\bidiedge w$ to $\cG$ does not change the induced independence model. This motivates the term \textit{potential} siblings.
The proof is provided in \cref{app:proofs_markov_equivalence}.

\paragraph{Potential Parents.}
Unlike for potential siblings, we do not define \emph{potential parents} solely with respect to an independence model $\cI$ but with respect to a DMG $\cG$. This is due to the more complicated nature of directed edges determining ancestral relations and especially due to the concept of unblockable non-colliders. Nevertheless we show that any two Markov equivalent DMGs have the same set of potential parents.

\begin{definition}[Potential Parents]\label{def:potential_parents}
Let $\cG = (V, \cE_d, \cE_{bi})$ be a DMG, $v, w \in V$ and let $\cI_W^\cG$ be the independence model induced by $\cG$. We call $v$ a \defined{potential parent} of $w$ in $\cG$ if all of the following hold:
\begin{enumerate}[label=\alph*)]
    \item \label{cond:potential_parents_a} $v\in u (w, \cI_W^\cG)$ %
    
    \item \label{cond:potential_parents_b}$\forall u, x \in \anc^\cG(v) \cap \des^\cG(w)$:
    $u\in u (x, \cI_W^\cG)$

    \item \label{cond:potential_parents_c} $\forall a, b \in V$, $C,D \subseteq V$, with $a \indep_W^\cG b \mid C, D$ and $a \indep_W^\cG b \mid C, D \cup \{w\}: a \indep_W^\cG b \mid C, D \cup \{v\}$

    \item \label{cond:potential_parents_d} $\forall a, b \in V$, $C,D \subseteq V$, with $a \indep_W^\cG b \mid C, D$ and $a \indep_W^\cG b \mid C\cup \{w\}, D : a \indep_W^\cG b \mid C\cup \{v\}, D $
 
\end{enumerate}
\end{definition}
    Note that if $v \in \anc(w)$ and there exist an a-inducing walk from $v$ to $w$, then $v$ is a potential parent of $w$ since in this case conditions \cref{cond:potential_parents_a,cond:potential_parents_b,cond:potential_parents_c,cond:potential_parents_d} trivially hold. \Cref{lem:pa_are_pot_pa} is an important special case of this. 
\begin{lemma}\label{lem:pa_are_pot_pa}
    Let $\cG = (V, \cE_d, \cE_{bi})$ be a DMG, $v,w \in V$ and $(v, w) \in \cE_{d}$. Then $v$ is a potential parent of $w$ in $\cG$.
\end{lemma}

\begin{proposition}%
\label{prop:adding_directed_edges}
Let $\cG = (V, \cE_d, \cE_{bi})$ be a DMG, $v, w \in V$.
Let $(v,w) \notin \cE_d$ and denote $\cG' = (V, \cE_d', \cE_{bi})$, where $\cE_d':=\cE_d \cup \{(v,w)\}$.
Then $\cI_W^\cG = \cI_W^{\cG'}$ if and only if $v$ is a potential parent of $w$ in $\cI_W^\cG$.
\end{proposition}
In words: If $v$ is a potential parent of $w$, then adding $v\myrightarrow w$ to $\cG$ does not change the induced independence model. This motivates the term \textit{potential} parent and is proven in \cref{app:proofs_markov_equivalence}.
\begin{remark}
    Adding directed edges is slightly more involved than adding bidirected edges, as they are asymmetric and may alter ancestral relations. 
    Intuitively, when adding directed edges, conditions \cref{cond:potential_parents_a,cond:potential_parents_b,cond:potential_parents_c,cond:potential_parents_d} of the \cref{def:potential_parents} of potential parents can be understood as follows: a) demands that there already exists an a-inducing walk since $v\myrightarrow w$ is a-inducing; b) ensures that no separable pair of nodes becomes inseparable because adding a directed edge creates a new strongly connected component including nodes that have previously been separable; c) and d) ensure that for two nodes that are W-separated given $C$ and $D$, there exist no walk that is blocked only by colliders and is opened by altering the ancestral relations of those.
\end{remark}
\begin{proposition}\label{prop:pot_pa_two_graphs}
Let $\cG^1 = (V, \cE^1_d, \cE^1_{bi})$ and  $\cG^2 = (V,\cE^2_d , \cE^2_{bi})$ be two DMGs over a common node set $V$, $v,w \in V$ and $\cG^1 \in [\cG^2]_W$. If  $v$ is a potential parent of $w$ in $\cG^2$, %
then $v$ is a potential parent of $w$ in $\cG^1$. 
\end{proposition}

\cref{prop:pot_pa_two_graphs} demonstrates that, even though we defined potential parents through graphical properties of a DMG, the set of potential parents is the same for all graphs within a Markov equivalence class. 

\begin{figure*}[t]
    \centering
    \includegraphics[width=0.83\linewidth]
    {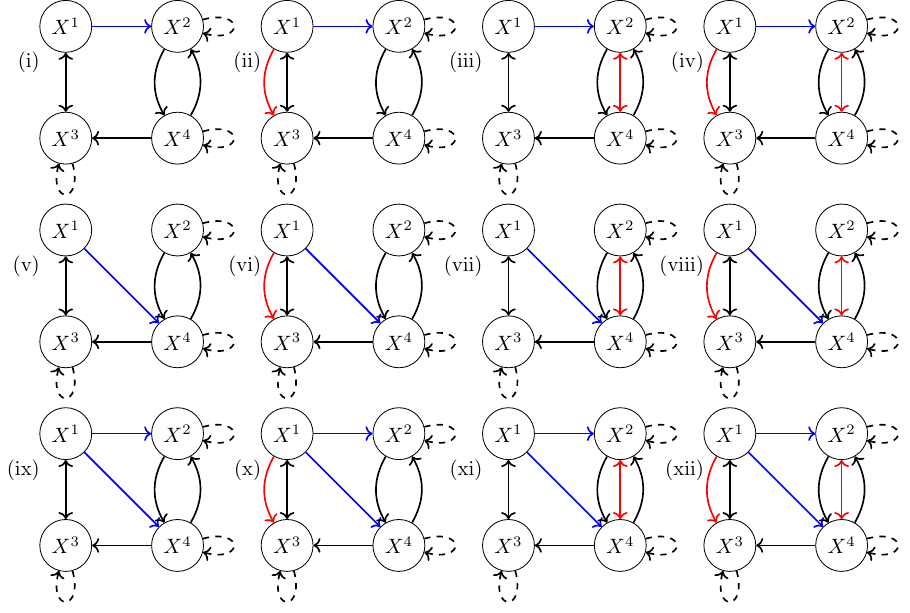}
    \caption{
    DMGs (i)-(xii) are the elements of one Markov equivalence class with (xii) being its greatest element. The dashed self-loops on $X^2,X^3$ and $X^4$ mark edges that may be removed without leaving the equivalence class. They are shown as dashed to avoid plotting all combinations.}\label{fig:dmg_equivalence_class}
\end{figure*}

\paragraph{Example:} \Cref{fig:dmg_equivalence_class} shows one Markov equivalence class and how potential parents and siblings permit edge additions that leave the independence model invariant.
    Adding $X^1 \myrightarrow X^3$ (red) is permitted: $X^1$ is a potential parent of $X^3$, since the a-inducing path $X^1 \bidiedge X^3$ already exists and $X^1 \in \anc(X^3)$, so no ancestral relation changes. The same holds for the dashed self-loops on $X^2,X^3,X^4$, whereas $X^1 \myrightarrow X^1$ is excluded as there is no a-inducing path from $X^1$ to $X^1$.
    In contrast, $X^3 \myrightarrow X^1$ cannot be added despite the a-inducing path from $X^3$ to $X^1$: it would merge all nodes into one strongly connected component, violating (b), and turn the collider path $X^4 \myrightarrow X^3 \bidiedge X^1$ into an a-inducing one, violating (c) and (d).
    As in \citet{manten2025}, some directed edge (blue) from $X^1$ into $\scc(X^2)=\{X^2,X^4\}$ is always present, in any variant.
    Finally, $X^2$ and $X^4$ are potential siblings, so $X^2 \bidiedge X^4$ (red) can be added: they are already connected by a-inducing walks in both directions, and bidirected edges do not alter ancestral relations. Adding or removing any other bidirected edge would change which pairs are connected by a-inducing walks.

We highlight two specific properties of potential siblings and parents that are particularly noteworthy.

\begin{proposition}
    \label{prop:a_ind_pot_sib}
    Let $\cG=(V, \cE_{d}, \cE_{bi})$ a DMG; $v, w \in V$. If $v \notin \anc(w)$ and there exists an a-inducing walk from $v$ to $w$ in $\cG$ %
    then $v$ and $w$ are potential sibling in $\cI_W^{\cG}$. 
\end{proposition}

\cref{lem:a_in_A} provides an ``inheritance'' rule for potential relations: it establishes that any ancestor of a potential parent $v$ that is not already an ancestor of the ``potential child'' $w$ must be a potential sibling of $w$, allowing us to propagate potential relations along ancestral lineages.

\begin{lemma}\label{lem:a_in_A}
Let $\cG = (V, \cE_d, \cE_{bi})$ be a DMG, $v,w \in V$ and $v \notin \anc(w)$.
If $v$ is a potential parent of $w$ in $\cG$, then for every $a \in V$ with $a \in \anc^\cG(v) \setminus \anc^\cG(w)$ it holds that $a$ is a potential sibling of $w$ in $\cI_W^{\cG}$.
\end{lemma}

The proof is given in \cref{app:proofs_markov_equivalence}. We provide a number of additional useful concepts and structural results in \cref{app:s_inducing_walks}.

\begin{theorem}[Greatest Markov equivalent DMG]\label{thr:greatest_dmg}
    For a DMG $\cG=(V,\cE_d ,\cE_{bi})$, its Markov equivalence class $[\cG]_W$ contains a greatest element. Hence there exist a $\hat{\cG}\in [\cG]_W$ that is a supergraph of all elements of $[\cG]_W$. 
    
    Furthermore, for a finite sequence of DMGs $\cG_0,G_1,...,G_m$ with $\cG_0=G$, $\cG_m=\hat{\cG}$ and $\cG_i=(V,\cE^i_d,\cE^i_{bi})$ such that $\cE^i_d \cup \cE^i_{bi} \subseteq \cE^{i+1}_d \cup \cE^{i+1}_{bi}$ for all $i\in \{0,1,...,m-1\}$. Then it holds that $\cG_i \in [{\cG}]_W$ for all $i\in \{0,1,...,m\}$.
\end{theorem}

The proof is provided in \cref{app:proofs_greatest}.

\begin{remark}
    From \cref{thr:greatest_dmg} it follows that the greatest element is the only maximal graph of its Markov equivalence class. Thus it provides a unique representative of the class and a target for structural learning algorithms. In other words, maximal DMGs are structurally identifiable from their W-independence models.
\end{remark}

\begin{corollary}\label{cor:greatest_char}
A DMG $\cG$ is the greatest element of $[\cG]_W$ if and only if every potential parent of a node is a parent of it and any two potential siblings are siblings in $\cG$.
\end{corollary}
The proof is provided in \cref{app:proofs_greatest}.

\section{Conclusion}
We introduced W-separation, an asymmetric graphical criterion for directed mixed graphs that refines and extends E-separation by enabling separate conditioning on past and future components. This added expressiveness resolves key characterization gaps: we establish the equivalence between graphical inseparability and inducing walks, characterize separability in the W-independence model, and fully characterize the corresponding Markov equivalence classes. In particular, every class admits a greatest element that serves as a canonical representative and principled learning target. By proving a global Markov property for a broad class of SDEs with latent confounding, we connect W-separation to conditional independence relations among coordinate processes that are testable from data. Together, these results provide a rigorous foundation for causal discovery in cyclic dynamical systems with latent confounding that leverages the direction of time.
Our contribution is foundational rather than algorithmic: a concrete discovery algorithm, e.g., adapting \citet{manten2025}, is left for future work and will need to address the combinatorial overhead of searching over pairs of separating sets $(C,D)$.

\subsubsection*{Acknowledgments}
This work has been supported by the German Federal Ministry of Education and Research (Grant: 01IS24082).
This work has received funding from the European Union (ERC, DYNAMICAUS, 101221985).
Funded by the Deutsche Forschungsgemeinschaft (DFG, German Research Foundation) -- GRK~3081/1 -- Project number 534429653.

\bibliography{bibliography}

\clearpage
\appendix
\thispagestyle{empty}

\onecolumn
\papertitle{Markov-Equivalent Conditional Independence Graphs on Path Spaces: \\
Supplementary Material}

\section{Proofs}\label{app:proofs}

\subsection{Proofs on Linking A-Inducing Walks and W-Separation}
\label{app:proof_aind_ssep}
We establish the following two lemmas that relate walks in a DMG $\cG$ to walks in its lifted dependency graph, before we proceed with the proof of \cref{prop:a_ind_eqv_s_sep}.

\begin{lemma}[Reduced Projection]
\label{lem:reduced_projection}
    Let $\cG = (V, \cE_d, \cE_{bi})$ a DMG, $\tilde{\cG} = (\tilde{V} = V_0 \,\dot\cup\, V_1, \tilde{\cE}_d, \tilde{\cE}_{bi})$ its lifted dependencies graph. Then
    for every walk $\tilde{\pi}: v_0 = v_{\ell_0}^0 \mysim[\tilde{e}_0] \cdots \mysim[\tilde{e}_n] v_{\ell_{n+1}}^{n+1}=w_{\ell_{n+1}}$ from $v_{\ell_0}$ to $v_{\ell_{n+1}}$ in $\tilde{\cG}$ there exists a walk $\pi: v = v^0 \mysim[e^0] \cdots \mysim[e^n] v^{n+1} = w$ from $v$ to $w$ in $\cG$ such that
    
    \begin{enumerate}[label=\roman*)]
        \item $\forall \, v^i \in \nodes_\pi$: $v_{\ell_i}^i \in \nodes_{\tilde{\pi}}$: $v_{\ell_i}^i \in \tilde{V}$ is the copy of $v^i \in V$ with subscript $\ell_i$
        
        \item $\forall \, e_i \in \edges_{\pi}, \tilde{e}_i \in \edges_{\tilde{\pi}}: \begin{cases} \tilde{e}_i \in (u_{\ell_u}, k_{\ell_k}) \\ \tilde{e}_i \in [(u_{\ell_u}, k_{\ell_k})] \end{cases} \Rightarrow \begin{cases} e_i \in (u, k) \\ e_i \in [(u, k)] \end{cases}$
        
    \end{enumerate}
    where $u_{\ell_i}, u_{\ell_j}$ is the copy of $u, u$ with subscript $\ell_i, \ell_j$ respectively.
    We call such a walk $\pi$ the \defined{reduced projection of the walk $\tilde{\pi}$}.
\end{lemma}
\begin{proof}[Proof of \cref{lem:reduced_projection}]
    i), ii) hold since by construction of $\tilde{\cG}$, all $v_{\ell_i}^i \in \tilde{V}$ and $\tilde{e}_i \in \tilde{\cE}_d \cup \tilde{\cE}_{bi}$ exist only if $v^i \in V$ and $e_i \in \cE_d \cup \cE_{bi}$. 
\end{proof}

\begin{lemma} \label{lem:properties_reduced_projection} 
Let $\cG = (V, \cE_d, \cE_{bi})$ a DMG, $\tilde{\cG} = (\tilde{V} = V_0 \cup V_1, \tilde{\cE}_d, \tilde{\cE}_{bi})$ its lifted dependencies graph, $\tilde{\pi}: v_{\ell_0} = v_{\ell_0}^0 \mysim[\tilde{e}_0] \cdots \mysim[\tilde{e}_n] v_{\ell_{n+1}}^{n+1}=w_{\ell_{n+1}}$ be a walk from $v_{\ell_0}$ to $w_{\ell_{n+1}}$ in $\tilde{\cG}$ and let the walk $\pi: v = v^0 \mysim[e^0] \cdots \mysim[e^n] v^{n+1} = w$ in $\cG$ be the reduced projection of $\tilde{\pi}$.
Then all of the following hold:

\begin{enumerate}[label=\roman*)]
    \item $(v^i \in \coll_{\pi} \Leftrightarrow v_{\ell_i}^i \in \coll_{\tilde{\pi}}) \wedge (v^i \in \ncoll_{\pi} \Leftrightarrow v_{\ell_i}^i \in \ncoll_{\tilde{\pi}})$
    
    \item $\forall v_{\ell_i}^i, v_{\ell_k}^k \in \nodes_{\tilde{\pi}}: v_{\ell_i}^i \in \scc^{\tilde{\cG}}(v_{\ell_k}^k) \Rightarrow \ell_i = \ell_k$
    
    \item $\forall v_{\ell_i}^i, v_{\ell_k}^k \in \nodes_{\tilde{\pi}}, v^u \in \nodes_{\pi}: v_{\ell_i}^i \in \scc^{\tilde{\cG}}(v_{\ell_k}^k) \Rightarrow v^i \in \scc^{\cG}(v^k)$ \\
    (note $\ell_i = \ell_k = \ell$)%
    
    \item $v_{\ell_i}^i \in \ncoll_{\tilde{\pi}}$ is unblockable $\Rightarrow v^i \in \ncoll_{\pi}$ is unblockable  
    
    \item $v_{\ell_i}^i \in \anc(v_{\ell_j}^j) \Rightarrow v^i \in \anc(v^j)$\\
\end{enumerate}
\end{lemma}
\begin{proof}[Proof of \cref{lem:properties_reduced_projection}]\hfill

\begin{enumerate}[label=\roman*)]
    \item holds trivially by construction of $\pi$ as a reduced projection of $\tilde{\pi}$.
    
    \item assume $\ell_i \neq \ell_k$ and w.l.o.g $\ell_i = 1$, $\ell_k = 0$
    
    it follows by construction of $\tilde{\cG}$ that there exists no directed path from $v_1^i$ to $v_0^k$ and thus $v_1^i \notin \scc(v_0^k)$.
    
    \item $v_{\ell_i}^i \in \scc^{\tilde{\cG}}(v_{\ell_k}^k) \Leftrightarrow$ There exist directed paths $v_{\ell_i}^i \myrightarrow \cdots \myrightarrow v_{\ell_k}^k$ and $v_{\ell_k}^k \myrightarrow \cdots \myrightarrow v_{\ell_i}^i$ in $\tilde{\cG}$.
    By construction of $\tilde{\cG}$, these only exist if there exist the directed paths $v^k \myrightarrow \cdots v^i$ and $v^i \myrightarrow \cdots \myrightarrow v^k$ in $\cG$ which means $v^i \in \scc^\cG(v^k)$.
    
    \item For $v_{\ell_i}^i \in \ncoll_{\tilde{\pi}}$ unblockable: it has only outgoing edges on $\tilde{\pi}$ to nodes in the same strongly connected component of $\tilde{\cG}$ and it is not an end node.
    
    Then for every $ v_{\ell_j}^j \in \nodes_{\tilde{\pi}}$ such that $v_{\ell_i}^i \myrightarrow v_{\ell_j}^j$ %
    is on $\tilde{\pi}$, it holds that $v_{\ell_i}^i \in \scc^{\tilde{\cG}}(v_{\ell_j}^j)$ and with iii) it follows $v^i \in \scc^\cG(v^j)$ for every $v_j$ such that $ v^i \myrightarrow v^j$ %
    and thus $v^i \in \ncoll_{\pi}$ it unblockable.
    
    \item If $v_{\ell_i}^i \in \anc(v_{\ell_j}^j)$ then there exists a directed path $v_{\ell_i}^i \myrightarrow \cdots \myrightarrow v_{\ell_j}^j$. By construction of $\tilde{\cG}$ it follows that there exists a directed path $v^i \myrightarrow \cdots \myrightarrow v^j$ and hence $v^i \in \anc(v^j)$.

\end{enumerate}
\end{proof}

In the proof of \cref{prop:a_ind_eqv_s_sep} we additionally show that i)--iii) are equivalent to
\begin{enumerate}[label=\roman*), start=4]
    \item $(\{v\}, \{w\}, C, D) \notin \cI_W^\cG $ for
    $H := \anc(v,w) \setminus \{v\}$, $Y:= H_0 \cup H_1 \setminus \{w_1\}$

    $C := \{v^i \in H: \exists$ a $\sigma$-$Y$-blocked walk ${\pi}$ in $\tilde{\cG}$ from $v_0$ to $w_1$ where $v_0^i \in \ncoll_{{\pi},\mathrm{bl}}\}$

    $D := \{ v^i \in \anc(w) \setminus \{w \}: \exists$ a $\sigma$-$Y$-blocked walk ${\pi}$ in $\tilde{\cG}$ from $v_0$ to $w_1$ where $v_1^i \in \ncoll_{{\pi},\mathrm{bl}}\}$,
\end{enumerate}
i.e., the explicit pair $(C,D)$ separates $w$ from $v$ whenever any pair does.

\begin{proof}[Proof of \cref{prop:a_ind_eqv_s_sep}]\hfill
\begin{description}[leftmargin=0pt, labelindent=0pt]
    \item[$\mathbf{i) \implies ii)}$]
    Let $\pi: v \mysim[e_0] v^1 \mysim[e_1] \cdots \mysim[e_n] w$ be an a-inducing walk from $v$ to $w$ in $\cG$. If there exist $v^i, v^j \in \nodes_{\pi}$ such that $i \neq j$, $v^i = v^j$, we drop the subwalk from $v^i$ to $v^j$ on $\pi$ and obtain the walk $\pi': v \mysim[e_0] v^1 \mysim[e_1] \cdots \mysim[e_{i-1}] v^i = v^j \mysim[e_{j+1}] \cdots \mysim[e_n] w$ in $\cG$.

    Since $\pi$ is a-inducing it holds that $\coll_{\pi} \subseteq \anc(v, w)$ and for every $v^k \in \ncoll_{\pi}\setminus \{v^0,v^{n+1}\}: v^k$ is unblockable.
    
    If $v^i \in \coll_{\pi'}$, then it must be that either $\{v^i, v^j\} \cap \coll_{\pi} \neq \emptyset$ and hence $v^i \in \anc(v, w)$, or $v^i, v^j \in \ncoll_{\pi}$ are unblockable, $e_{i-1}$ is into $v^i$ and $e_j$ is into $v^j$. In the latter it then follows that $e_i \in \edges_{\pi}$ is out of $v^i$ and $e_{j-1} \in \edges_{\pi}$ is out of $v^j$ and hence there must exist a $v^h \in \coll_{\pi} \subseteq \anc(v, w)$ with $v^i \in \anc(v^h)$ and hence $v^i \in \anc(v, w)$.
    
    If $v^i \in \ncoll_{\pi'}$ then $v^i$ is unblockable since then $e_{i-1} \in \edges_{\pi}$ is out of $v^i$ or $e_j \in \edges_{\pi}$ is out of $v^j$ and $v^{i-1} \in \scc(v^i)$ or $v^{j+1} \in \scc(v^j)$, respectively.
    
    Hence $\pi'$ is an a-inducing walk. By iteratively applying this step to all $v^i, v^j \in \nodes_{\pi}$ with $i \neq j$, $v^i = v^j$ we obtain an a-inducing path.

    \item[$\mathbf{ii) \implies iii)}$] This step is analogously to that of Proposition 3.19 in \cite{manten2025} with only minor modifications to adapt it to our \cref{def:s_sep} of W-separation.
    Let us consider the following cases:
    \begin{enumerate}[label=\arabic*)]
        \item $v \in \scc^\cG(w)$
        
        Then there exists a directed path $\pi: v \myrightarrow v^1 \myrightarrow \cdots \myrightarrow v^n \myrightarrow w$
        from $v$ to $w$ in $\cG$ that lies entirely in $\scc(w)$ %
        and we can construct the directed path $\tilde{\pi}: v_0 \myrightarrow v_1^1 \myrightarrow \cdots \myrightarrow v_1^n \myrightarrow w_1$
        where $\forall v_{\ell_i}^i \in \nodes_{\tilde{\pi}}$ with $i > 0:\ell_i = 1 \wedge v_{\ell_i}^i \in \scc(w_1)$.
        
        Then it follows that $ \coll_{\tilde{\pi}} = \emptyset$ and $\ncoll_{\tilde{\pi},\mathrm{bl}} = \{v_0, w_1\}$ and hence
        $ \tilde{\pi}$ is $\sigma$-${Z}$-open for every $ {Z} = C_0 \cup D_1, $ with $ C \subseteq V\setminus \{v\}, D \subseteq V\setminus \{w\}$
    
        \item $v \notin \scc^\cG(w)$
    
        Then let $\pi: v = v_0 \mysim[e_0] \cdots \mysim[e_n] v^{n+1} = w$ be an a-ind. path in $\cG$ from $v$ to $w$
        
        Let $k := \max\{k \in [n]: v^k \notin \scc^\cG(w)\}$
        
        Then there exists an a-ind. path in $\cG$ from $v$ to $w$ that is into $w$.
        We can construct such path $\mu$ by concatenating the walk in $\pi$ until node $v^{k+1}$ with the directed path $v^{k+1} \myrightarrow \cdots \myrightarrow w$ that lies entirely in $\scc(w)$ %
        to obtain $ \mu: v \mysim[e_0] \cdots \mysim[e_{k-1}] v^k \circright[e_k] v^{k+1} \myrightarrow \cdots \myrightarrow w$.
        
        We construct the path $\tilde{\mu}$ in $\tilde{\cG}$ between $v_0$ and $w_1$ as:
        
        $\tilde{\mu}: v_0 \mysim[\tilde{e}_0] \cdots \mysim[\tilde{e}_{k-1}] v_{0}^k \circright[\tilde{e}_k] v_{1}^{k+1} \myrightarrow \cdots \myrightarrow w_1$ where $\begin{cases} \forall v_{\ell_i}^i\in \nodes_{\tilde{\pi}}, i \leq k: \ell_i = 0 \\ \forall v_{\ell_i}^i \in \nodes_{\tilde{\pi}}, i > k: \ell_i = 1 \end{cases}$.\
        We find that
        \begin{enumerate}[label=\alph*)]
            \item $\coll_{\tilde{\mu}} \subseteq \anc(v_0, w_1)$,
            
            since by construction it holds that $\coll_{\tilde{\mu}} \subseteq V_0$ and $\forall v_0^i \in \coll_{\tilde{\mu}}: v^i \in \coll_{\pi} \subseteq \anc(v, w)$.
            
            \item $\ncoll_{\tilde{\mu},\mathrm{bl}} \subseteq \{v_0, w_1\}$,
            
            since for every $ v_{\ell_i}^i \in \ncoll_{\tilde{\mu}} \setminus \{v_0, w_1\}$ it holds $v_{\ell_i}^i$ is unblockable due to the construction of $\tilde{\mu}$ from $\mu$.
            For $i \neq k$ this is obvious since $\mu$ is a-inducing.
            Note that $v_0^k \in \ncoll_{\tilde{\mu},\mathrm{bl}}$ only if $v_0^k = v_0$, since by definition of $k$ we have that $v^k \notin \scc(v^{k+1})$ or $v^k = v$, but $\ncoll_{\mu,\mathrm{bl}} = \{v, w\}$.
        \end{enumerate}
        
        Hence for every $Z:=C_0 \cup D_1$ with $C \subseteq V\setminus \{v\}, D \subseteq V\setminus \{w\}$ we know that $\ncoll_{\tilde{\pi},\mathrm{bl}} \cap Z = \emptyset $ and $\coll_{\tilde{\mu}} \subseteq \anc^{\tilde{\cG}}(\{v_0, w_1\} \cup Z)$.
    
        We want to show that $\coll_{\tilde{\mu}} \subseteq \anc^{\tilde{\cG}}(Z)$ meaning that in this case we have found a $\sigma$-$Z$-open walk from $v_0$ to $w_1$ in $\tilde{\cG}$ and hence $(\{v\}, \{w\}, C, D) \notin \cI_W $.
        
        Remember $\coll_{\tilde{\mu}} \subseteq V_0$.
        If there exists a $v_0^i \in \coll_{\tilde{\mu}} $ with $ v_0^i \in \anc^{\tilde{\cG}}(v_0) \setminus \anc^{\tilde{\cG}}(Z)$ (we assume $i$ to be minimal with this property), then it follows that there exists a directed path $v_0^i \myrightarrow \cdots \myrightarrow v_0$ that does not intersect $Z$
        and we can replace the subwalk from $v_0$ to $v^i_0$ on $\tilde{\mu}$ to construct the $\sigma$-$Z$-open path $\tilde{\mu}' : v_0 \myleftarrow \cdots \myleftarrow u_0^i \mysim[\tilde{e}_i] \cdots \mysim w_1$.
        
        We proceed analogously for a collider $v_0^i \in \anc^{\tilde{\cG}}(w_1) \setminus \anc^{\tilde{\cG}}(Z)$ (assuming $i$ minimal with this property).
        
    \end{enumerate}
    
    \item[$\mathbf{iii) \implies iv)}$] is trivial.
    
    \item[$\mathbf{iv) \implies i)}$]
    Then we have $H := \anc(v,w) \setminus \{v\}$, $Y:= H_0 \cup H_1 \setminus \{w_1\}$
        
    $C := \{v^i \in H: \exists$ a $\sigma$-$Y$-blocked walk ${\pi}$ in $\tilde{\cG}$ from $v_0$ to $w_1$ where $v_0^i \in \ncoll_{{\pi},\mathrm{bl}}\}$   
    
    $D := \{ v^i \in \anc(w) \setminus \{w \}: \exists$ a $\sigma$-$Y$-blocked walk ${\pi}$ in $\tilde{\cG}$ from $v_0$ to $w_1$ where $v_1^i \in \ncoll_{{\pi},\mathrm{bl}}\}$.

    We are going to show that every walk $\pi: v_0 \mysim[e_0] v_{\ell_1}^1 \mysim[e_1] \cdots \mysim v_{\ell_n}^n \mysim[e_n] w_1$ in $\tilde{\cG}$ from $v_0$ to $w_1$ is either $\sigma$-${Z}$-blocked or we can use ${\pi}$ to construct an a-inducing walk $\mu: v \mysim[\hat{e}_0] v^1 \mysim[\hat{e}_1] \cdots \mysim[\hat{e}_n] w$  from $v$ to $w$ in $\cG$.

    Since $(\{v\}, \{w\}, C, D) \notin \cI_W$ there must exist a $\pi$ that is $\sigma$-$Z$-open. Let ${\pi}$ be $\sigma$-${Z}$-open.
    
    We are going to look on subwalks of $\pi$ that lie entirely in $V_0$ or $V_1$. Through this we are going to characterize all $\coll_{\pi}$ and all $\ncoll_{\pi}$ (hence all nodes on $\pi$). We will then use the found characterization to show that it exist an a-inducing walk $\mu$ in $\cG$.
    
    \underline{Subwalks on $\pi$ in the future}
    
    There exist the nodes $v_0^j \circright v_1^{j+1}$ on $\pi$ such that $\forall v_{\ell_i}^i \in \nodes_{\pi}$ with $i > j: \ell_i = 1$.
    
    We define $\pi_1: v_1^{j+1} \mysim[e_{j+1}] \cdots \mysim[e_n] w_1$ as the subwalk from $v_1^{j+1}$ to $w_1$ on $\pi$.
    We further divide $\pi_1$ in $m$ subwalks $\pi_1^s$ with $s \in \{1, \ldots, m\}$. 
    
    We define $\pi_1^s: v_1^{l_s} \mysim[e_{\ell_s}] \cdots \mysim[e_{r_{s-1}}] v_1^{r_s}$ as a subwalk from $v_1^{l_s}$ to $v_1^{r_s}$ on $\pi_1$ such that it holds:

    \begin{enumerate}[label=\alph*)]
        \item $v_1^{l_1} = v_1^{j+1}$ \quad $\wedge$ \quad $v_1^{r_m} = w_1$
        
        \item $v_1^{r_i} = v_1^{l_{i+1}}$ \quad $\forall i \in \{1, \ldots, m-1\}$
        
        \item $\pi_1^s$ has maximal length w.r.t. being one of the following structures:
        
        \begin{enumerate}[label=\roman*), leftmargin=*]
            \item $\pi_1^s: v_1^{l_s} \myleftarrow \cdots \myleftarrow v_1^{r_s}$ is a left chain directed walk %
            
            \item $\pi_1^s: v_1^{l_s} \myrightarrow \cdots \myrightarrow v_1^{r_s}$ is a right chain directed walk %
            
            \item $\pi_1^s: v_1^{l_s} \bidiedge v_1^{r_s}$
            
            \item $\pi_1^s: w_1$ is the trivial walk \quad (only if $v_1^{j+1} = w_1$)
            
        \end{enumerate}
        
    \end{enumerate}
    
    Note that every $\pi_1$ can be divided into such subwalks $\pi_1^s$.

    For the following case distinction we are going to assume that for every $ s \in \{1, \ldots, m\}$ it holds that $v_1^{r_s} \in \anc(w_1)$.
    For $s=m$ this trivially holds since $v_1^{r_m} = w_1$. We are going to show that from $v_1^{r_s} \in \anc(w_1)$ it follows $v_1^{l_s} \in \anc(w_1)$ for every $\pi_1^s$. Since $v_1^{r_i} = v_1^{l_{i+1}}$ for$ i \in \{1, \ldots, m-1\}$, from the assumption holding for $s=m$ it then follows for all $ s \in \{1, \ldots, m\}$.

    Note that by definition $C \subseteq \anc(v,w) $, $ D \subseteq \anc(w)$ and hence $C_0 \subseteq \anc(v_1,w_1)$, $ D_1 \subseteq \anc(w_1)$.
    
    We consider the following cases:

    \begin{enumerate}[label=\arabic*), ref=\arabic*] 
    
        \item \label{case:Prop_eqv:pi_1:1} $\pi_1^s: v_1^{l_s} \myleftarrow \cdots \myleftarrow v_1^{r_s}$ is a left chain directed walk
        
        Then $v_1^{l_s} \in \coll_{\pi}$ (otherwise $\pi_1^s$ would not be of maximal length).
        \begin{enumerate}[label=\theenumi.\arabic*), ref=\theenumi.\arabic* ]
            
            \item $v_1^{l_s} \notin \anc(w_1) \Rightarrow v_1^{l_s} \notin \anc(D_1) \Rightarrow v_1^{l_s} \notin \anc(Z)$
            
            $\Rightarrow \pi$ is $\sigma$-$Z$-blocked, a contradiction. assumption
            
            \item $v_1^{l_s} \in \anc(w_1) \Rightarrow \forall v_1^i \in \nodes_{\pi_1^s}: v_1^i \in \anc(w_1)$
        
            Therefore $\forall v_1^i \in \nodes_{\pi_1^s} \setminus \{v_1^{l_s}\}$ one of the following holds:
            
            \begin{enumerate}[label=\theenumii.\arabic*)]
                \item $v_1^i \in \ncoll_{\pi}$ is unblockable or $ v_1^i = w_1$
                
                \item $v_1^i \in \ncoll_{\pi,\mathrm{bl}}$ and then by definition of $D$ it holds $v_1^i \in D_1$
                
                $\Rightarrow \pi$ is $\sigma$-$Z$-blocked, a contradiction. assumption
                
            \end{enumerate}
            
        \end{enumerate}

        \item \label{case:Prop_eqv:pi_1:2} $\pi_1^s: v_1^{l_s} \myrightarrow \cdots \myrightarrow v_1^{r_s}$ is a right chain directed walk
        
        Then $v_1^{r_s} \in \coll_{\pi} \cup \{w_1\}$.
        From $v_1^{r_s} \in \anc(w_1) $ it follows that for every $ v_1^i \in \nodes_{\pi_1^s} $ it holds $ v_1^i \in \anc(w_1)$ and either
        \begin{enumerate}[label=\theenumi.\arabic*), ref=\theenumi.\arabic* ]
            \item $v_1^i \in \ncoll_{\pi}$ is unblockable or $ v_1^i = v_1^{r_s} \in \coll_{\pi} \cup \{w_1\} $
                    
            \item $v_1^i \in \ncoll_{\pi,\mathrm{bl}}$ and then by definition of $D$ it holds $v_1^i \in D_1$
            
            $\Rightarrow \pi$ is $\sigma$-$Z$-blocked, a contradiction. assumption 
        \end{enumerate}

        \item \label{case:Prop_eqv:pi_1:3} $\pi_1^s: v_1^{l_s} \bidiedge v_1^{r_s}$
        
        We distinguish between:
        \begin{enumerate}[label=\theenumi.\arabic*), ref=\theenumi.\arabic* ]
            \item $v_1^{l_s} \notin \anc(w_1) \Rightarrow v_1^{l_s} \notin D_1$
            
            \begin{enumerate}[label=\theenumii.\arabic*), ref=\theenumii.\arabic* ]
                \item $v_1^{l_s} \in \coll_{\pi} $
                
                $ \Rightarrow \pi$ is $\sigma$-$Z$-blocked, a contradiction. assumption
                
                \item $v_1^{l_s} \in \ncoll_{\pi}$
                
                Then $e_{l_s-1}\in \edges_\pi$ is out of $v_1^{l_s}$. Thus there need to be a $v_1^k \in \coll_{\pi}$ 
                with $v_1^k \in \des(v_1^{l_s}) $ and hence $ v_1^k \notin \anc(w_1)$. Therefore it follows $ v_1^{l_s} \notin \anc(D_1) $ and hence $ v_1^{l_s} \in \coll_\pi \setminus \anc(Z)$.
                
                $\Rightarrow \pi$ is $\sigma$-$Z$-blocked, a contradiction. assumption
            \end{enumerate}
            
            \item $v_1^{l_s} \in \anc(w_1)$
            
            Then for both $v_1^{l_s}$ and $v_1^{r_s}$ one of the following holds:
            
            \begin{enumerate}[label=\theenumii.\arabic*), ref=\theenumii.\arabic* ]
                \item $v_1^{l_i}, v_1^{r_s} \in \coll_{\pi} \cap \anc(w_1)$
                
                \item $v_1^{l_i}, v_1^{r_s} \in \ncoll_{\pi}$ is unblockable or $v_1^{r_s} = w_1$
                \item $v_1^{l_i}, v_1^{r_s} \in \ncoll_{\pi,\mathrm{bl}} $ and then by definition of $D$ it holds $ v_1^{l_i}, v_1^{r_s} \in D_1$ 
                
                $\Rightarrow \pi$ is $\sigma$-$Z$-blocked, a contradiction. assumption
            \end{enumerate}

        \end{enumerate}

        \item \label{case:Prop_eqv:pi_1:4} $\pi_1^s: w_1$ is the trivial walk\\
        
    \end{enumerate}

With cases \ref{case:Prop_eqv:pi_1:1}-\ref{case:Prop_eqv:pi_1:4} we have shown that for every $ s \in \{1, \ldots, m\}$ it holds $ v_1^{l_s} \in \anc(w_1)$ and hence we confirmed our assumption that for every $ s \in \{1, \ldots, m\}$ it holds that $v_1^{r_s} \in \anc(w_1)$.
Furthermore we have shown that for every $ v_1^i \in \nodes_{\pi_1}$ one of the following holds:

\begin{itemize}

    \item $v_1^i \in \coll_{\pi} \cap \anc(w_1)$
    
    \item $v_1^i \in \ncoll_{\pi}$ is unblockable
    
    \item $v_1^i = w_1$\\
\end{itemize}

We are now going to cover nodes $v_1^i \in \nodes_{\pi} $ for which $ i \leq j$.

We therefore define $\pi_f: v_i^p \mysim[e_{\ell_0}] \cdots \mysim[e_q] v_i^q$ to be a subwalk on $\pi$ such that it holds:

\begin{enumerate}[label=\alph*)]
    \item $\forall v_{\ell_i}^i \in \nodes_{\pi_f}: \ell_i = 1$
    
    \item $v_{\ell_{p-1}}^{p-1}, v_{\ell_{q+1}}^{q+1} \in V_0$
\end{enumerate}

Then we have that $v_1^q \in \anc(w_1)$, since otherwise there would exist a $ v_1^k \in \coll_{\pi}$ with $v_1^k \notin \anc(w_1) \supseteq \anc(D_1) $ and hence $ v_1^k \notin \anc(Z)$ meaning $\pi$ is $\sigma$-$Z$-blocked, a contradiction to our assumption.

We further divide $\pi_f$ in $m$ subwalks $\pi_f^s$ with $s \in \{1, \ldots, \hat{m}\}$.
We define $\pi_f^s: v_1^{l_s} \mysim[e_{\ell_s}] \cdots \mysim[e_{r_s-1}] v_1^{r_s}$ as a subwalk from $v_1^{l_s}$ to $v_1^{r_s}$ on $\pi_f$ such that it holds:

\begin{enumerate}[label=\roman*)]
    \item $v_1^{l_1} = v_1^p$ $\wedge$ $v_1^{r_{\hat{m}}} = v_1^q$
    
    \item $v_1^{r_i} = v_1^{l_{i+1}}$ $\forall i \in \{1, \ldots, \hat{m}-1\}$
    
    \item $\pi_f^s$ has maximal length w.r.t. being one of the following structures:
    
    \begin{enumerate}[label=\alph*)]
        \item $\pi_f^s: v_1^{l_s} \myleftarrow \cdots \myleftarrow v_1^{r_s}$ is a left chain directed walk %
        
        \item $\pi_f^s: v_1^{l_s} \myrightarrow \cdots \myrightarrow v_1^{r_s}$ is a right chain directed walk %
        
        \item $\pi_f^s: v_1^{l_s} \bidiedge v_1^{r_s}$
        
        \item $\pi_f^s: v_1^q$ is the trivial walk \quad (only if $v_1^p = v_1^q$)
    \end{enumerate}
\end{enumerate}

Note that every $\pi_f$ can be divided into such subwalks $\pi_f^s$.

If $\pi_f: v_1^q$ it follows that $ v_1^q \in \coll_{\pi} \cap \anc(w_1)$.

Otherwise we can use cases \ref{case:Prop_eqv:pi_1:1}-\ref{case:Prop_eqv:pi_1:3} (from our case distinction on $\pi_1^s$) and
to show that for every
$ v_1^i \in \nodes_{\pi_f}$ one of the following holds:

\begin{itemize}
    \item $v_1^i \in \coll_{\pi} \cap \anc(w_1)$
    
    \item $v_1^i \in \ncoll_{\pi}$ is unblockable
    
\end{itemize}

Note that $w_1 \notin \nodes_{\pi_f}$.

Note that $\nodes_{\pi_1} \cup \nodes_{\pi_f} = \nodes_\pi \cap V_1$. Therefore we showed that for every $ v_{\ell_i}^i \in \nodes_{\pi}$ with $\ell_i = 1$ one of the following holds:

\begin{itemize}
    \item $v_1^i \in \coll_{\pi} \cap \anc(w_1)$
    
    \item $v_1^i \in \ncoll_{\pi}$ is unblockable
    
    \item $v_1^i = w_1$\\
    
\end{itemize}

\underline{Now consider subwalks on $\pi$ in the past}

Note that by the definition of $C,D$ we have $Z = C_0 \cup D_1 \subseteq \anc(v_0, w_1) \setminus \{v_0, w_1\}$

Then for every $ v_0^i \in \nodes_{\pi}$ one of the following holds:
\begin{enumerate}[label=\Alph*), ref=\Alph*]
    \item $v_0^i \in \coll_{\pi}$. Then either

    \begin{enumerate}[label=\theenumi.\arabic*)]
        \item $v_0^i \in \anc(Z) \quad \Rightarrow \quad v_0^i \in \coll_\pi \cap \anc(v_0, w_1)$
        
        \item $v_0^i \notin \anc(Z) \quad \Rightarrow \quad \pi$ is $\sigma$-$Z$-blocked, a contradiction.
    \end{enumerate}

    \item $v_0^i \in \ncoll_{\pi}$. Then either

    \begin{enumerate}[label=\theenumi.\arabic*), ref=\theenumi.\arabic*]
        \item $v_0^i$ is unblockable or $v_0^i = v_0$
        
        \item $v_0^i \in \ncoll_{\pi,\mathrm{bl}} \setminus \{v_0\}$. Then either
    
        \begin{enumerate}[label=\theenumii.\arabic*)]
            \item $v^i \in \anc(v, w) $ and then by definition of $C$ it holds $ v_0^i \in C_0 \subseteq Z $
            
            $\Rightarrow \pi$ is $\sigma$-$Z$-blocked, a contradiction. assumption
            
            \item $v^i \notin \anc(v, w) $
            
            Then $v_0^i \in \ncoll_{\pi} \setminus \anc(v_0, w_1)$ and hence there must exist a $v_{\ell_k}^k \in \coll_{\pi} \setminus \anc(v_0, w_1) $ and thus by definition of $C$ and $D$ it follows $ v_{\ell_k}^k \notin \anc(Z)$ 
            
            $\Rightarrow \pi$ is $\sigma$-$Z$-blocked, a contradiction.
            
        \end{enumerate}
    
    \end{enumerate}

\end{enumerate}

Hence we just showed that for every $ v_0^i \in \nodes_{\pi}$ one of the following holds

\begin{itemize}
    \item $v_i^i \in \coll_{\pi} \cap \anc(v_0, w_1)$
    
    \item $v_0^i \in \ncoll_{\pi}$ is unblockable
    
    \item $v_0^i = v_0$
\end{itemize}

Overall we showed that there exists a ($\sigma$-$Z$-open) walk $\pi: v_0 \mysim[e_0] \cdots \mysim[e_n] w_1$ in $\tilde{\cG}$ from $v_0$ to $w_1$ for which it holds:

\begin{itemize}
    \item $\coll_{\pi} \subseteq \anc(v_0, w_1)$
    
    \item $\forall v_{\ell_i}^i \in \ncoll_{\pi} \setminus \{v_0, w_1\}$: $v_{\ell_i}^i$ is unblockable\\
    
\end{itemize}

Let us do some considerations about the \underline{last edge on $\pi$ before entering $\scc(w)$}

Let us define $s := \min\{s \in [n] \text{ such that }\forall v_{\ell_i}^i \in \nodes_\pi \text{ with } s<i: v^i \in \scc(w)\}$.

If $v \in \scc(w)$ then there exists an a-ind. path from v to w in $\cG$ and we are done.

If $v \notin \scc(w)$ then for $e_s \in \edges_\pi \setminus \{e_n\}$ it holds that $e_s$ is into $v_{\ell_{s+1}}^{s+1} \in \nodes_\pi$, as otherwise $v_{\ell_{s+1}}^{s+1} \in \ncoll_{\pi,\mathrm{bl}} \setminus \{v_0, w_1\}$ a contradiction since we showed $\ncoll_{\pi,\mathrm{bl}} = \{v_0, w_1\}$.

For $e_s = e_n$ it follows that $e_s$ is into $v_{\ell_{s+1}}^{s+1} = w_1$. To see this assume the contrary. Then $e_s \in \pi_1^m$ a left chain directed walk and hence $v_1^s \in \des(w_1)$. As we have shown in case \ref{case:Prop_eqv:pi_1:1} when considering subwalks on $\pi_1$, it then follows that $v_1^s \in \anc(w_1)$ and hence $v_1^s \in \scc(w_1)$, a contradiction to how we defined $s$.

Therefore we showed that $e_s \in \edges_\pi$ is into $v_{\ell_{s+1}}^{s+1}$.\\

We are now going to show that we can \underline{construct an a-inducing walk from $\pi$.}

Let $\mu : v \mysim[\hat{e}_0] \cdots \mysim[\hat{e}_n] w$
be the \textit{reduced projection of $\pi$} and thus a walk from $v$ to $w$ in $\cG$.

By \cref{lem:reduced_projection,lem:properties_reduced_projection}, we know that such $\mu$ exist and that the properties of $\pi$:

\begin{itemize}
    \item $\coll_{\pi} \subseteq \anc(v_0, w_1)$
    \item $\forall v_{\ell_i}^i \in \ncoll_{\pi} \setminus \{v_0, w_1\}: v_{\ell_i}^i$ is unblockable
    \item $e_s$ is into $v_{\ell_{s+1}}^{s+1}$
\end{itemize}

transfer to $\mu$ such that it holds:

\begin{itemize}
    \item $\coll_{\mu} \subseteq \anc(v, w)$
    \item $\forall v^i \in \ncoll_{\mu} \setminus \{v, w\}: v^i$ is unblockable
    \item $\hat{e}_s$ is into $v^{s+1}$
    
\end{itemize}

Since we defined ${e}_s$ such that $\hat{e}_s$ is the last edge on $\mu$ before entering $\scc(w)$, we find that the last edge on $\mu$ before entering $\scc(w)$ is into $\scc(w)$.

Hence $\mu$ is an a-inducing walk.

\end{description}
\end{proof}

\subsection{Properties of Inducing Walks and Additional Structural Results.}
\label{app:s_inducing_walks}
In the following we provide a number of additional useful concepts and structural results.
We introduce the notion of \emph{s-inducing walks}, which provide %
a versatile tool, leveraged across multiple proofs (e.g. of \cref{prop:pot_sib_in_G}) due to their ability to represent the effects of bidirected edges through path-based constructions.

\begin{definition}[S-inducing walks] \label{def:s-ind-walks}
Let $\cG = (V, \cE_d, \cE_{bi})$ be a DMG, $v, w \in V$, $v \notin \scc(w)$.
An s-inducing walk from $v$ to $w$ in $\cG$ is a non-trivial path
$\pi: v \mysim[e_0] v^1 \mysim[e_1] \cdots \mysim[e_n] w$ such that $\coll_\pi \subseteq \anc(\{v,w\})$, $\ncoll_{\pi,\mathrm{bl}}\setminus \en_\pi = \emptyset$,
the last edge before entering $\scc(w)$ is into $\scc(w)$ and
the first edge after leaving $\scc(v)$ is into $\scc(v)$.
\end{definition}
Note that trivially every s-inducing walk in $\cG$ is also an a-inducing walk.
\begin{remark}
    We defined s-inducing walks on a DMG $\cG$ only between nodes that are not in the same strongly connected component. This ensures the existence of a bidirected edge on every s-inducing walk. %
    \cref{lem:s_ind_LDG} shows how this enables $\sigma$-inducing walks on the lifted dependency graph of $\cG$ that in a sense correspond to bidirected edges. 
\end{remark}

\cref{prop:s-ind-walks} provides a remarkably powerful result: it allows us to immediately conclude that $w$ is inseparable from $v$ as soon as $v \notin \anc(w)$ and $v$ is inseparable from $w$.

\begin{proposition}
    \label{prop:s-ind-walks}
Let $\cG=(V, \cE_{d}, \cE_{bi})$ a DMG; $v, w \in V$. If $v \notin \anc(w)$ and there exists an a-inducing walk from $v$ to $w$ in $\cG$ (i.e. $v \in u(w,\cI_W^{\cG})$), then there exists an s-inducing walk from $v$ to $w$ in $\cG$.%
\end{proposition}

The following Lemma establishes an important property that allows to relate s-inducing walks to bidirected edges.

\begin{lemma}
    \label{lem:s_ind_LDG}
    Let $\cG=(V, \cE_{d}, \cE_{bi})$ a DMG; $v, w \in V$ and $\tilde{\cG}=(\tilde{V}, \tilde{\cE}_{d}, \tilde{\cE}_{bi})$ the lifted dependency graph of $\cG$. If there exists an s-inducing walk between $v$ and $w$ in $\cG$, then it follows:

\begin{enumerate}[label=\roman*)]
    \item There exists an s-inducing walk $\mu: v \circleft[e_{0}] v^{1} \sim \cdots \sim v^{n} \circright[e_{n}] w$ in $\cG$ that is into $v$ and into $w$. 
    There exists at least one bidirected edge on $\mu$. 
    \item For every $i,j \in \{0,1\}$ there exists a $\sigma$-inducing walk $\mu_{ij}: v_{i} \circleft[e_{0}] v_{\ell_1}^{1} \sim \cdots \sim v_{\ell_n}^{n} \circright[e_{n}] w_{j}$ in $\tilde{\cG}$
\end{enumerate}
\end{lemma}

We establish the following - almost trivial but highly useful - properties of a-inducing walks and s-inducing walks.
\begin{lemma}
\label{lem:intow}
Let $\cG=(V, \cE_{d}, \cE_{bi})$ be a DMG and $v, w \in V$, and let 
$\tilde{\cG}=(\tilde{V}, \tilde{\cE}_{d}, \tilde{\cE}_{bi})$ be the lifted dependency graph of $\cG$. %
If there exists an a-inducing path $\pi: v \mysim[e_0] \cdots \mysim[e_n] w$ from $v$ to $w$ in $\cG$, then %
it follows:
\begin{enumerate}[label=\roman*)]
    \item There exists an a-inducing path $\pi: v \mysim[e_0] \cdots \circright[e_n] w$ from $v$ to $w$ in $\cG$ that is into $w$.

    \item There exists a $\sigma$-inducing path $\tilde{\pi}: v_0 \mysim[\tilde{e}_0] \cdots \circright[\tilde{e}_n] w_1$ from $v_0$ to $w_1$ in $\tilde{\cG}$ that is into $w_1$.
    \item There exists a $\sigma$-inducing path $\tilde{\pi}_0: v_0 \mysim[\tilde{e}_0] \cdots \circright[\tilde{e}_n] w_0$ from $v_0$ to $w_0$ in $\tilde{\cG}$ that is into $w_0$.%
    \item There exists a $\sigma$-inducing path $\tilde{\pi}_1: v_1 \mysim[\tilde{e}_0] \cdots \circright[\tilde{e}_n] w_1$ from $v_1$ to $w_1$ in $\tilde{\cG}$ that is into $w_1$.%
\end{enumerate}
\end{lemma}

\begin{lemma} \label{lem:eqv_s-ind_2}
Let $\cG^1 = (V, \cE_d^1, \cE_{bi})$, $\cG^2 = (V, \cE_d^2, \cE_{bi})$ be DMGs, $\cI_W^\cG = \cI_W^{\cG'}$ and $v,w \in V$.
Then the following are equivalent:

\begin{enumerate}[label=\roman*)]
    \item Either $v \in \scc^{\cG^1}(w)$ or there exists an s-inducing path between $v$ and $w$ in $\cG^1$.
    
    \item Either $v \in \scc^{\cG^2}(w)$ or there exists an s-inducing path between $v$ and $w$ in $\cG^2$.
    
    \item There exist a-inducing paths from $v$ to $w$ and from $w$ to $v$ in $\cG^1$.
    
    \item There exist a-inducing paths from $v$ to $w$ and from $w$ to $v$ in $\cG^2$.
\end{enumerate}
\end{lemma}

We relate s-inducing walks and paths.

\begin{lemma}\label{lem:eqv_s-ind_1}
Let $\cG = (V, \cE_d, \cE_{bi})$ be a DMG, $v, w \in V$.
Then the following are equivalent

\begin{enumerate}[label=\roman*)]
    \item there exists an s-inducing walk from $v$ to $w$ in $\cG$
    
    \item there exists an s-inducing path from $v$ to $w$ in $\cG$
    
    \item there exists an s-inducing path from $w$ to $v$ in $\cG$
\end{enumerate}
\end{lemma}

\begin{lemma}\label{lem:s-ind_inseparability}
    Let $\cG = (V, \cE_d, \cE_{bi})$ be a DMG, $v, w \in V$. If there exist an s-inducing walk between $v$ and $w$ then $v\in u(w,\cI_W^\cG)$ and $w\in u(v,\cI_W^\cG)$.
\end{lemma}

\begin{proof}[Proof of Proposition \ref{prop:s-ind-walks}]
Let $\cG=(V, \cE_{d}, \cE_{bi})$ a DMG; $v, w \in V$ and $v \notin \anc(w)$.
 Let $\pi: v \mysim[e_0] v^1 \mysim[e_1] \cdots \mysim[e_n] v^{n+1} = w$ be an a-inducing path from $v$ to $w$ in $\cG$, $v \notin \scc(w)$.
    Let $s := \min\{s \in [n]: v^{s+1} \notin \scc(w)\}$. Such $s$ exist since $w = v^{n+1} \notin \scc(w)$.    
    Since $\ncoll_{\pi,\mathrm{bl}} \setminus \en_\pi = \emptyset$ it must be that $e_s = \text{edges}_\pi$ is into $\scc(v)$ or that $v^s = v$ (meaning $s=0$). But if $e_s = e_0$ is out of $v$ then there must exist a $v^k \in \coll_\pi$, $k > 0$ such that $v \in \anc(v^k)$.
    Since $v \notin \anc(w)$ and $\coll_{\pi} \subseteq \anc(\{v, w\})$ it follows that $v^k \in \anc(v)$ and hence for every $x \in [n]$ with $0 \leq x \leq k$ it holds $v^x \in \scc(v)$, a contradiction to $s = 0$.
    
    Hence $e_s$ is into $v^s \in \scc(v)$ and since ${\pi}$ is a-inducing it immediately follows that ${\pi}$ is an s-inducing path in $\cG$.
\end{proof}

\begin{proof}[Proof of Lemma \ref{lem:s_ind_LDG}]
Let $\cG=(V, \cE_{d}, \cE_{bi})$ a DMG; $v, w \in V$ and $\tilde{\cG}=(\tilde{V}, \tilde{\cE}_{d}, \tilde{\cE}_{bi})$ the lifted dependency graph of $\cG$ and let there exists an s-inducing walk from $v$ to $w$ in $\cG$.
We show that all of the following hold:
\begin{enumerate}[label=\roman*)]
    \item There exists an s-inducing walk $\mu: v \circleft[e_{0}] v^{1} \sim \cdots \sim v^{n} \circright[e_{n}] w$ in $\cG$ that is into $v$ and into $w$. 
    There exists at least one bidirected edge on $\mu$. 
    \item There exists a $\sigma$-inducing walk $\mu_{00}: v_{0} \circleft[e_{0}] v_{0}^{1} \sim \cdots \sim v_{0}^{n} \circright[e_{n}] w_{0}$ in $\tilde{\cG}$
    \item There exists a $\sigma$-inducing walk $\mu_{11}: v_{1} \circleft[e_{0}] v_{1}^{1} \sim \cdots \sim v_{1}^{n} \circright[e_{n}] w_{1}$ in $\tilde{\cG}$
    \item There exists a $\sigma$-inducing walk $\mu_{01}: v_{0} \circleft[e_{0}] v_{\ell_{1}}^{1} \sim \cdots \sim v_{\ell_{n}}^{n} \circright[e_{n}] w_{1}$ in $\tilde{\cG}$
    \item There exists a $\sigma$-inducing walk $\mu_{10}: v_{1} \circleft[e_{0}] v_{\ell_{1}}^{1} \sim \cdots \sim v_{\ell_{n}}^{n} \circright[e_{n}] w_{0}$ in $\tilde{\cG}$
\end{enumerate}

We consider:

\begin{enumerate}[label=\roman*)]

    \item Let $\mu$ be an s-inducing walk 
    $\mu: v \mysim[e_{0}] v^{1} \mysim \cdots \mysim v^{n} \mysim[e_{n}] w$ in $\cG$.
    
    Let $e_{t}$ be the first edge on $\mu$ after leaving $\scc(v)$ and $e_{s}$ be the last edge before entering $\scc(w)$.  
    Then the following valid $\mu$ exists:
    $\mu: v \myleftarrow \cdots \myleftarrow v^{t} \bidiedge[e_{t}] v^{t+1} \sim \cdots \\
    \sim v^{s} \bidiedge[e_{s}] v^{s+1} \myrightarrow 
    \cdots \myrightarrow w$
    
    Note that possibly $e_{t}=e_{s}=e_{0}=e_{n}$. \\
    Since $e_{0}$ and $e_{n}$ are into $v$ and $w$, respectively, $\ncoll_{\mu,\mathrm{bl}}=\{v, w\}$ and $v \neq \scc(w)$, there must exist at least one bidirected edge on $\mu$. If only one bidirected edge exists, then $e_{t}=e_{s}$.
    
    \item[\textnormal{(iii)/(iv)}] $\mu_{00} / \mu_{11}$ exist by construction of $\tilde{\cG}$ as the lifted dependency graph of $\cG$. They lie entirely in $V_{0} / V_{1}$ and every node $v_{0}^{i} \in \mu_{00} / v_{1}^{i} \in \mu_{11}$ is a copy of $v^{i} \in \mu$, respectively. 
    Thus $\coll_{\mu_{00}} \subseteq \anc(v_{0}, w_{0})$, $\ncoll_{\mu_{00},\mathrm{bl}}=\{v_{0}, w_{0}\}$, $\coll_{\mu_{11}} \subseteq \anc(v_{1}, w_{1})$, $\ncoll_{\mu_{11},\mathrm{bl}}=\{v_{1}, w_{1}\}$, and hence they are $\sigma$-inducing.
    
    \setcounter{enumi}{4}
    \item From ii) we know the s-inducing walk 
    $
    \mu: v \myleftarrow \cdots \myleftarrow v^{t} \bidiedge[e_{t}] v^{t+1} \sim \cdots \sim v^{s} \bidiedge[e_{s}] v^{s+1} \myrightarrow \ldots \myrightarrow w
    $
    exists, with $e_{t} / e_{s}$ the last edge on $\mu$ before leaving $\scc(v) /$ entering $\scc(w)$, respectively. \\
    By construction of $\tilde{\cG}$ as the lifted dependency graph of $\cG$, we know there exists the walk in $\tilde{\cG}$
    $
    \mu_{01}: v_{0} \myleftarrow \cdots \myleftarrow v_{0}^{t} \bidiedge v_{0}^{t+1} \sim \cdots \sim v_{0}^{s} \bidiedge v_{1}^{s+1} \myrightarrow \cdots \myrightarrow w_{1},
    $
    where $\forall v_{\ell_{i}}^{i} \in \mu_{01}: \ell_{i}=\begin{cases}0 & i \leq s \\ 1 & i > s\end{cases}$.\\
    Then $\ncoll_{\mu_{01},\mathrm{bl}}=\{v_{0}, w_{1}\}$, $\coll_{\mu_{01}} \subseteq V_{0}$. Since $\coll_{\mu} \subseteq \anc(v, w)$ it follows that $\coll_{\mu_{01}} \subseteq \anc(v_{0}, w_{0}) \subseteq \anc(v_{0}, w_{1})$.\\
    Hence $\mu_{01}$ is $\sigma$-inducing, into $v_{0}$ and into $w_{1}$.
    
    \item Analogously to v), we find that there exists the walk in $\tilde{\cG}$
    $
    \mu_{10}: v_{1} \myleftarrow \cdots \myleftarrow v_{1}^{t} \bidiedge v_{0}^{t+1} \sim \cdots \sim v_{0}^{s} \bidiedge v_{0}^{s+1} \myrightarrow \cdots \myrightarrow w_{0},
    $
    where $\forall v_{\ell_{i}}^{i} \in \mu_{10}: \ell_{i}=\begin{cases}1 & i \leq t \\ 0 & i>t\end{cases}$.\\
    Then $\ncoll_{\mu_{10},\mathrm{bl}}=\{v_{1}, w_{0}\}$ and $\coll_{\mu_{10}} \subseteq V_{0}$. Since $\coll_{\mu} \subseteq \anc(v, w)$ it follows that $\coll_{\mu_{01}} \subseteq \anc(v_{0}, w_{0}) \subseteq \anc(v_{1}, w_{0})$.\\ 
    Hence $\mu_{10}$ is $\sigma$-inducing, into $v_{1}$ and into $w_{0}$.
\end{enumerate}
\end{proof}

\begin{proof}[Proof of Lemma \ref{lem:intow}]
Let $e_s$ be the last edge on $\pi$ before entering $\mathrm{scc}(w)$.  
Since $\pi$ is a-inducing, it is of the form $v^s \circright[e_s] v^{s+1}$.  

\begin{enumerate}[label=\roman*)]
    \item From $\pi$ we can construct the a-inducing walk
    $$
    \pi': v \mysim[e_0] \cdots \mysim[] v^s \circright[{e}_s] v^{s+1} \myrightarrow[] \cdots \myrightarrow[e_m] w
    $$ 
    in $\cG$ that is into $w$, where $v^{s+1} \myrightarrow \cdots \myrightarrow[{e}_m] w$ is a directed walk in $\scc(w)$.
    If $\pi'$ is not already a path we can construct an a-inducing path from it, analogously as in the proof of \cref{prop:a_ind_eqv_s_sep}. Note that the last edge on $\pi'$ $e_m$ is not affected in this construction and is hence still into $w$.
    
    \item From $\pi'$ we can construct the walk 
    $$
    \tilde{\pi}: v_0 \mysim[\tilde{e}_0] \cdots \sim v_0^s \circright[\tilde{e}_s] v_1^{s+1} \myrightarrow \cdots \myrightarrow[\tilde{e}_m] w_1
    $$
     in $\tilde{\cG}$ where $\forall v_{\ell_i}^i \in \tilde{\pi}: \ell_i = 0$ if $i \leq s$, $\ell_i = 1$ if $i > s$  and $\ncoll_{\pi,\mathrm{bl}} = \{v_0, w_1\}$, $\coll_\pi \subseteq \anc(v_0, w_1) \cap V_0$. Hence $\tilde{\pi}$ is a $\sigma$-inducing path.\\
    
    We show that $\coll_\pi \subseteq \anc(Z) $ for all $ Z \subseteq \tilde{V}\setminus\{v_0,w_1\} $.
    Assume there exists $u_0^i \in \coll_{\tilde{\pi}}$ such that $u_0^i \in \anc(v_0) \setminus \anc(Z)$ (let $i$ be maximal with this property).  
    Then there exists a directed path 
    $u_0^i \myrightarrow \cdots \myrightarrow[v_0] v_0$ that does not intersect $Z$, and we can construct the $\sigma$-$Z$-open path
    $$
    \tilde{\mu}: v_0 \myleftarrow \cdots \myleftarrow u_0^i \mysim[\tilde{e}_i] \cdots \myrightarrow[\tilde{e}_m] w_1.
    $$

    Analogously for a collider $u_0^i \in \anc(w_1) \setminus \anc(Z)$ (take $i$ minimal with this property).  
    Note that $\tilde{\mu}$ is still into $w_1$.  
    
    Hence, a $\sigma$-$Z$-open path from $v_0$ to $w_1$ in $\tilde{\cG}$ exists that is into $w_1$.  

    \item[\textnormal{(ii)/(iii)}] follow analogously to (i), with 
    $$
    \tilde{\pi_0}: v_0 \mysim[\tilde{e}_0] \cdots \sim v_0^s \circright[\tilde{e}_s] v_0^{s+1} \myrightarrow \cdots \myrightarrow[\tilde{e}_m] w_0
    $$
    where $\forall v_{\ell_i}^i \in \tilde{\pi}_0: \ell_i = 0$, and
    $$
    \tilde{\pi_1}: v_1 \mysim[\tilde{e}_0] \cdots \sim v_1^s \circright[\tilde{e}_s] v_1^{s+1} \myrightarrow \cdots \myrightarrow[\tilde{e}_m] w_1
    $$
    where $\forall v_{\ell_i}^i \in \tilde{\pi}_1: \ell_i = 1$.
\end{enumerate}
\end{proof}

\begin{proof}[Proof of \cref{lem:eqv_s-ind_2}]\hfill

\begin{description}
    \item[$i) \Rightarrow iii)$] (analogously for $ii) \Rightarrow iv)$)
    
    If $v \in \scc(w)$ then this is trivial. Otherwise there exists an s-inducing path $\pi$ from $v$ to $w$ in $\cG$ and since $\pi^{-1}$ is s-inducing from $w$ to $v$ in $\cG$ this trivially concludes the statement.
    
    \item[$iii) \Rightarrow i)$] (analogously $iv) \Rightarrow ii)$)
    
    Follows trivially by using \cref{prop:s-ind-walks}.
    
    \item[$ii) \Leftrightarrow iv)$]
    
    Follows since $\cI_W^\cG = \cI_W^{\cG'}$ by using \cref{prop:a_ind_eqv_s_sep}.
\end{description}
\end{proof}

\begin{proof}[Proof of \cref{lem:eqv_s-ind_1}]\hfill
\begin{description}
    \item[${ii) \Leftrightarrow iii)}$] 
    Let $\pi$ be an s-inducing path from $v$ to $w$ in $\cG$. Then this proof trivially follows since $\pi^{-1}$ is s-inducing.

    \item[$ii) \Rightarrow i)$] Trivial.

    \item[$i) \Rightarrow ii)$]
    Let $\mu: v \mysim[e_0] v^1 \mysim \cdots \mysim v^n \mysim[e^n] w$ be an s-inducing walk from $v$ to $w$ in $\cG$ and let $s := \max\{s \in [n]: v^s \in \scc(v)\}$.
    
    By replacing the subwalk from $v$ to $v^s$ on $\pi$ with the shortest directed walk $v \myleftarrow \cdots \myleftarrow v^s$ we obtain an s-inducing walk $\mu': v \myleftarrow \cdots \myleftarrow v^s \bidiedge[e_s] \cdots \mysim w$ in $\cG$.
    
    Let $t := \min\{t \in [s+1,n]: v^t \in \scc(w)\}$.    
    By replacing the subwalk from $v^t$ to $w$ on $\pi$ with the shortest directed walk $v^t \myrightarrow \cdots \myrightarrow w$ we obtain the s-inducing walk $\pi: v \myleftarrow \cdots \myleftarrow v^s \bidiedge[e_s] \cdots \bidiedge[e_{t-1}] v^t \myrightarrow \cdots \myrightarrow w$ in $\cG$.
    
    Then by definition of $s$ and $t$ and by construction of
    $\pi$ it follows that if there exist $v^i,v^j\in \nodes_\pi$ such that $i \neq j$ but $v^i = v^j$,
    then $s < i, j < t$. We drop the subwalk from $v^i$ to $v^j$ on $\pi$ and obtain the walk $\pi': v \myleftarrow \cdots \myleftarrow v^s \bidiedge[e_s] \cdots \mysim[e_{i-1}] v^i =v^j \mysim[e_j] \cdots \bidiedge[e_{t-1}] v^t \myrightarrow \cdots \myrightarrow w$
    
    Since $\pi$ is s-inducing it holds that $\coll_\pi \subseteq \anc(v, w)$ and $\ncoll_{\pi,\mathrm{bl}}\setminus \en_\pi = \emptyset$.
    
    If $v^i \in \coll_{\pi'}$, then it must be that either $\{v^i, v^j\} \cap \coll_\pi \neq \emptyset$ and hence
    $v^i \in \anc(v, w)$, or $v^i,v^j \in \ncoll_{\pi}$ are unblockable and $e_{i-1}$ is into $v^i$ and $e_j$ is into 
    $v^j$. In the latter, it follows that $e_{i}\in \edges_\pi$ is out of $v^i$ and $e_{j-1}\in \edges_\pi$ is out of $v^j$. Hence there must exist a $v^k \in \coll_\pi \subseteq \anc(v, w)$ with $v^i \in \anc(v^k)$ and hence $v^i \in \anc(v, w)$.
    
    If $v^i \in \ncoll_{\pi'}$ then $v^i$ is unblockable, since then $e_{i-1}\in \edges_\pi$ is out of $v^i$ or $e_j \in \edges_\pi$, is out of $v^j$ and $v^{i-1} \in \scc(v^i)$ or $v^{j+1} \in \scc(v^j)$, respectively.
    
    Hence $\pi'$ is an s-inducing walk. By iteratively applying this step to all $v^i,v^j\in \nodes_\pi$ with $i \neq j$, $v^i = v^j$ we obtain an s-inducing path. Note that $e_s$ and $e_{t-1}$ are unaffected in this procedure since $s < i, j < t$ and hence they are still into $\scc(v)$ and $\scc(w)$, respectively.
\end{description}
\end{proof}

\begin{proof}[Proof of \cref{lem:s-ind_inseparability}]
    Let $\cG = (V, \cE_d, \cE_{bi})$ be a DMG, $v, w \in V$ and let there exist an s-inducing walk $\pi:v\mysim \cdots \mysim w$ in $\cG$. Then $v\in u(w,\cI_W^\cG)$ and $w\in u(v,\cI_W^\cG)$ trivially follow form \cref{prop:a_ind_eqv_s_sep} since $\pi$ and $\pi^{-1}$ are a-inducing walks. 
\end{proof}

\begin{proof}[Proof of \cref{lem:equivalent_DMGs_inducing_walks}]
    The proof immediately follows from \cref{def:markov_eqv} of Markov equivalence and \cref{prop:a_ind_eqv_s_sep}.
\end{proof}

\subsection{Proofs on Markov Equivalent Supergraphs of DMGs.}
\label{app:proofs_markov_equivalence}

\begin{proof}[Proof of \cref{prop:pot_sib_in_G}]
    The proof follows trivially from \cref{prop:a_ind_eqv_s_sep} and the definition of potential siblings.
\end{proof}

\begin{proof}[Proof of \cref{lem:sib_are_pot_sib}]
    The proof follows trivially from \cref{prop:pot_sib_in_G} since $a \bidiedge b$ and $b \bidiedge a$ are a-inducing paths in $\cG$.
\end{proof}

\begin{proof}[Proof of Proposition \ref{prop:adding_bidi_edges}]
Let $\tilde{\cG}=\left(\tilde{V}, \tilde{\cE_{d}}, \tilde{\cE_{bi}}\right)$ and $\tilde{\cG}'=\left(\tilde{V}, \tilde{\cE_{d}}, \tilde{\cE'_{bi}}\right)$ be the lifted dependency graphs of $\cG$ and $\cG'$, respectively.
Note that $v$ and $w$ being a potential siblings is equivalent to the existence of both an a-ind. path from $v$ to $w$ in $\cG$ and an a-ind. path from $w$ to $v$ in $\cG$. We will therefore proof that the following are equivalent:
\begin{enumerate}[label=\roman*)]
    \item $\cI_{W}^{\cG}=\cI_{W}^{\cG'}$
    
    \item There exist both an a-ind. path from $v$ to $w$ in $\cG$ and an a-ind. path from $w$ to $v$ in $\cG$
\end{enumerate}

\begin{description}[style=nextline]
    \item[$\mathbf{i) \implies ii)}$] 
    Let us assume the contrary: $\lnot ii) \Rightarrow \lnot i)$\\
    W.l.o.g., let there be no a-inducing path from $v$ to $w$ in $\cG$.\\
    Then it follows by \cref{prop:a_ind_eqv_s_sep} that there exist $ C, D \subseteq V$ such that $(\{v\},\{w\}, C, D) \in \cI_{W}^{\cG}$.\\
    
    But in $\tilde{\cG'}$ it exist the path $\tilde{\mu}: v_{0} \bidiedge[] w_{1}$, and thus $v_0 \notindep^{\tilde{\cG}'}_\sigma w_1 \mid Z \quad \forall Z \subseteq \tilde{v}\{v_{0}, w_{1}\}$, hence $(\{v\},\{w\}, C, D) \notin \cI_{W}^{\cG'}$ and therefore $\cI_{W}^{\cG} \neq \cI_{W}^{\cG'}$. %

    \item[$\mathbf{ii) \implies i)}$] 
    Let us assume the contrary:\\
    Assume there exist both a-ind. paths from $v$ to $w$ and from $w$ to $v$ in $\cG$, but there exist $a, b \in V; C, D \subseteq V$ such that $(\{a\},\{b\}, C, D) \in \cI_{W}^{\cG}$ but $(\{a\},\{b\}, C, D) \notin \cI_{W}^{\cG'}$.\\
    Let $Z:=C_{0} \cup D_{1}$.\\
    Then there exists a $\sigma$-$Z$-open walk $\pi': a_{0} \mysim[e'_0] v_{\ell_1}^{1} \sim \ldots \mysim[e'_m] b_{1}$ in $\tilde{\cG}'$.\\
    
    If $[(v_{0}, w_{0})], [(v_{0}, w_{1})], [(v_{1}, w_{0})], [(v_{1}, w_{1})] \in \pi'$, then we can replace each of these edges by a respective $\sigma$-$Z$-open walk $\mu_{ij}: v_{\ell_i}^{i} \mysim[\tilde{e_0}] \ldots \mysim[\tilde{e_0}] v_{\ell_j}^{j}$, $i,j \in [m]$, in $\tilde{\cG}$.\\
    The obtained walk $\pi: a_{0} \mysim[e_0] \ldots \mysim[e_n] b_{1}$ is then a walk in $\tilde{\cG}$.\\
    We show that for each $[(v_i, w_j)]$ a respective $\mu_{ij}$ exists such that:
    
    \begin{itemize}
      \item $\coll \mu_{ij} \subseteq \anc(\{v_{\ell_i}^{i}, v_{\ell_j}^{j}\})$ and $\ncoll_{\mu_{ij},\mathrm{bl}} = \{v_{\ell_i}^{i}, v_{\ell_j}^{j}\}$
      \item $v_i$ and $w_i$ do not block $\pi$.
    \end{itemize}
    Let us consider how to replace each edge.
    
    \begin{description}[font=\normalfont\underline]
      \item[{1) Replace $[(v_{0}, w_{0})]$:}]
      \hfill \\
      Let us do a case distinction:
      \begin{description}[font=\normalfont]
            \item[{1.1) $v_{0} \in \scc(w_{0})$}]\hfill \\
            Then replace $[(v_{0}, w_{0})]$ in $\pi'$ by the directed walk $\mu_{00}: v_{0} \myrightarrow[\tilde{e}_0] \ldots \myrightarrow[\tilde{e_n}] w_{0}$ that lies entirely in $\scc(w_{0})$.\\
            Thus
            \begin{itemize}
              \item $v_0 \in \ncoll_\pi$ and is unblockable or $v_0 \notin Z$\\ (since if $v_0 \in \ncoll_{\pi,\mathrm{bl}}$ then $v_0 \in \ncoll_{\pi',\mathrm{bl}} $ but $ \pi' $ is $\sigma$-$Z$-open.)
              \item since $\mu_{00}$ is into $w_0$ as is $[(v_0, w_0)] \in \pi'$ and $\pi'$ is $\sigma$-$Z$-open, $w_1$ does not $\sigma$-$Z$-block $\pi$.
              \item $\coll_{\mu_{00}}=\emptyset$, $\ncoll_{\mu_{00},\mathrm{bl}}=\{v_0, w_0\}$
            \end{itemize}
            
            \item[{1.2) $v_0 \notin \scc(w_0)$}]\hfill \\
            Then by \cref{prop:s-ind-walks,lem:s_ind_LDG} we know that there exists a $\sigma$-inducing walk $\mu_{00}: v_0 \circleft[\tilde{e_0}] v_0^1 \sim \cdots \sim v_0^n \circright[\tilde{e}_u] w_0$ in $\tilde{\cG}$.\\
            Such that
            \begin{itemize}
              \item $\coll \mu_{00} \subseteq \anc(v_0, w_0), \ncoll_{\mu_{00},\mathrm{bl}}=\{v_0, w_0\}$
              \item since $\mu_{00}$ is into $v_0$ and $w_0$, as is $[(v_0, w_0)] \in \pi'$ and $\pi'$ is $\sigma$-$Z$-open, $v_0$ and $w_0$ do not $\sigma$-$Z$-block $\pi$.
            \end{itemize}
      \end{description}

      \item[{2) Replace $[(v_{1}, w_{1})]$:}] Follows analogously to replacing $[(v_{0}, w_{0})]$.

      \item[{3) Replace $[(v_0, w_1)]$:}]
      \hfill \\
      Let us do a case distinction:
      \begin{description}[font=\normalfont]
        \item [{3.1) $v_0 \notin Z$}] \hfill \\
        Then we replace $\left[\left(v_{0}, w_{1}\right)\right]$ on $\pi'$ by the $\sigma$-inducing walk in $\widetilde{\cG}$\\
        $$
        \mu_{01}: v_{0} \circright[\tilde{e_0}] v_{\ell_1}^1 \sim \cdots \circright[\tilde{e_n}] w_1
        $$
        We know such walk exists since it exist an a-inducing walk from $v$ to $w$ in $\cG$. (Note that if $\tilde{e_0}$ was out of $v_{\ell_1}^1$ then $v_{\ell_1}^1 \in \scc(v_0)$ and hence $\mu_{01}$ with $\tilde{e_0}$ into $v_{\ell_1}^1$ also exist.)\\
        Thus
        
        \begin{itemize}
          \item $\coll_{\mu_{01}} \leq \anc\left(v_{0}, w_{1}\right), \quad \ncoll_{\mu_{01},\mathrm{bl}}=\left\{v_{0}, w_{1}\right\}$
          \item \textbf{either} $v_{0} \in \ncoll_\pi \wedge v_{0} \notin Z$ \textbf{or} $v_{0} \in \coll_\pi \wedge v_{0} \in \anc(Z)$ since then $v_{0} \in \coll_{\pi'}$ and $\pi'$ is $\sigma$-$Z$-open.\\ 
          Hence for both cases it follows $ v_{0}$ does not $\sigma$-$Z$-block $\pi$.
          \item since $\mu_{01}$ is into $w_{1}$ as is $\left[\left(v_{0}, w_{1}\right)\right] \in \pi'$ and $\pi'$ is $\sigma$-$Z$-open, $w_{1}$ does not $\sigma$-$Z$-block $\pi$.\\
        \end{itemize}

        \item [{3.2) $v \notin \scc (w)$}]\hfill \\
        Then by \cref{prop:s-ind-walks,lem:s_ind_LDG} we know that it exist a $\sigma$-inducing walk in $\tilde{\cG}$
        $$
        \mu_{01}: v_0 \circleft[\tilde{e_0}] v_0^1 \sim \cdots \sim v_0^n \circright[\tilde{e}_u] w_1 
        $$
        Thus
        \begin{itemize}
          \item $\coll \mu_{0} \subseteq \anc\left(v_{0}, w_{1}\right)$, $ \ncoll_{\mu_{01},\mathrm{bl}}=\left\{v_{0}, w_{1}\right\}$
          \item since $\mu_{01}$ is into $v_{0}$ and into $w_{1}$, as is $\left[\left(v_{0}, w_{1}\right)\right] \in \pi'$, and $\pi'$ is $\sigma$-$Z$-open, $v_{0}$ and $w_{1}$ do not $\sigma$-$Z$-block $\pi$.\\
        \end{itemize}

        \item [{3.3) $v_0 \in Z \wedge v \in \scc (w)$}] \label{case:3_3_adding_bidi}\hfill \\ 
        Remember we start with the $\sigma$-$Z$-open walk $\pi': a_{0} \mysim[e'_0] v_{\ell_1}^{1} \sim \ldots \mysim[e'_m] b_{1}$ in $\tilde{\cG}'$ and we want to replace the edge $e_{\beta}'=\left[\left(v_{0}, w_1 \right)\right] \in \pi'$.\\
        
        Let 
        $$
        d := \max \left\{d \in [\beta-1]: e_{d} \in \cE_{bi} \lor v_{l_{d}}^{d} \notin Z \right\}
        $$
        Note that $d$ is well defined as $v_{\ell_{0}}^{0}=a_{0} \notin Z $ since $Z \subseteq \tilde{V} \setminus \{a_0, b_1\}$.\\
        
        Note that 
        
        \begin{itemize}
            \item $\forall v_{\ell_i}^{i} \in \pi'$ with $d < i \leq \beta $ it holds $ v_{\ell_i}^{i} \in Z$
            \item $\forall e_{i}' \in \pi'$ with $d < i < \beta $ it holds $ e_{i}' \in \tilde{\cE_{d}}$
            \item $\forall v_{\ell_{i}}^{i} \in \pi'$ with $d < i \leq \beta$ it holds $v_{\ell_{i}}^{i} \in \scc(v_{0}) \wedge l_{i}=0$\\
            (This is a consequence of the previous two, since $\pi' $ is $\sigma$-$Z$-open.)
        \end{itemize}
        
        Hence the following $\pi'$  exists:  
        $
        \pi': a_{0} \mysim[e_0'] \cdots \sim v_{\ell_d}^{d} \circright[e_d'] v_{0}^{d+1} \myrightarrow \cdots \myrightarrow[] v_{0}^{\beta}=v_{0} \bidiedge[e_\beta'] w_{1} \sim \cdots \mysim[e_m'] b_{1}
        $.
        
        (Possibly $v_{0}^{d+1}=v_{0}^{\beta}\coloneqq v_{0}$).
        Since from $v_{0}^{d+1} \in \scc(v_0), v \in \scc (w)$ it follows $v_{1}^{d+1} \in \scc (w_1)$, one can choose the walk in $\tilde{\cG}$
        $$
        \mu_{d 1}: v_{\ell_{d}}^{d} \circright[\hat{e}_{d}'] v_{1}^{d+1} \myrightarrow[] \cdots \myrightarrow[] w_{1}
        $$
        to replace the subwalk from $v_{l_{d}}^{d}$ to $w_{1}$ on $\pi'$, thereby replacing $e_{\beta}'=\left[\left(v_{0}, w_{1}\right)\right]$ as well.\\
        
        Note that we can assume that all $\left[\left(v_{0}, w_{0}\right)\right] \in \pi'$ have been removed previously and therefore $e_{d}'\neq \left[\left(v_{0}, w_{0}\right)\right]$ and we do not introduce a new edge $\hat{e}_{d}'=\left[\left(v_{0}, w_{1}\right)\right]$ in $\pi$. If $e_{d}'=\left[\left(v_{1}, w_{0}\right)\right]$ and thus $\hat{e}_{d}'=\left[\left(v_{1},w_1\right)\right]$, we can remove it afterwards.\\
        
        Note $\hat{e}_{d}'$ exists since $e_{d}'$ was into $v_{0}^{d+1}$.
        
        We find that:
        \begin{itemize}
          \item $\coll_{\mu_{d 1}}=\emptyset \subseteq \anc\left(v_{0}, w_{1}\right), \ncoll_{\mu_{d 1},\mathrm{bl}}=\left\{v_{l_{d}}^d, w_{1}\right\}$
          \item since $\mu_{d1}$ is endpoint identical and direction preserving and $\pi'$ is $\sigma$-$Z$-open, $v_{l_{d}}^{d}$ and $w_{1}$ do not $\sigma$-$Z$-block $\pi$.
        \end{itemize}
      \end{description}

    \item[{4) Replace $[(v_{1}, w_{0})]$:}]    
    \hfill \\
    
    We do a case distinction:\\
    \begin{description}[font=\normalfont]

        \item[{4.1) $w_{0} \notin Z$}] 
        \hfill \\
        Then we replace $\left[\left(v_{1}, w_{0}\right)\right]$ on $\pi'$ by the $\sigma$-inducing walk in $\tilde{\cG}$
        $$
        \mu_{10}: v_{1} \circleft[\tilde{e}_{0}] v_{\ell_1}^1 \sim \cdots \sim v_{\ell_n}^{n} \circleft[\tilde{e}_n] w_{0}
        $$
        We know such walk exists since it exist an a-inducing walk from $w$ to $V$ in $\cG$.\\
        Thus
        
        \begin{itemize}
          \item $\coll_{\mu_{10}} \subseteq \anc\left(v_{1}, w_{0}\right), \quad \ncoll_{\mu_{10},\mathrm{bl}}=\left\{v_{1}, w_{0}\right\}$
          \item \textbf{either} $w_{0} \in \ncoll_\pi \wedge \, w_{0} \notin Z$ \textbf{or} $w_{0} \in \coll_\pi \wedge \, w_{0} \in \anc(Z)$ since then $w_{0} \in \coll_{\pi'}$ and $\pi'$ is $\sigma$-$Z$-open.\\ 
          Hence for both cases it follows $ v_{0}$ does not $\sigma$-$Z$-block $\pi$.
          \item since $\mu_{10}$ is into $v_{1}$ as is $\left[\left(v_{1}, w_{0}\right)\right] \in \pi'$ and $\pi'$ is $\sigma$-$Z$-open, $v_{1}$ does not $\sigma$-$Z$-block $\pi$.\\
        \end{itemize}
        
        \item [{4.2) $v \notin \scc (w)$}]\hfill \\
        Then by \cref{prop:s-ind-walks,lem:s_ind_LDG} we know it exist a $\sigma$-inducing walk in $\tilde{\cG}$
        $$
        \mu_{10}: v_{1} \circleft[\tilde{e}_{0}] v_{\ell_{1}}^{1} \sim \cdots \sim v_{\ell_{n}}^{n} \circright[\tilde{e}_{n}] w_{0}
        $$
        Thus
        \begin{itemize}
          \item $\coll_{\mu_{10}} \subseteq \anc\left(v_{1}, w_{0}\right),  \ncoll_{\mu_{10},\mathrm{bl}}=\left\{v_{1}, w_{0}\right\}$
          \item since $\mu_{10}$ is into $v_{1}$ and into $w_{0}$, as is $\left[\left(v_{1}, w_{0}\right)\right] \in \pi'$, and $\pi'$ is $\sigma$-$Z$-open, $v_{1}$ and $w_{0}$ do not $\sigma$-$Z$-block $\pi$.\\
        \end{itemize}
        \item[{4.3) $w_{0} \in Z \wedge v \in \scc(w)$}] \label{case:4_3_adding_bidi}\hfill \\ 
        Remember we start with the $\sigma$-$Z$-open walk $\pi': a_{0} \mysim[e'_0] v_{\ell_1}^{1} \sim \ldots \mysim[e'_m] b_{1}$ in $\tilde{\cG}'$ and we want to replace the edge $e_{\beta}'=\left[\left(v_{1}, w_0\right)\right] \in \pi'$.\\
        Let 
        $
        d := \min \left\{d \in \{\beta+1, \ldots, m\}: e_{d}' \in \cE_{b i} \lor v_{l_{d+1}}^{d+1} \notin Z \right\}
        $
        Note that $d$ is well defined as $b_{1}=v_{\ell_{n+1}}^{n+1} \notin Z $ since $Z \subseteq \tilde{V}\setminus \{a_{0}, b_{1}\}$\\
        
        Note that 
        
        \begin{itemize}
          \item $\forall v_{\ell_{i}}^{i} \in \pi'$ with $\beta < i \leq d $ it holds $ v_{\ell_{i}}^{i} \in Z$
          \item $\forall e_{i}' \in \pi'$ with $\beta < i < d $ it holds $ e_{i}' \in \tilde{\cE}_{d}$
          \item $\forall v_{\ell_{i}}^{i} \in \pi'$ with $\beta < i \leq d$ it holds $v_{\ell_{i}}^{i} \in \scc(w_{0}) \wedge l_{i}=0$\\
          (This is a consequence of the previous two, since $\pi' $ is $\sigma$-$Z$-open.)
        \end{itemize}
        
        Hence the following is an existing $\pi'$ :
        
        $
        \pi': a_{0} \mysim[e'_0] \cdots \sim v_{1}^{\beta}=v_{1} \bidiedge[e_{\beta}'] w_{0}=v_{0}^{\beta +1} \myleftarrow[] \cdots \myleftarrow[] v_{0}^{d} \circright[e_d'] v_{\ell_{d+1}}^{d+1} \sim \ldots \mysim[e'_m]{\sim} b_{1} 
        $
        (Possibly $v_{0}^{d}=v_{0}^{\beta+1} \coloneqq w_{0}$)\\
        
        Since $v_{0}^{d} \in \scc (w_{0})$, $v \in \scc (w)$ it follows $v_{1}^{d} \in \scc (v_{1})$, one can choose the walk in $\tilde{\cG}$ 
        $$
        \mu_{1 d}: v_{1} \myleftarrow[] \cdots \myleftarrow[] v_{1}^{d} \circleft[\hat{e}_{d}'] v_{\ell_{d+1}}^{d+1}
        $$
        
        to replace the subwalk from $v_{1}$ to $v_{\ell_{d+1}}^{d+1}$ on $\pi'$, thereby replacing $e_{\beta}'=\left[\left(v_{1}, w_{0}\right)\right]$ as well.\\

        Note that we can assume that all $\left[\left(v_{0}, w_{0}\right)\right] \in \pi'$ have been removed previously and therefore $e_{d}'\neq \left[\left(v_{0}, w_{0}\right)\right]$ and we do not introduce a new edge $\hat{e}_{d}'=\left[\left(v_{0}, w_{1}\right)\right]$ in $\pi$. If $e_{d}'=\left[\left(v_{1}, w_{0}\right)\right]$ and thus $\hat{e}_{d}'=\left[\left(v_{1},w_1\right)\right]$, we can remove it afterwards.\\
        
        Note $\hat{e}_{d}'$ exists since $e_{d}'$ was into $v_{0}^{d}$.\\
        
        We find that:
        \begin{itemize}
          \item $\coll_{\mu_{1 d}}=\emptyset \subseteq \anc\left(v_{1}, w_{0}\right), \\ \ncoll_{\mu_{1 d},\mathrm{bl}}=\left\{v_{1}, v_{\ell_{d+1}}^{d+1}\right\}$
          \item since $\mu_{1 d}$ is endpoint identical and direction preserving and $\pi'$ is $\sigma$-$Z$-open, $v_{1}$ and $v_{l_{d+1}}^{d+1}$ do not $\sigma$-$Z$-block $\pi$.
        \end{itemize}

    \end{description}
    \end{description}
    \hfill \\
    We have now shown that if any of the edges $\left[\left(v_{0}, w_{0}\right)\right], \left[\left(v_{0}, w_{1}\right)\right], \left[\left(v_{1}, w_{0}\right)\right], \left[\left(v_{1}, w_{1}\right)\right] $ is in $ \pi'$, then we can replace each of these edges by a respective walk 
    $$
    \mu_{ij}: v_{\ell_i}^{i} \mysim[\tilde{e_0}] \ldots \mysim[\tilde{e_0}] v_{\ell_j}^{j}
    $$
    in $\tilde{\cG}$ such that the obtained walk 
    $$
    \pi: a_{0} \mysim[e_0] \cdots \mysim[e_n] b_{1}
    $$
    is a walk in $\tilde{\cG}$ and
    
    \begin{itemize}
      \item the endnodes of $\mu_{i j}$, $v_{\ell_{i}}^{i}$ and $v_{\ell_{j}}^{j}$, do not $\sigma$-$Z$-block $\pi$
      \item $\ncoll_{\mu_{i j},\mathrm{bl}}=\left\{v_{\ell_{i}}^{i},v_{\ell_{j}}^j\right\}$
      \item $ \coll_{\mu_{i j}} \subseteq \anc\left(\left\{v_{\ell_{i}}^{i},v_{\ell_{j}}^j\right\}\right)$
    \end{itemize}

    We now want to show that $\pi$ is $\sigma$-$Z$-open. Since $\pi'$ was $\sigma$-$Z$-open, the only thing left to show is that $\coll_{\mu_{i j}}$ do not $\sigma$-$Z$-block $\pi$.
    
    Assume it exist $ v_{\ell_{k}}^{k} \in \coll \mu_{i j}$ with $v_{\ell_{k}}^{k} \in \anc\left(v_{\ell_{i}}^{i}\right) \setminus \anc(Z)$.
    Let it be maximal w.r.t. this property.\\
    Then we could replace the subwalk from $v_{\ell_{i}}^{i} \sim \ldots \circright[] v_{\ell_k}^{k}$ on $\pi$ by the directed walk $v_{\ell_{i}}^{i} \myleftarrow[] \cdots \myleftarrow[] v_{\ell_{k}}^{k}$ that does not intersect $Z$.
    
    Then 
    $$
    \pi: a_{0} \sim \cdots \sim v_{\ell_{i}}^{i} \myleftarrow[e_i] \cdots \myleftarrow[] v_{\ell_{k}}^{k} \sim \cdots \sim b_{1} 
    $$
    would be $\sigma$-$Z$-open, since now $v_{\ell_{k}}^{k} \in \ncoll_{\pi}$ and we only added non-colliders that are not in $Z$.\\
    
    Note that $e_{i}$ being into $v_{\ell_{i}}^{i}$ does not cause any issues, since in \ref{case:3_3_adding_bidi} and \ref{case:4_3_adding_bidi} it holds $\coll_{\mu_{i j}}=\emptyset$ and for all other cases $v_{\ell_{i}}^{i} \in \{v_{0}, v_{1}\}$ and $e_{i}'=\left[\left(v_{\ell_{i}}, w_{\ell_{i}}\right)\right]$ was into $v_{\ell_{i}}^{i}$ but $\pi'$ was $\sigma$-$Z$-open.\\
    Analogously for $v_{\ell_{k}}^{k} \in \anc\left(v_{\ell_{j}}^{j}\right) \setminus \anc(Z)$.\\
    
    Thus we showed that there exists a walk $\pi: a_{0} \sim \cdots \sim b_{1}$ in $\tilde{\cG}$ that is $\sigma$-$Z$-open and hence
    $$
    (\{a\},\{b\}, C, D) \notin \cI_{W}^{\cG},
    $$
    a contradiction to our assumption.
\end{description}
\end{proof}

\begin{proof}[Proof of \cref{lem:pa_are_pot_pa}]
    Let $\cG = (V, \cE_d, \cE_{bi})$ be a DMG, $v,w \in V$ and $(v, w) \in \cE_{d}$. We test conditions \cref{cond:potential_parents_a,cond:potential_parents_b,cond:potential_parents_c,cond:potential_parents_d} of \cref{def:potential_parents} for potential parents.
    \begin{enumerate}
        \item a) $v\in u (w, \cI_W^\cG)$ trivially holds by \cref{prop:a_ind_eqv_s_sep} since $v\myrightarrow w $ is an a-inducing path.

        \item b) trivially holds since then $u,x \in \scc(w)$.

        \item c) Assume the contrary, i.e. there exist $a, b \in V$, $C,D \subseteq V$, with $a \indep_W^\cG b \mid C, D$ and $a \indep_W^\cG b \mid C, D\cup \{w\} : a \notindep_W^\cG b \mid C, D\cup \{v\} $. Define $Z:=C_0 \cup D_1$ and let $\tilde{\cG}$ be the lifted dependency graph of $\cG$.
        Then w.l.o.g. there exist a $\sigma$-$Z$-blocked walk $\pi$ in $\tilde{\cG}$ with $\ncoll_{\pi,\mathrm{bl}}\cap Z = \emptyset$ and $\coll_\pi \setminus \anc(Z) = \{v_1\}$. Since $v_1\myrightarrow w_1$ exist it follows that $\coll_\pi \setminus \anc(Z\cup \{w_1\}) = \emptyset$ and hence $\pi$ is $\sigma$-$Z\cup \{w_1\}$-open, a contradiction to $a \indep_W^\cG b \mid C, D\cup \{w\}$.

        \item d) Follows analogously to c).
    \end{enumerate}
\end{proof}   

\begin{proof}[Proof of \cref{prop:adding_directed_edges}]
Let $\tilde{\cG} = (\tilde{V}, \tilde{\cE}_d, \tilde{\cE}_{bi})$ and $\tilde{\cG}' = (\tilde{V}, \tilde{\cE}_d', \tilde{\cE}_{bi})$, with $\cE_d'=\cE_d \cup \{(v,w)\}$, be the lifted dependency graphs of $\cG$ and $\cG'$, respectively.
We proof that the following are equivalent:\\
\textbf{i)} $\cI_W^\cG = \cI_W^{\cG'}$\\
\textbf{ii)} $v$ is a potential parent of $w$ i.e. all of the following hold:
\begin{enumerate}[label=\alph*)]
    \item  $v\in u (w, \cI_W^\cG)$ %
    
    \item $\forall u, x \in \anc^\cG(v) \cap \des^\cG(w)$:
    $u\in u (x, \cI_W^\cG)$

    \item $\forall a, b \in V$, $C,D \subseteq V$, with $a \indep_W^\cG b \mid C, D$ and $a \indep_W^\cG b \mid C, D \cup \{w\}: a \indep_W^\cG b \mid C, D \cup \{v\}$

    \item $\forall a, b \in V$, $C,D \subseteq V$, with $a \indep_W^\cG b \mid C, D$ and $a \indep_W^\cG b \mid C\cup \{w\}, D : a \indep_W^\cG b \mid C\cup \{v\}, D $
 
\end{enumerate}

Note that $v\in u (w, \cI_W^\cG)$ is equivalent to the existence of an a-inducing walk from $v$ to $w$ in $\cG$, according to \cref{prop:a_ind_eqv_s_sep}.

\underline{$\mathbf{i) \Rightarrow ii)}$} \hfill    (via $\neg ii) \Rightarrow \neg i)$)
    
    \begin{description}[style=nextline, font=\normalfont]
        \item[$\neg ii a) \Rightarrow \neg i)$] Assume $v\notin u (w, \cI_W^\cG)$.
        Then it follows that there exist $C,D \subseteq V $ such that $(\{v\}, \{w\}, C, D) \in \cI_W^\cG$.
        But in $\tilde{\cG}'$ it exist  the walk $\pi': v_0 \myrightarrow w_1$  that is $\sigma$-$Z$-open in $\tilde{\cG}'$ 
        $\forall \, Z \coloneqq C_0 \cup D_1$ with ${C} , {D} \subseteq V$, and hence $(\{v\}, \{w\}, C, D) \notin \cI_W^{\cG'}$.
        
        \item[ $\neg ii b)\Rightarrow \neg i)$]
        
        Assume it exist $u, x \in \anc^\cG(v) \cap \des^\cG(w)$ such that there exists no a-inducing walk from $u$ to $x$ in $\tilde{\cG}$.
        
        Then there exist $C, D \in V$ such that $u \indep_W^\cG x \mid C, D$ \quad (according to \cref{prop:a_ind_eqv_s_sep})
        
        But since $(v, w) \in \cE_d'$ it follows that $v \in \anc^{\cG'}(w)$
        and hence $u, x \in \anc^{\cG'}(w) \cap \des^{\cG'}(w) = \scc^{\cG'}(w)$ and especially
        $u \in \scc^{\cG'}(x)$ and $|\scc^{\cG'}(x)|\geq 2$. Hence the non trivial directed walk $\pi': u \myrightarrow \cdots \myrightarrow x$ that lies entirely in $\scc^{\cG'}(x)$ is an a-inducing path in $\cG'$ and hence $u \notindep_W^{\cG'} x \mid C, D$ according to \cref{prop:a_ind_eqv_s_sep}.
        
        Hence we showed $(\{u\}, \{x\}, C, D) \in \cI_W^\cG$ but $(\{u\}, \{x\}, C, D) \notin \cI_W^{\cG'}$.

        \item[ $\neg ii c)\Rightarrow \neg i)$]
        
        Then there exist $a, b \in V$, $C,D \subseteq V$ such that \\
        $a \indep_W^\cG b \mid C, D$ and $a \indep_W^\cG b \mid C, D\cup \{w\} $ \\
        but $ a \notindep_W^\cG b \mid C, D\cup \{v\} $. 
        
        Define $Z:=C_0 \cup D_1$.
        Then w.l.o.g. there exist a $\sigma$-$Z$-blocked walk $\pi$ in $\tilde{\cG}$ with $\ncoll_{\pi,\mathrm{bl}}\cap Z = \emptyset$ and $\coll_\pi \setminus \anc^{\tilde{\cG}}(Z) = \{v_1\}$. Since $v_1\in \anc^{\tilde{\cG'}}(w_1)$ exist it follows that $\coll_\pi \setminus \anc^{\tilde{\cG'}}(Z\cup \{w_1\}) = \emptyset$ and hence $\pi$ is $\sigma$-$Z\cup \{w_1\}$-open in $\tilde{\cG'}$. Therefore $a \indep_W^{\cG'} b \mid C, D\cup \{w\}$ and $\cI_W^\cG \neq \cI_W^{\cG'}$.
    
        \item[$\neg ii d)\Rightarrow \neg i)$] Follows analogously to $\neg ii c)\Rightarrow \neg i)$.
                
    \end{description}

    \underline{$\mathbf{ii) \Rightarrow i)}$} \hfill \\
    Assume the contradiction: $ii)$ holds but
    there exist $a, b \in V$; $C,D \subseteq V $ such that $(\{a\}, \{b\}, C, D) \in \cI_W^\cG$ but $(\{a\}, \{b\}, C, D) \notin \cI_W^{\cG'}$.
    
    Let $Z \coloneqq C_0 \cup D_1$.
    Then there exists a $\sigma$-$Z$-open walk in $\tilde{\cG'}$:
    $$
    \pi': a_0 \mysim[e'_0] v_{\ell_1}^1 \cdots \mysim[e'_m] b_1
    $$ 
    
    If the edge $e_d' := (v_i, w_j) \in \{(v_0, w_0), (v_0, w_1), (v_1, w_1)\}$ is $e'_d \in \edges_{\pi'}$, then one can replace $e_d'$ by the $\sigma$-$Z$-open walk in $\tilde{\cG}$: $$
    \mu_{ij}: v_i \mysim[\tilde{e}_0] u_{\ell_1}^1 \mysim[\tilde{e}_1] \cdots \circright[\tilde{e}_n] w_j
    $$
    $\mu_{ij}$ exist since it exist an a-inducing walk from $v$ to $w$ in $\cG$.
    
    The obtained walk 
    $$
    \pi: a_0 \mysim[e'_0] \cdots \mysim[e'_{d-1}] \underbrace{v_i \mysim[\tilde{e}_0] u_{\ell_1}^1 \mysim[\tilde{e}_1] \cdots \circright[\tilde{e}_n] w_j}_{\mu_{ij}}  \mysim[e'_{d+1}] \cdots \mysim[e'_m] b_1
    $$
    is then a walk in $\tilde{\cG}$.
    
    Let $\cK_\pi := \{v_{\ell_k}^k \in \coll_\pi \setminus \{v_0, v_1\} : v_{\ell_k}^k \in \anc^{\tilde{\cG}}(\{v_0, v_1\}) \setminus \anc^{\tilde{\cG}}(Z) \cap \anc^{\tilde{\cG'}}(Z)\}$
    
    Let $\cU_\pi := \{v_{\ell_u}^u \in \ncoll_{\pi,\mathrm{bl}}^{\tilde{\cG}} \cap\, Z\}$  
    
    Let $\hat{\cU}_\pi := \{v_{\ell_u}^u \in \nodes_{\pi}: v^u \in \anc^\cG(v) \cap \des^\cG(w)\}$ 

    Note that $\cU_\pi \subseteq \hat{\cU}_\pi \subseteq \scc^{\tilde{\cG'}}(\{v_0, v_1, w_1, w_0\})$ %

    Also note that $\cK_\pi$ is the set of all colliders on $\pi$ that are $\sigma$-$Z$-blocking $\pi$, except for possibly $v_0$ and $v_1$. They are not in $\anc^{\tilde{\cG}}(Z)$, but they are in $\anc^{\tilde{\cG'}}(Z)$ since they did not $\sigma$-$Z$-block $\pi'$.
    
    Since $\pi'$ and $\mu_{ij}$ are $\sigma$-$Z$-open, $\pi$ is $\sigma$-$Z$-open if $v_i, w_j \in \nodes_\pi$ do not $\sigma$-$Z$-block $\pi$
    and $\cU_\pi = \cK_\pi = \emptyset$ (we show that such $\pi$ exist).\\

    \underline{Let us assume it exist an edge $e_d' = (v_i, w_j) \in \edges_{\pi'}$.}

    Since $\mu_{ij}$ is $\sigma$-$Z$-open it holds $\cK_{\mu_{ij}} = \cU_{\mu_{ij}} \setminus \operatorname{endnodes}_{\mu_{ij}} = \emptyset$.
    Since $\mu_{ij}$ is into $w_j$, as was $e_d'$, and $\des^\cG(w) = \des^{\cG'}(w)$, because $w_j$ was not $\sigma$-$Z$-blocking $\pi'$,
    it also does not block $\pi$ unless $w_j \in \cU_\pi$. Therefore $w_j \in \cU_\pi$ is not considered explicitly here but just in general later.
    
    Let us consider if $v_i$ can $\sigma$-$Z$-block $\pi$.
    Note that $v_i \in \ncoll_\pi$.
    
    Let us consider all cases:
    \begin{enumerate}[label=\arabic*), ref=\arabic*]
    
        \item \underline{$v_i \in Z$.}
        Then $v_i \in \ncoll_{\pi'}^{\cG'}$ is unblockable, hence $i = j$, $v \in \scc^{\cG'}(w)$, $w \in \anc^{\cG}(v)$ and $v_i, w_j \in \anc^{\tilde{\cG}}(Z)$
        
        \begin{enumerate}[label=\theenumi.\arabic*) 
        ]

            \item \underline{$v \in \scc^\cG(w)$.}
            Then let $\mu_{ij}: v_i \myrightarrow \cdots \myrightarrow w_j$ be a directed path in $\scc^{\tilde{\cG}}(w)$.
            Then $\mu_{ij}$ is $\sigma$-inducing, $v_i \in \ncoll_\pi$ is unblockable, $\cK_\pi = \cU_\pi = \emptyset$ and $w_j$ does not block $\pi$.
            
            Hence $\pi$ is $\sigma$-$Z$-open and $a_0 \notindep_\sigma^{\tilde{\cG}} b_1 \mid Z$, a contradiction to our assumption.
            
            \item \underline{$v \notin \scc^\cG(w)$.}
            Then $v \notin \anc^\cG(w)$ and since there exist an a-inducing walk from $v$ to $w$ in $\cG$, by \cref{prop:s-ind-walks,lem:s_ind_LDG} it follows that  there exist a $\sigma$-$Z$-open walk $\mu_{ij}: v_i \circleft \cdots \circright w_j$ in $\tilde{\cG}$ that is into $v_i$ and $w_j$. Hence $v_i, w_j \in \anc^{\tilde{\cG}}(Z)$ do not $\sigma$-$Z$-block $\pi$ due to $\mu_{ij}$.
            Note that since $v_i \in Z$, if $v_i \in \coll_\pi$ it will not $\sigma$-$Z$-block $\pi$.
            If $v_i \in \ncoll_\pi$ then either $v_i$ is unblockable or $v_i = v_i^d \in \cU_\pi$ only
            because of $v_{\ell_{d-1}}^{d-1} \myleftarrow v_i^d$ and $v_{\ell_{d-1}}^{d-1} \in \hat{\cU}_\pi$. This is just a special case
            of $\cU_\pi \neq \emptyset$ which is considered later in general. The same holds for $w_j$.
            Note that then $v_i^d \neq \min(\hat{\cU}_\pi)$ and $\tilde{e}_0 \in \edges_\pi \cap \edges_{\mu_{ij}}$ is into $v_i^d \in \hat{\cU}_\pi$.  
            
            With $\cU_\pi = \emptyset$, as shown later, it follows that $a_0 \notindep_\sigma^{\tilde{\cG}} b_1 \mid Z$,
            a contradiction to our assumption.

        \end{enumerate}
        
        \item \label{case:adding_directed_edges_2} \underline{$v_i \notin Z$.}
        Then $v_i$ does not $\sigma$-$Z$-block $\pi$ unless $v_i \in \coll_\pi \setminus \anc(Z,\{b_1\})$.
        
        Therefore assume there exists no $\pi$ such that $v_i \notin \coll_\pi \setminus \anc(Z,\{b_1\})$. Then $v \notin \scc(w)$, otherwise $v_i \myrightarrow \cdots \myrightarrow w_j$
        that lies entirely in $\scc(w_j)$ (except for $v_i$ if $i=0$, $j=1$) would exist.
        
        Note we thus have that $v_i \notin \cU_\pi$. Hence we have showed to this point that replacing $e'_d = (v_i, w_j)$ does not add nodes to $\cU_\pi$. We leave this case unsolved for now and are
        first showing that $\cU_\pi = \emptyset$.\\
        
    \end{enumerate}

    Note that if $v \in \anc^\cG(w)$ then $\cU_\pi = \cK_\pi = \emptyset$.\\

    \underline{Let us assume $\cU_\pi \neq \emptyset$.} 
    
    Then $v \notin \anc^\cG(w)$ and $|\hat{\cU}_\pi| \geq 2$ since it must be that
    $v_{\ell_t}^t \myrightarrow v_{\ell_{t+1}}^{t+1}$ or $v_{\ell_{t-1}}^{t-1} \myleftarrow v_{\ell_t}^t$ is on $\pi$ and $v_{\ell_t}^t, v_{\ell_{t+1}}^{t+1} \in \hat{\cU}_\pi$ or $v_{\ell_t}^t, v_{\ell_{t-1}}^{t-1} \in \hat{\cU}_\pi$ since $v_{\ell_t}^t \in \cU_\pi$ but $v_{\ell_t}^t \notin \ncoll_{\pi',\mathrm{bl}}$.
    
    Let $v_{\ell_u}^u := \min(\{v_{\ell_u}^u \in \hat{\cU}_\pi : v_{\ell_u}^u \in \cU_\pi \text{ or } v_{\ell_{u+1}}^{u+1} \in \cU_\pi\})$ and $v_{\ell_x}^x := \max(\{v_{\ell_u}^u \in \hat{\cU}_\pi : v_{\ell_u}^u \in \cU_\pi \text{ or } v_{\ell_{u-1}}^{u-1} \in \cU_\pi\})$.
    
    Since $\forall v_{\ell_t}^t \in \hat{\cU}_\pi: v^t \in \anc^\cG(v) \cap \des^\cG(w)$ by condition \cref{cond:potential_parents_b} it follows there exist
    both a-inducing walks from $v^u$ to $v^x$ and from $v^x$ to $v^u$ in $\cG$.
    
    Note that $v_{\ell_{u-1}}^{u-1}, v_{\ell_{x+1}}^{x+1} \notin \hat{\cU}_\pi \supseteq \cU_\pi$.\\
    
    If $v^u \in \scc(v^x)$ we are considering the following cases. In those we assume w.l.o.g.\\
    that if $\ell_u \neq \ell_x$ then $\ell_u = 0$, $\ell_x = 1$. This is valid since the problem is symmetric w.r.t. $v^u$ and $v^x$.
    
    \begin{enumerate}[label=\alph*)]

        \item \underline{$v_{\ell_u}^u \in \cU_\pi \subseteq Z$.}
        Then $e_{u-1} \in \edges_\pi$ is into $v_{\ell_u}^u$ and $v_{\ell_{u-1}}^{u-1} \notin \scc(v_{\ell_u}^u)$.
        Thus there exists the $\sigma$-inducing walk $\gamma: v_{\ell_{u-1}}^{u-1} \circright[\hat{e}_0] v_{\ell_x}^u \myrightarrow \cdots \myrightarrow[\hat{e}_{\hat{n}}] v_{\ell_x}^x$ in $\tilde{\cG}$ that lies entirely in $\scc(v_{\ell_x}^x)$
        for all nodes except $v_{\ell_{u-1}}^{u-1}$. Hence, by replacing the subwalk from $v_{\ell_{u-1}}^{u-1}$ to $v_{\ell_x}^x$
        on $\pi$ with $\gamma$, we obtain the walk 
        $\tilde{\pi}: a_0 \mysim \cdots \mysim v_{\ell_{u-1}}^{u-1} \circright v_{\ell_u}^u \myrightarrow \cdots \myrightarrow[\hat{e}_{\hat{n}}] v_{\ell_x}^x \mysim \cdots \mysim b_1$ in $\cG$.
        
        \item \underline{$v_{\ell_u}^u \notin Z \supseteq \cU_\pi$.}
        Then it exists the $\sigma$-inducing walk $\gamma: v_{\ell_u}^u \myrightarrow[\hat{e}_0] \cdots \myrightarrow[\hat{e}_{\hat{n}}] v_{\ell_x}^x$ in $\cG$ that lies entirely in $\scc(v_{\ell_x}^x)$
        except for $v_{\ell_u}^u$ if $\ell_u \neq \ell_x$. Hence by replacing the subwalk from $v_{\ell_{u}}^{u}$ to $v_{\ell_x}^x$
        on $\pi$ with $\gamma$ we obtain the walk $\tilde{\pi}: a_0 \mysim \cdots \mysim v_{\ell_u}^u \myrightarrow[\hat{e}_0] \cdots \myrightarrow[\hat{e}_{\hat{n}}] v_{\ell_x}^x \mysim \cdots \mysim b_1$ in $\cG$.
        
    \end{enumerate}
    
    Note that if $v_{\ell_x}^x \in \cU_\pi \subseteq Z$ then $e_x \in \edges_\pi$ is into $v_{\ell_x}^x$ and $v_{\ell_x}^x \in \coll_{\tilde{\pi}} \cap \anc(Z)$.
    If $v_{\ell_x}^x \notin \cU_\pi$ then $e_{x-1} \in \edges_\pi$ is into $v_{\ell_x}^x$ as is $\hat{e}_{\hat{n}} \in \edges_{\tilde{\pi}}$.
    
    Since $\cU_{\gamma} \setminus \operatorname{endnodes}_{\gamma} = \emptyset$ we have $\cU_{\tilde{\pi}} = \emptyset$. Note that $K_{\gamma} = \emptyset$ and hence $K_{\pi} = K_{\tilde{\pi}}$.\\
    
    If $v^u \notin \scc(v^x)$ it follows by \cref{prop:s-ind-walks,lem:s_ind_LDG} that it exist an $\sigma$-inducing walk $\gamma: v_{\ell_u}^u \circleft[\hat{e}_0] \cdots \circright[\hat{e}_n] v_{\ell_x}^x$ that is into $v_{\ell_u}^u$ and $v_{\ell_x}^x$.
    
    We then replace the subwalk from $v_{\ell_x}^u$ to $v_{\ell_x}^x$ on $\pi$ by $\gamma$ and obtain the walk $\tilde{\pi}: a_0 \mysim \cdots \mysim v_{\ell_u}^u \circleft[\hat{e}_0] \cdots \circright[\hat{e}_n] v_{\ell_x}^x \mysim \cdots \mysim b_1$ in $\cG$.
    
    Hence if $v_{\ell_u}^u \in \cU_{\pi} \subseteq Z$ then $e_{u-1} \in \edges_{\pi}$ is into $v_{\ell_u}^u$ and $v_{\ell_u}^u \in \coll_{\tilde{\pi}} \cap \anc(Z)$. If $v_{\ell_u}^u \notin \cU_{\pi}$ then $e_u \in \edges_{\pi}$ is into $v_{\ell_u}^u$ as is $\hat{e}_0 \in \edges_{\tilde{\pi}}$. Analogously the same holds for $v_{\ell_x}^x$.
    
    Since $\cU_{\gamma}\setminus \operatorname{endnodes}_{\gamma} = \emptyset$ we have $\cU_{\tilde{\pi}} = \emptyset$. Note that $\cK_{\gamma} = \emptyset$ and hence $\cK_{\pi} = \cK_{\tilde{\pi}}$.
    
    Hence $\tilde{\pi}$ has the same properties as $\pi$ except for $\cU_{\tilde{\pi}} = \emptyset$.\\
    
    Therefore, to keep notation simple, we are in the following assuming $\cU_{\pi} = \emptyset$ since we have shown that we can easily construct such $\pi$.
    
    Note that \ref{case:adding_directed_edges_2} $v \notin Z$ when replacing $e_d' = (v_i, w_j) \in \edges_{\pi'}$ does not create any difficulties in the argument for $\cU_{\pi} = \emptyset$, since in case $v_i = v_{\ell_u}^u$ and $w_j = v_{\ell_x}^x$ we are in case $v^u \notin \scc(v^x)$ and hence to show $\cU_{\tilde{\pi}} = \emptyset$ it holds that $\gamma = \mu_{ij}$ exist.\\

    \underline{Next assume $\cK_{\pi} \neq \emptyset$.}

    Let $v_{\ell_k}^k \in \cK_{\pi}$. Then by definition $v_{\ell_k}^k \in (\anc^{\tilde{\cG}}(\{v_0, v_1\}) \setminus \anc^{\tilde{\cG}}(Z)) \cap \anc^{\tilde{\cG}'}(Z)$
    and hence it must be that $\{w_0, w_1\} \cap \anc^{\tilde{\cG}}(Z) \neq \emptyset$.
    
    Since $v_{\ell_k}^k \in \anc^{\tilde{\cG}}(\{v_0, v_1\}) = \anc^{\tilde{\cG}}(v_1)$ we can then insert the walk $v_{\ell_k}^k \myrightarrow \cdots \myrightarrow v_1 \myleftarrow \cdots \myleftarrow v_{\ell_k}^k$ that does not intersect $Z$ into $\pi$ and obtain
    $\tilde{\pi}: a_0 \mysim \cdots \circright[e'_{k-1}] v_{\ell_k}^k \myrightarrow \cdots \myrightarrow v_1 \myleftarrow \cdots \myleftarrow v_{\ell_k}^k \circleft[e'_k] \cdots \mysim b_1$.      
    
    Then $v_{\ell_k}^k \notin \coll_{\tilde{\pi}}$. We can do so for all $v_{\ell_k}^k \in \cK_{\pi}$ and hence $\cK_{\tilde{\pi}} = \emptyset$.
    Note that potentially $v_0 \in \coll_{\tilde{\pi}} \setminus \anc(Z)$ due to the removing of $e'_d \in \edges_{\pi'}$ in \ref{case:adding_directed_edges_2} when constructing ${\pi}$ from $\pi'$.
    
    If so, then $e'_t, e'_{t-1} \in \edges_\pi$ are into $v_0 =: v^t_0$.
    
    If $v_{\ell_{t-1}}^{t-1} \in \scc(v_0)$ then $v_{\ell_{t-1}}^{t-1} \notin Z$, otherwise $v_0 \in \anc(Z)$ a contradiction.
    If $v_{\ell_{t-1}}^{t-1} \notin \scc(v_0)$ then either
    $v_{\ell_{t-1}}^{t-1} \notin Z$ or $e'_{t-1} \in \tilde{\cE}_{bi}$. Analogously the same holds for $v_{\ell_{t+1}}^{t+1}$ and $e'_t$.
    
    Hence if $v_0 \in \coll_{\tilde{\pi}} \setminus \anc(Z)$, we can just replace it  with $v_1$ on $\tilde{\pi}$.
    We just showed the therefore needed edges exist in $\tilde{\cG}$ since $e'_{t}, e'_{t-1}$ are into
    $v_0$ and if $v_{\ell_{t-1}}^{t-1}, v_{\ell_{t+1}}^{t+1}$ did not block $\pi$ then they still dont when
    replacing $v_0$ with $v_1$ on $\tilde{\pi}$.
    Note that if $v_1 \in \coll_{\tilde{\pi}}$ then $v_1 \notin \ncoll_{\tilde{\pi}}$ since otherwise we could
    shorten $\tilde{\pi}$ by cutting out the subwalk on $\tilde{\pi}$ between $v_i^{min}$ and $v_i^{max}$ with
    \begin{equation}
    v_1^{min} \coloneqq \min\{v_1^i \in \nodes_\pi: v_1^i = v_1\}
    \end{equation}
    
    \begin{equation}
    v_1^{\max} \coloneqq \max\{v_1^i \in \nodes_\pi: v_1^i = v_1\}.
    \end{equation}

    Therefore we have that
    \begin{itemize}
        \item $\{v_1\} = \coll_{\tilde{\pi}} \cap \anc(Z)$
        
        \item $\ncoll_{\tilde{\pi},\mathrm{bl}} \cap Z = \emptyset$
        
    \end{itemize}
    
    Hence $a_0 \notindep_\sigma^{\tilde{\cG}} b_1 \mid Z \cup \{v_1\}$
    and since we assume $a_0 \indep_\sigma^{\tilde{\cG}} b_1 \mid Z$ it follows by condition \cref{cond:potential_parents_c} that $a_0 \notindep_\sigma^{\tilde{\cG}} b_1 \mid Z \cup \{w_1\}$. From this it follows that there must exist a $\sigma$-$Z$-blocked walk $\tau$ from $a_0$ to $b_1$ in $\tilde{\cG}$ with $\{w_1\}=\coll_\tau \setminus \anc^{\tilde{\cG}}(Z)$ and $\ncoll_{\tau,\mathrm{bl}}\cap Z = \emptyset$ and hence $w\notin \anc^\cG(D,\{b\})$. Using this it furthermore follows that $v_1 \notin \anc^{\tilde{\cG'}}(Z)$ and hence, by definition of $\cK_\pi$ (and $v_1 \notin \anc^{\tilde{\cG}}(w_0)$), it must be that $\cK_\pi \subseteq V_0$.
    
        Since $\cK_\pi \subseteq V_0$ it holds $\forall v_{\ell_k}^k \in \cK_\pi$: $\ell_k = 0$ and $v_0^k \in \anc(v_0) \setminus \anc(Z)$.
        
        Hence we can insert the walk $v_0^k \myrightarrow \cdots \myrightarrow v_0 \myleftarrow \cdots \myleftarrow v_0^k$
        that does not intersect $Z$ into $\pi$ and obtain
        $\tilde{\pi}: a_0 \mysim \cdots \circright[e'_{k-1}] v_0^k \myrightarrow \cdots \myrightarrow v_0 \myleftarrow \cdots \myleftarrow v_0^k \circleft[e'_k] \cdots \mysim b_1$
        
        Hence w.l.o.g. we can assume that $\cK_{\tilde{\pi}} = \emptyset$
        and
        
        \begin{itemize}
            \item $\coll_{\tilde{\pi}} \setminus \anc(Z) \subseteq \{v_0, v_1\}$

            \item $\ncoll_{\tilde{\pi},\mathrm{bl}} \cap Z = \emptyset$
            
        \end{itemize}

    Note that $v_1 \in \coll_{\tilde{\pi}} \setminus \anc(Z)$ if and only if it comes from
    removing $e'_d = (v_1, w_1) \in \edges_{\pi'}$ during the construction of $\pi$ from $\pi'$.
    Thus if $v_1 \in \coll_{\tilde{\pi}} \setminus \anc(Z)$ then it must be that $e'_{d+1} \in \edges_{\pi'} \cap \edges_{{\pi}}$
    is out of $w_1$, since $e_d' \in \edges_{\pi'}$ was into $w_1$, $w_1 \notin \anc^{\tilde{\cG'}}(Z)$ and $\pi'$ was $\sigma$-$Z$-open.

    But then, since $w\notin \anc^\cG(D,\{b\})$, it must be that there exist $v_1^k \in \coll_{{\pi}}\setminus \anc^{\tilde{\cG}}(Z)$, and since $\pi'$ was $\sigma$-$Z$-open and hence it must be that $v_1^k\in\cK_\pi$, a contradiction to $\cK_\pi \subseteq V_0$.

    Hence we have
    
    \begin{itemize}
        \item $\{v_0\} = \coll_{\tilde{\pi}} \setminus \anc(Z)$
        
        \item $\ncoll_{\tilde{\pi},\mathrm{bl}} \cap Z = \emptyset$
        
    \end{itemize}
    
    Note again that w.l.o.g. $v_0 \notin \ncoll_{\tilde{\pi}}$
    with the same argument as before.
    
    Therefore we have that $a_0 \indep_\sigma^{\tilde{\cG}} b_1 \mid Z$ but $a_0 \notindep_\sigma^{\tilde{\cG}} b_1 \mid Z \cup \{v_0\}$
    and hence by \cref{cond:potential_parents_d} it follows that $a_0 \notindep_\sigma^{\tilde{\cG}} b_1 \mid Z \cup \{w_0\}$. From this it follows that there must exist a $\sigma$-$Z$-blocked walk $\tau$ from $a_0$ to $b_1$ in $\tilde{\cG}$ with $\{w_0\}=\coll_\tau \setminus \anc^{\tilde{\cG}}(Z)$ and $\ncoll_{\tau,\mathrm{bl}}\cap Z = \emptyset$ and hence $w\notin \anc^\cG(C,D)$. 
    
    Thus it follows that $\anc^{\cG}(v)\cap \anc^{\cG}(C,D) = \anc^{\cG'}(v)\cap \anc^{\cG'}(C,D)$
    and hence by definition $\cK_\pi = \emptyset$.\\

    By now we found that $\cK_\pi = \cU_\pi = \emptyset$ and that we can replace
    every potential edge $e_d' = (v_i, w_j) \in \{(v_0, w_0), (v_0, w_1), (v_1, w_1)\} \subseteq \tilde{\cE}_d'$ on $\pi'$ with
    a $\sigma$-$Z$-open walk $\mu_{ij}$ and obtain the walk $\pi$ in $\tilde{\cG}$.
    
    We also showed that using $\mu_{ij}$ does not lead to $w_j$ blocking $\pi$
    and more generally that $\pi$ is $\sigma$-$Z$-open for all cases except
    for case \ref{case:adding_directed_edges_2} of replacing $e_d'$.
    
    \underline{Therefore, lets consider case \ref{case:adding_directed_edges_2}.}
    Then we have that $v_i \in \coll_{\pi} \setminus \anc(Z,\{b_1\})$, $\ncoll_{\pi,\mathrm{bl}} \cap Z = \emptyset$, $\coll_{\pi} \setminus \anc(Z) \subseteq \{v_0, v_1\}$.

    Assume $v_1 \in \coll_{\pi} \setminus \anc(Z,\{b_1\})$.
    Then just like before, we can w.l.o.g. assume $\coll_{\pi} \setminus \anc(Z) = \{v_1\}$. Hence $a_0 \notindep_\sigma^{\tilde{\cG}} b_1 \mid Z \cup \{v_1\}$ and since we assume $a_0 \indep_\sigma^{\tilde{\cG}} b_1 \mid Z$ it follows by condition \cref{cond:potential_parents_c} that $a_0 \notindep_\sigma^{\tilde{\cG}} b_1 \mid Z \cup \{w_1\}$. From this it follows that there must exist a $\sigma$-$Z$-blocked walk $\tau$ from $a_0$ to $b_1$ in $\tilde{\cG}$ with $\{w_1\}=\coll_\tau \setminus \anc^{\tilde{\cG}}(Z)$ and $\ncoll_{\tau,\mathrm{bl}}\cap Z = \emptyset$ and hence $w\notin \anc^\cG(D,\{b\})$. As before, since $e_d'$ was into $w_1$, it follows that there must exist a collider $v_1^k \in \coll_{{\pi}}\setminus \anc^{\tilde{\cG}}(Z)$, and since $\pi'$ was $\sigma$-$Z$-open and hence it must be that $v_1^k\in\cK_\pi$, a contradiction to $\cK_\pi = \emptyset$.
    
    Therefore $\coll_{\pi} \setminus \anc(Z) = \{v_0\}$ and, as before, we have that $a_0 \indep_\sigma^{\tilde{\cG}} b_1 \mid Z$ but $a_0 \notindep_\sigma^{\tilde{\cG}} b_1 \mid Z \cup \{v_0\}$
    and hence by \cref{cond:potential_parents_d} it follows that $a_0 \notindep_\sigma^{\tilde{\cG}} b_1 \mid Z \cup \{w_0\}$. From this it follows that there must exist a $\sigma$-$Z$-blocked walk $\tau$ from $a_0$ to $b_1$ in $\tilde{\cG}$ with $\{w_0\}=\coll_\tau \setminus \anc^{\tilde{\cG}}(Z)$ and $\ncoll_{\tau,\mathrm{bl}}\cap Z = \emptyset$ and hence $w\notin \anc^\cG(C,D,\{a,b\})$. As before, since $e_d'$ was into $w_0$, it follows that there must exist a collider $v_{\ell_k}^k \in \coll_{{\pi}}\setminus \anc^{\tilde{\cG}}(Z)$, and since $\pi'$ was $\sigma$-$Z$-open and hence it must be that $v_{\ell_k}^k\in\cK_\pi$, a contradiction to $\cK_\pi = \emptyset$.
    
Therefore we finally showed that all cases lead to a contradiction of
our assumption. This concludes our proof.
\end{proof}

\begin{proof}[Proof of \cref{prop:a_ind_pot_sib,}]
    This follows directly from \cref{prop:s-ind-walks,lem:s-ind_inseparability}.
\end{proof}

\begin{proof}[Proof of \cref{lem:a_in_A}]
Let $\cG = (V, \cE_d, \cE_{bi})$ be a DMG, $v,w \in V$ and $v \notin \anc(w)$. Let $\cG' = (V,\cE_d' , \cE_{bi})$, $\cE_d':=\cE_d \cup \{(v,w)\}$.
If $v$ is a potential parent of $w$ in $\cG$, then by \cref{prop:adding_directed_edges} it follows that $\cI_W^{\cG'} = \cI_W^\cG$. Furthermore, from \cref{def:potential_parents} it follows  that there exists an a-inducing path from $v$ to $w$ in $\cG$.
Denote $A := \{a \in \anc^\cG(v) \setminus \anc^\cG(w): \text{there exists an a-inducing path from } a \text{ to } w \text{ in } \cG\}$.
Note that $v \in A$. 

Assume the contrary: There exists a $b \in \anc^\cG(v) \setminus \anc^\cG(w)$, $C,D \subseteq V$ such that $b \indep_W^\cG w \mid C, D$. %
Let $b \in \pa^\cG(A)$. Such $b$ exists since $v \in A$ and $b \in \anc^\cG(v) \setminus \anc^\cG(w)$ and hence there exists
the directed walk $\tau: b \myrightarrow \cdots \myrightarrow v$ and a last node $b'$ on $\tau$ with $b' \in \pa^\cG(A) \cap (\anc^\cG(v) \setminus \anc^\cG(w))$.
Let $a \in A$ such that $b \in \pa^\cG(a)$. Then by using \cref{prop:s-ind-walks} it follows that there exists the s-inducing walk $\pi: a \circleft[e_{0}] \cdots \circright[e_{n}] w$
in $\cG$.
By concatenating the walk $b \myrightarrow a$ with $\pi$, we obtain the walk $\mu: b \myrightarrow a \circleft[e_0] \cdots  \circright[e_n] w$ in $\cG$. Since $b \notin A$, $\mu$ cannot be an a-inducing walk in $\cG$.
Note that $\coll_\mu \subseteq \anc^\cG(a,w)$, $\ncoll_{\mu,\mathrm{bl}} \setminus \text{endnodes}_\mu = \emptyset$.
Since $v \in \anc^{\cG'}(w)$ and $a \in \anc^{\cG'}(v)$ it follows that $\coll_\mu \subseteq \anc^{\cG'}(w)$,
$\ncoll_{\mu,\mathrm{bl}} \setminus \text{endnodes}_\mu = \emptyset$ and hence $\mu$ is an a-inducing walk in $\cG'$ (the last edge
before entering $\scc(w)$ is into $\scc(w)$ by construction of $\mu$ from $\pi$).
Hence we have $b \notindep_W^{\cG'} w \mid C, D$, a contradiction to $\cI_W^{\cG'} = \cI_W^\cG$.
Thus for every $a \in \anc^\cG(v) \setminus \anc^\cG(w) = A$ it holds that there exists an a-inducing path from $v$ to $w$ in $\cG$.
By \cref{prop:s-ind-walks} it follows the existence of an s-inducing path and from \cref{lem:s-ind_inseparability,def:potential_siblings} it follows that $a$ is a potential sibling of $w$.

\end{proof}

\subsection{Proofs on the Greatest Markov Equivalent DMGs}\label{app:proofs_greatest}

\begin{proof}[Proof of \cref{prop:pot_pa_two_graphs}]
    Let $\cG^1 = (V, \cE^1_d, \cE^1_{bi})$ and  $\cG^2 = (V,\cE^2_d , \cE^2_{bi})$ be two DMGs over a common node set $V$, $v,w \in V$, $\cG^1 \in [\cG^2]_W$. Then if  $v$ is a potential parent of $w$ in $\cG^2$ then there exist a Markov equivalent graph that contains the edge $v\myrightarrow w$. We therefore simply assume w.l.o.g. that $(v,w)\in \cE^2_d$. Furthermore, w.l.o.g. we assume that $\cE_{bi}^1=\cE_{bi}^2=\hat{\cE}_{bi}$ where for every $a,b\in V$ it holds $[(a,b)] \in \hat{\cE}_{bi}$ if and only if $a$ and $b$ are potential siblings in $\cI_W^{\cG^1}$. Note that otherwise we could just add them without altering the independence model or any ancestral relations and hence all conditions for $v$ being a potential parent of $w$ in $\cG^1$ are not affected. We start by restating those conditions for an arbitrary DMG $\cG$ and follow by proving that they hold in $\cG^1$:
    \begin{enumerate}[label=\alph*)]
    \item  $v\in u (w, \cI_W^\cG)$ %
    
    \item $\forall u, x \in \anc^\cG(v) \cap \des^\cG(w)$:
    $u\in u (x, \cI_W^\cG)$

    \item $\forall a, b \in V$, $C,D \subseteq V$, with $a \indep_W^\cG b \mid C, D$ and $a \indep_W^\cG b \mid C, D \cup \{w\}: a \indep_W^\cG b \mid C, D \cup \{v\}$

    \item $\forall a, b \in V$, $C,D \subseteq V$, with $a \indep_W^\cG b \mid C, D$ and $a \indep_W^\cG b \mid C\cup \{w\}, D : a \indep_W^\cG b \mid C\cup \{v\}, D $

\end{enumerate}
Since $(v,w)\in \cE^2_d$ by \cref{lem:pa_are_pot_pa} it follows that these conditions hold in $\cG^2$. Since $\cG^1 \in [\cG^2]_W$, conditions a),c),d) trivially hold in $\cG^1$ as well. Hence it is only left to proof that condition b) holds in $\cG^1$.
Note that this would trivially be the case if $v\in \anc^{\cG^1}(w)$, since then $v,u,x\in \scc^{\cG^1}(w)$. Therefore assume $v\notin \anc^{\cG^1}(w)$. Then, since $v\in u(w,\cI_W^{\cG^1})=u(w,\cI_W^{\cG^2})$, it follows by \cref{prop:s-ind-walks} that there exist an s-inducing walk between $v$ and $w$ and hence $w\in u(v,\cI_W^{\cG^1})$. Therefore they are potential siblings in $\cI_W^{\cG^2}$ and thus $[(v,w)]\in \hat{\cE}_{bi} = \cE^1_{bi} = \cE^2_{bi}$. 
Now assume there exists a node $v^u \in \anc^{\cG^1}(v) \cap \des^{\cG^1}(w)$ (otherwise b) trivially holds). Then there exist a directed path $\pi:w=v^0\myrightarrow[e_0] v^1 \myrightarrow[e_1] \cdots \myrightarrow[e_{u-1}] v^u \myrightarrow[e_u] \cdots \myrightarrow[e_n] v^{n+1}=v$ in $\cG^1$. %
We first show that for every $v^i\in \nodes_\pi $ the following are equivalent:
\begin{enumerate}[label=\roman*)]
    \item there exist an a-inducing path from $v^i$ to $v$ in $\cG^1$ (and hence equivalently in $\cG^2$)
    \item there exist an a-inducing path from $v^i$ to $w$ in $\cG^1$ (and hence equivalently in $\cG^2$)
    \item $[(v^i,w)]\in \hat{\cE}_{bi}$
    \item $[(v^i,v)]\in \hat{\cE}_{bi}$
\end{enumerate}
To see this assume i) to hold. Then there exists an a-inducing walk $\mu: v^i\mysim \cdots \circright v$ in $\cG^2$ and by concatenating $\mu$ with $v\bidiedge w$ we obtain the a-inducing walk $\mu': v^i\mysim \cdots \circright v\bidiedge w$ in $\cG^2$ (note that $v\in \anc^{\cG^2}(w)$ and hence $\coll_{\mu'} \subseteq \anc^{\cG^2}(v^i,v)\subseteq \anc^{\cG^2}(v^i,w)$) and thus it follows that ii) holds.

From ii), and since $w\in \anc^{\cG^1}(v^i)$, it follows that either $w \in \scc^{\cG^1}(v^i)$ or, using \cref{prop:s-ind-walks}, there exist an s-inducing walk between $v$ and $w$ in $\cG^1$. In both cases it follows that $[(v^i,w)] \in \hat{\cE}_{bi}$ and hence iii) holds. 

Then from iii) it follows the existence of the s-inducing walk $v^i \circleft w \circright v$ in $\cG^1$ (note that $w \in \anc^{\cG^1}(v)$) and hence $[(v^i,v)] \in \hat{\cE}_{bi}$ and hence iv) holds.

From iv) it trivially follows i) which completes the cycle.

Now assume $[(v^u,w)] \notin \hat{\cE}_{bi}$. Then there must exist a $j := \max\{j \in [n] : [(v^j,w)] \notin \hat{\cE}_{bi}\}$ (note $u \in [n]$). Then $[(v^{j+1},w)] \in \hat{\cE}_{bi}$ and hence, as we just showed, $[(v^{j+1},v)] \in \tilde{\cE}_{bi}$. Therefore there exist the  walk $v^j \myrightarrow v^{j+1} \bidiedge v$ in $\cG^1$ that is a-inducing since $v^{j+1} \in \anc^{\cG^1}(v)$. We just showed that this is equivalent to $[(v^j,w)] \in \hat{\cE}_{bi}$, a contradiction to the definition of $j$.

Hence it must be that $[(v^u,w)] \in \hat{\cE}_{bi}$ for every $u \in \anc^{\cG^1}(v) \cap \des^{\cG^1}(w)$.

Let $u,x \in \anc^{\cG^1}(v) \cap \des^{\cG^1}(w)$. Then it exist the a-inducing walk $u \bidiedge w \bidiedge x$ (note that $w \in \anc^{\cG^1}(u,x)$) in $\cG^1$. This concludes the proof.
\end{proof}

\begin{proof}[Proof of \cref{thr:greatest_dmg}]
    Let $\cG=(V,\cE_d ,\cE_{bi})$ be a DMG and $\hat{\cG}=(V,\hat{\cE}_d ,\hat{\cE}_{bi})$ be the DMG where $(v,w) \in \hat{\cE}_d$ if and only if $v$ is a potential parent of $w$ in $\cG$, and $[(v,w)] \in \hat{\cE}_{bi}$ if and only if $v$ and $w$ are potential siblings in $\cI_W^\cG$.

    Then by \cref{lem:sib_are_pot_sib} it follows that if there exist a DMG $\cG'=(V,\cE'_d,\cE'_{bi})$ with $\cG'\in [\cG]_W$ and $[(v,w)] \in \cE'_{bi}$ then $[(v,w)] \in \hat{\cE}_{bi}$. From \cref{prop:pot_pa_two_graphs} the same follows for directed edges and therefore there exist no DMG in $[\cG]_W$ that contains an edge that is not also existent in $\hat{\cG}$, i.e. $\hat{\cG}$ is a supergraph of all graphs in $[\cG]_W$. Next \cref{prop:adding_bidi_edges,prop:adding_directed_edges} proof that when we sequentially add all the edges in $(\hat{\cE_d} \cup \hat{\cE_{bi}})\setminus (\cE_d \cup \cE_{bi})$ to $\cG$ (in a finite sequence of DMGs $\cG_0,G_1,...,G_m$ with $\cG_0=G$, $\cG_m=\hat{\cG}$) it follows that $\cG_i \in [{\cG}]_W$ for all $i\in \{0,1,...,m\}$ and hence also $\hat{\cG}\in [\cG]_W$ which concludes the proof.
\end{proof}

\begin{proof}[Proof of \cref{cor:greatest_char}]
By the proof of \cref{thr:greatest_dmg}, the greatest element $\hat{\cG}$ of $[\cG]_W$ is the DMG whose directed edges are exactly the pairs $(v,w)$ such that $v$ is a potential parent of $w$ in $\cG$, and whose bidirected edges are exactly the pairs of potential siblings in $\cI_W^\cG$; moreover, $\cG \subseteq \hat{\cG}$. Since the greatest element is unique, $\cG$ is the greatest element of $[\cG]_W$ if and only if $\cG = \hat{\cG}$, i.e., if and only if every potential parent of a node is a parent of it and any two potential siblings are siblings in $\cG$.
\end{proof}

\section{A Markov Property for an SDE Model}
\subsection{Precise Formulation of the SDE Model}\label{app:sde_assumptions}

Given a filtered probability space $(\Omega, \mathcal{F}, \mathbb{F}, P)$ with filtration $\mathbb{F}=(\mathcal{F}_t)_{t\in I}$ satisfying the usual conditions over $I=[0,s+h]$
where $h \in \mathbb{R}_{>0}$ fixed and $s \in \mathbb{R}_{>0}$ the fixed time point separating the past and the future. Consider the multivariate stochastic process $X = (X^1, \ldots, X^d)$ with $X^j = (X_t^j)_{t \in I}$
for $j \in [d]$ evolving according to a stationary, possibly path-dependent SDE-model of the form 
    \begin{align}\label{eq:integral_SDE_equation} 
        X_t^{k} = \xi^{k} + \int_{0}^{t} \mu^{k} (X_{[0,r]}) dr + \int_{0}^{t} \sigma^{k} (X_{[0,r]}) dW^{k}_r,
    \end{align}
for $t \in I$, $k \in [d]$, with $X_t^k \in \mathbb{R}^{n_k}$, $n_k \in \mathbb{N}_{>0}$, Brownian motions $W_t^k \in \mathbb{R}^{m_k}$ w.r.t. $\mathbb{F}$, $m_k \in \mathbb{N}_{>0}$, initial conditions $\xi^k$ $\mathbb{R}^{n_k}$-valued, $\mathcal{F}_0$-measurable random variables and drift- and diffusion
functionals $\mu^k : \mathbb{D}(I: \mathbb{R}^n) \rightarrow \mathbb{R}^{n_k}$, $\sigma^k : \mathbb{D}(I: \mathbb{R}^n) \rightarrow \mathbb{R}^{n_k \times m_k}$ that may depend on the full solution
history up to time $t$ where $n = \sum_{k=1}^{d} n_k$. More precisely we assume the following assumptions on our SDE system: 
\begin{enumerate}
\item[(A1)] \defined{Exogenous noise independence:} The family of Brownian motions and initialization values  $W^1,\dots,W^d, \xi^1, \ldots, \xi^d$  are mutually independent.
In particular, for each $s,h>0$, the increments $(W^k_{s+t}-W^k_s)_{t\in[0,h]}$ are independent of $\sigma(X_{[0,s]},W_{[0,s]})$.
\item[(A2)] \defined{Non-anticipativity:} For each $k \in [d]$, $\mu^k: I \times \mathbb{D}(I: \mathbb{R}^n) \rightarrow \mathbb{R}^{d_k}$ and $\sigma^k: I \times \mathbb{D}(I: \mathbb{R}^n)\rightarrow \mathbb{R}^{d_k \times m_k}$ are Borel maps on path space and satisfy: 
\begin{enumerate}
    \item $t \mapsto \mu^{k}(t,x) \in \mathbb{D}(I: \mathbb{R}^{d_k})$, $t \mapsto \sigma^{k}(t,x) \in \mathbb{D}(I: \mathbb{R}^{d_k \times m_k})$ for $x \in \mathbb{D}(I: \mathbb{R}^n)$
    \item $\mathcal{B}(I) \otimes \mathcal{B}(\mathcal{J}_1)-\mathcal{B}(\mathbb{R}^{d_k})$ resp. $\mathcal{B}(I) \otimes \mathcal{B}(\mathcal{J}_1)-\mathcal{B}(\mathbb{R}^{d_k \times m_k})$ jointly measurable where $\mathcal{J}_1$ denotes the Skorokhod topology on the space $\mathbb{D}(I: \mathbb{R}^n)$ 
    of cadlag functions from $I$ to $\mathbb{R}^n$.
    \item $\mu^k(t, x)=\mu^k(t, x_{\cdot \wedge t}),\qquad \sigma^k(t, x)=\sigma^k(t, x_{\cdot \wedge t})$ where $x_{\cdot \wedge t}: I \rightarrow \mathbb{R}^{n}$ is $x_{\cdot \wedge t}(s) = x(s \wedge t)$, meaning the functions at time $t$ depend only on the past.
\end{enumerate}
\item[(A3)] \defined{Functional Lipschitz:} there exists an $L>0$ such that for all $t\in I$ and $x,y\in \mathbb{D}(I: \mathbb{R}^n)$:
\begin{align*}
\|\mu^k(t,x)-\mu^k(t,y)\|+\|\sigma^k(t,x)-\sigma^k(t,y)\| \le L \,\sup_{u\le t}\|x(u)-y(u)\|;
\end{align*}
\item[(A4)] \defined{Linear growth:} there exists an $L>0$ s.t. for all $t\in I$ and $x\in \mathbb{D}(I:  \mathbb{R}^n)$,
\begin{align*}
    \|\mu^k(t,x)\|+\|\sigma^k(t,x)\| \leq L \Bigl(1+\sup_{u\le t}\|x(u)\|\Bigr).
\end{align*}
\end{enumerate}

\begin{remark}[Existence of a solution map] An adapted continuous process $X$ solves \eqref{eq:integral_SDE_equation}
if its integrands are stochastically integrable and, on an event of probability one, all coordinate
identities hold for every $t \in I$. The standard assumption (A1)-(A4) to ensure existence and uniqueness of solutions of SDEs, see \citet{Protter1992}. 
\end{remark}

\subsection{Dependency Graph and Markov Property of the SDE Model} 
The coordinate process dependence in \Cref{eq:SDE_model} defines a directed \defined{dependency graph} $\cG = (V, \cE_{d})$ with vertices $V = [d] := \{1, \ldots, d\}$
representing the individual coordinate processes and where, for any $i,j\in[d]$, we include the directed edge $i\to j$ whenever $\mu^j$ or $\sigma^j$ is non-constant in its $i$-th argument.
For convenience, we introduce the following notation for the past and future increments of a coordinate process $X^k$ of a solution of \eqref{eq:SDE_model} on the subintervals $[0,s], [s,s+h]\subseteq[0,1]$:
\begin{align*}
\widetilde{X}^k=(X^k(t)-X^k(0))_{t\in[0,s]}, \qquad \hat{X}^k=(X^k(t)-X^k(s))_{t\in[s,s+h]}.
\end{align*}

\begin{proposition}[\defined{Global Markov property for $W$-separation}] Let $\{X_t^i\}_{i\in[d]}$ be the coordinate processes of a solution of \eqref{eq:integral_SDE_equation}, which satisfies (A1)-(A4) and $[0,s], [s,s+h]\subseteq I$ be subintervals for the fixed $s,h>0$. 
For $\mathcal{G}=(V,\mathcal{E}_d)$ the adjacency graph defined by \eqref{eq:integral_SDE_equation} and $A,B,C,D\subseteq V$, it holds
\begin{align}
A \indep^{\mathcal{G}}_{W} B \mid C,D \Rightarrow X^{A}_{[0,s]} \indep  \hat{X}^B \mid X^{C}_{[0,s]},\hat{X}^D.
\end{align}
\end{proposition}

\begin{remark} Note that information capture by past resp. future paths is the same as the sigma-algebras generated by the corresponding increments plus the initial value at the start of the interval:
\begin{align*}
\sigma (X^{C}_{[0,s]}) = \sigma (X^{C}_0,\widetilde{X}^C), \quad \sigma (X^{D}_{[s,s+h]}) = \sigma (X^{D}_s , \hat{X}^D).
\end{align*}
\end{remark}

\subsection{Proof of the Global Markov Property for $W$-separation}\label{app:proof_markov_property}
The proof of the global Markov property proceeds in the following steps:
\paragraph{Full pasts and future increments representation}  We represent the coordinate process of a solution of the SDE with the help of 
\begin{align*}
U^k(t):=X^k(t\wedge s), \qquad Q^k(t):=X^k(t)-X^k(t\wedge s).
\end{align*}
which satisfy 
\begin{align*}
U^v(t)=
\begin{cases}
X^v(t),&t\le s,\\
X^v(s),&t\ge s,
\end{cases}
\qquad
Q^v(t)=
\begin{cases}
0,&t\le s,\\
X^v(t)-X^v(s),&t\ge s.
\end{cases}
\end{align*}
allowing to express the coordinate process as the sum of its past and future increments:
\begin{align*}
X^{k}=U^k+Q^k.
\end{align*}
The whole path $U^k$ contains exactly the information in $X^k_{[0,s]}$ while the whole path $Q^k$ contains exactly the information in $\hat{X}^k$.
\begin{remark} We consider the full pasts and future increments representation in order for the future to start at zero and not contain the additive carry-over term $X_s^k$.
Thus dependence on the past remains whenever it enters through the drift or diffusion functionals. The graph representation differs from the one in \citet{boekenmooij2026SDEs}, which keeps the 
boundary value in the future component thus introducing self-loops at each node.
\end{remark}
\paragraph{Splitting the initial values and driving noises} In a similar fashion we can split initial values and driving noises into past and future increments:  For every coordinate $k$, define
\begin{align*}
W^{k,-}(t):=W^k(t\wedge s), \qquad W^{k,+}(t):=W^k(t)-W^k(t\wedge s), \qquad t\in I.
\end{align*}
meaning $W^{k,-}$ is the original Brownian motion up to $s$, remaining constant afterwards, while $W^{k,+}$ is zero up to $s$ and records the subsequent increments together giving
\begin{align*}
W^k=W^{k,-}+W^{k,+}.
\end{align*}
Note that each split driver is a continuous martingale in $\mathbb{F}$. As a consequence of independent-increments property and the mutual independence demanded in (A1) we have 
\begin{align*}
\bigl\{
\xi^k,W^{k,-},W^{k,+}:k\in[d]
\bigr\}
\end{align*}
to be mutually independent as random variables. Define the deterministic clocks
\begin{align*}
a^-(t):= t \wedge s := \min \lbrace t, s \rbrace, \qquad a^+(t):=\max \lbrace (t-s), 0 \rbrace.
\end{align*}
Then the stopped past and future increment processes satisfy
\begin{align*}\tag{A}
    U^k(t) = X_0^k + \int_0^t\mu^k(r,U)\,da^-(r) + \int_0^t\sigma^k(r,U)\,dW^{k,-}(r), \quad  Q^k(t) = \int_0^t\mu^k(r,U+Q)\,da^+(r) +\int_0^t\sigma^k(r,U+Q)\,dW^{k,+}(r).
\end{align*}
\paragraph{Equivalence of observational solutions:} $"\Leftarrow"$: an original solution $\left(X^{k} \right)_{k \in [d]}$ gives a solution $(U,Q)$ of (A) as for $r\le s$, the histories of $U$ and $X$ agree, so nonanticipativity gives
\begin{align*}
\mu^v(r,U)=\mu^v(r,X), \qquad \sigma^v(r,U)=\sigma^v(r,X).
\end{align*}
After $s$, the past clock and noise are constant. The future equation uses $U+Q=X$.\\
$"\Rightarrow"$: A solution $(U,Q)$ of (A) reconstructs an original solution $X$ of \eqref{eq:integral_SDE_equation}, as its future component $Q(t) = 0$ for $t \leq s$, and its past component is constant after $s$.
Adding the equations and using
\begin{align*}
a^-+a^+=t,\qquad W^{v,-}+W^{v,+}=W^v
\end{align*}
gives the original SDE for $X=U+Q$. Existence and uniqueness of the original system consequently give the observational solution of the constructed system.
\paragraph{Identify with the lifted dependence graph:} For the dependency graph $\cG$ created by the SDE, use its existing lift $\tilde{\cG}=(\tilde V,\tilde{\cE}_d)$, with
$\tilde V=V_0\sqcup V_1$, and identify its vertices with
\begin{align*}
 k_0\longleftrightarrow U^k,\qquad k_1\longleftrightarrow Q^k.
\end{align*}
\paragraph{Functional Representation:} To construct the measurable functionals between the strongly connected components, we can leverage \citet[Theorem~3.1]{Przybylowicz2024}, which is also used in \citet{boekenmooij2026SDEs}:
we set $Y=(U,Q)$ and define the set of noises as
\begin{align*}
 E^{k_0}:=(\xi^k,W^{k,-}),\qquad E^{k_1}:=W^{k,+}.
\end{align*}
For $v\in V$, consider its past strongly connected component
$\scc^{\tilde{\cG}}(v_0)\subseteq V_0$ and its external parents
\begin{align*}
 \pa^{\tilde{\cG}}\bigl(\scc^{\tilde{\cG}}(v_0)\bigr)
 \setminus\scc^{\tilde{\cG}}(v_0)\subseteq V_0.
\end{align*}
For a set of past nodes $A\subseteq V_0$, write $U^A:=(U^k)_{k_0\in A}$ and $E^A:=(E^{k_0})_{k_0\in A}$. The same coordinate convention holds for $\xi$, $W^-$ and future components.
Only the component and these external parents enter its subsystem.
The past subsystem then takes the form
\begin{align*}
 U^{\scc^{\tilde{\cG}}(v_0)}(t) =\xi^{\scc^{\tilde{\cG}}(v_0)}+\int_0^t
 \Bigl[
   \bigl(\mu^k(r,U)\bigr)_{k_0\in\scc^{\tilde{\cG}}(v_0)}\quad
   \operatorname{diag}_{k_0\in\scc^{\tilde{\cG}}(v_0)}
   \bigl(\sigma^k(r,U)\bigr)
 \Bigr]\,d\begin{pmatrix}
 a^-(r)\\
 W^{\scc^{\tilde{\cG}}(v_0),-}(r)
 \end{pmatrix}.
\end{align*}
As the non-anticipativity, Lipschitz and linear-growth bounds in (A2)--(A4) carry over to the $U$-subsystem  (where growth bounds may depend on the external input paths)
on the full c\`adl\`ag path space and $\bigl(a^-,W^{\scc^{\tilde{\cG}}(v_0),-}\bigr)$ is a $\mathbb F$-semimartingale, \citet[Theorem~3.1]{Przybylowicz2024} yields a Borel solution map
$\Gamma_{v_0}$ s.t.
\begin{align*}
 U^{\scc^{\tilde{\cG}}(v_0)}
 =\Gamma_{v_0}\Bigl(
 U^{\pa^{\tilde{\cG}}(\scc^{\tilde{\cG}}(v_0))\setminus\scc^{\tilde{\cG}}(v_0)},
 E^{\scc^{\tilde{\cG}}(v_0)}
 \Bigr)\quad P\text{-a.s.}
\end{align*}
by pathwise uniqueness of this subsystem.\\
Likewise, for $\scc^{\tilde{\cG}}(v_1)\subseteq V_1$, apply the same theorem to its $Q$-subsystem in (A), with zero initial value, driver $\bigl(a^+,W^{\scc^{\tilde{\cG}}(v_1),+}\bigr)$ and all external parent paths treated as inputs.
After substituting $X=U+Q$, (A2)--(A4) give the required hypotheses, with growth bounds allowed to depend on these inputs, so the theorem and subsystem pathwise uniqueness yield a Borel map $\Gamma_{v_1}$ satisfying
\begin{align*}
 Q^{\scc^{\tilde{\cG}}(v_1)}
 =\Gamma_{v_1}\Bigl(
 Y^{\pa^{\tilde{\cG}}(\scc^{\tilde{\cG}}(v_1))\setminus\scc^{\tilde{\cG}}(v_1)},
 E^{\scc^{\tilde{\cG}}(v_1)}
 \Bigr)\quad P\text{-a.s.}
\end{align*}
Here, the linear-growth bound follows from substituting $X = U + Q$ as 
\begin{align*}
\|\mu(t,x)\|\le L(1+\|x\|_t) \: \Rightarrow \:  \|\mu(t,u+q)\| \le \underbrace{L(1+\|u\|_t)}_{K(t,u)} (1+\|q\|_t).
\end{align*}

\paragraph{Application of the Markov-Property:} Having established the Borel SCC functionals, we apply \citet[Theorem~B.2]{Forre_Constraint} to $Y=(U,Q)$ and $\tilde{\cG}$.
Since $\tilde{\cG}$ has no bidirected edges, mutual independence of the noises $(E^w)_{w\in\tilde V}$ ensures the required noise Markov property (requirement (2.)); the SCC identities hold jointly $P$-a.s. 
because there are finitely many components. The theorem therefore yields
\begin{align*}
 A_0\indep_{\sigma}^{\tilde{\cG}}B_1\mid C_0\cup D_1 \: \Rightarrow \: U^A\indep Q^B\mid U^C,Q^D.
\end{align*}

\section{Additional Examples}\label{app:examples}

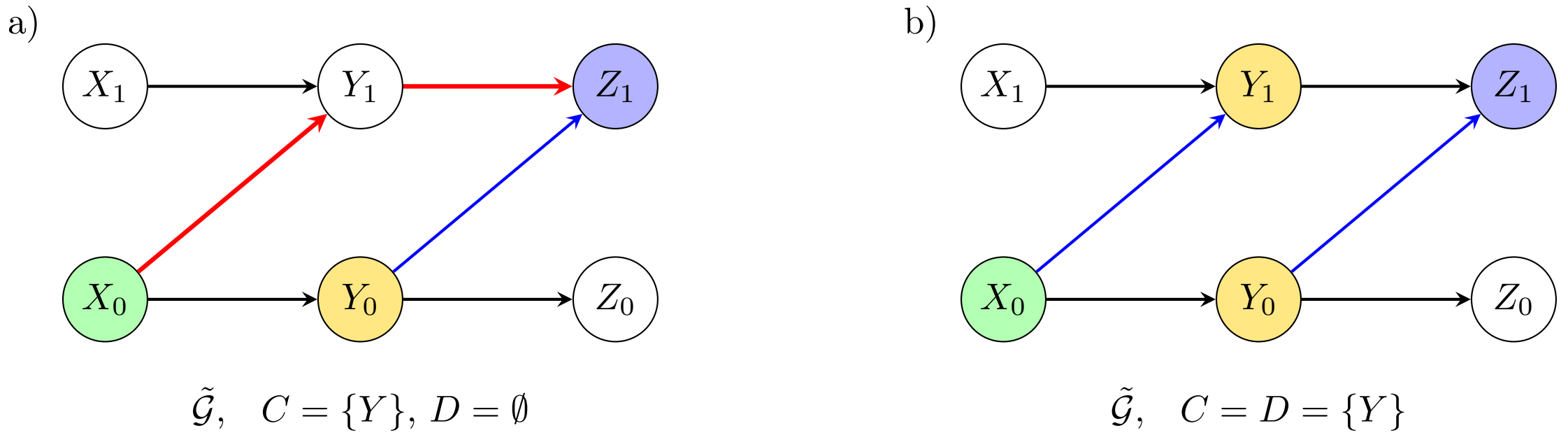
\begin{figure}[h]
    \centering
    \tikzset{
        nd/.style={circle, draw, minimum size=8mm, inner sep=0pt},
        cond/.style={nd, fill=Goldenrod!70},
        src/.style={nd, fill=green!30},
        tgt/.style={nd, fill=blue!30},
        e/.style={->, >=stealth, thick},
        xe/.style={e, blue},
        op/.style={e, red, very thick}
    }
    \begin{tikzpicture}[node distance=12mm and 16mm]
        \node[nd]                 (x1) {$X_1$};
        \node[nd,   right=of x1]  (y1) {$Y_1$};
        \node[tgt,  right=of y1]  (z1) {$Z_1$};
        \node[src,  below=of x1]  (x0) {$X_0$};
        \node[cond, right=of x0]  (y0) {$Y_0$};
        \node[nd,   right=of y0]  (z0) {$Z_0$};
        \draw[e] (x1) -- (y1); \draw[op] (y1) -- (z1);
        \draw[e] (x0) -- (y0); \draw[e] (y0) -- (z0);
        \draw[op] (x0) -- (y1); \draw[xe] (y0) -- (z1);
        \node[above left=0mm and 2mm of x1] {a)};
        \node[below=3mm of y0] {$\tilde{\cG}$, \; $C=\{Y\}$, $D=\emptyset$};
        \begin{scope}[xshift=85mm]
        \node[nd]                 (x1) {$X_1$};
        \node[cond, right=of x1]  (y1) {$Y_1$};
        \node[tgt,  right=of y1]  (z1) {$Z_1$};
        \node[src,  below=of x1]  (x0) {$X_0$};
        \node[cond, right=of x0]  (y0) {$Y_0$};
        \node[nd,   right=of y0]  (z0) {$Z_0$};
        \draw[e] (x1) -- (y1); \draw[e] (y1) -- (z1);
        \draw[e] (x0) -- (y0); \draw[e] (y0) -- (z0);
        \draw[xe] (x0) -- (y1); \draw[xe] (y0) -- (z1);
        \node[above left=0mm and 2mm of x1] {b)};
        \node[below=3mm of y0] {$\tilde{\cG}$, \; $C=D=\{Y\}$};
        \end{scope}
    \end{tikzpicture}
    \caption{Example where separation requires the future conditioning set $D$: lifted dependency graph $\tilde{\cG}$ of $X \to Y \to Z$; conditioned nodes are shaded yellow. a) Without future conditioning, the walk $X_0 \to Y_1 \to Z_1$ (red) remains open for every $C$, so $X$ and $Z$ cannot be separated. b) Conditioning on the future of $Y$ blocks it, and $X \indep_{W}^{\cG} Z \mid \{Y\},\{Y\}$.}
    \label{fig:future_conditioning}
\end{figure}

\end{document}